\documentclass[10pt,twocolumn,letterpaper]{article}

\usepackage{iccv}
\usepackage{times}
\usepackage{epsfig}
\usepackage{graphicx}
\usepackage{amsmath}
\usepackage{amssymb}

\usepackage[dvipsnames,table,xcdraw]{xcolor}
\usepackage[utf8]{inputenc} %
\usepackage[T1]{fontenc}    %
\usepackage{url}            %
\usepackage{amsfonts}       %
\usepackage{nicefrac}       %
\usepackage{microtype}      %
\usepackage{multirow}
\usepackage{bbm}
\usepackage{booktabs} 
\usepackage{rotating}
\usepackage[english]{babel}
\usepackage{pifont}%
\usepackage{comment}
\usepackage{soul}
\usepackage{algorithm}
\usepackage{algpseudocode}
\usepackage{capt-of} %

\definecolor{green}{RGB}{0,150,10}

\newcommand{\figLabel}{Figure~}
\newcommand{\eqLabel}[1]{{Eq (#1)}}
\newcommand{\secLabel}{Section~}

\newcommand{\mysection}[1]{\noindent\textbf{#1.}}
\newcommand{\supp}{\textit{Appendix}\xspace}

\newcommand{\sota}{state-of-the-art\xspace}

\newlength\mytmplen

\newcommand{\methodname}{UKBOB\xspace}
\usepackage[pagebackref=true,breaklinks=true,letterpaper=true,colorlinks,bookmarks=false]{hyperref}

\iccvfinalcopy %

\def\iccvPaperID{****} %

\ificcvfinal\fi

\title{\methodname: One Billion MRI Labeled Masks for Generalizable 3D Medical Image Segmentation}

\author{Emmanuelle Bourigault \quad\quad Amir Jamaludin \quad\quad  Abdullah Hamdi \\  
\normalsize Visual Geometry Group, University of Oxford \\
\tt\small{emmanuelle@robots.ox.ac.uk}\\
}
\begin{document}
\twocolumn[{
\renewcommand\twocolumn[1][]{#1}%
\maketitle
 \vspace{-2em}
\includegraphics[page=3, trim={0cm 0 0cm 0cm},clip, width=0.42\linewidth]{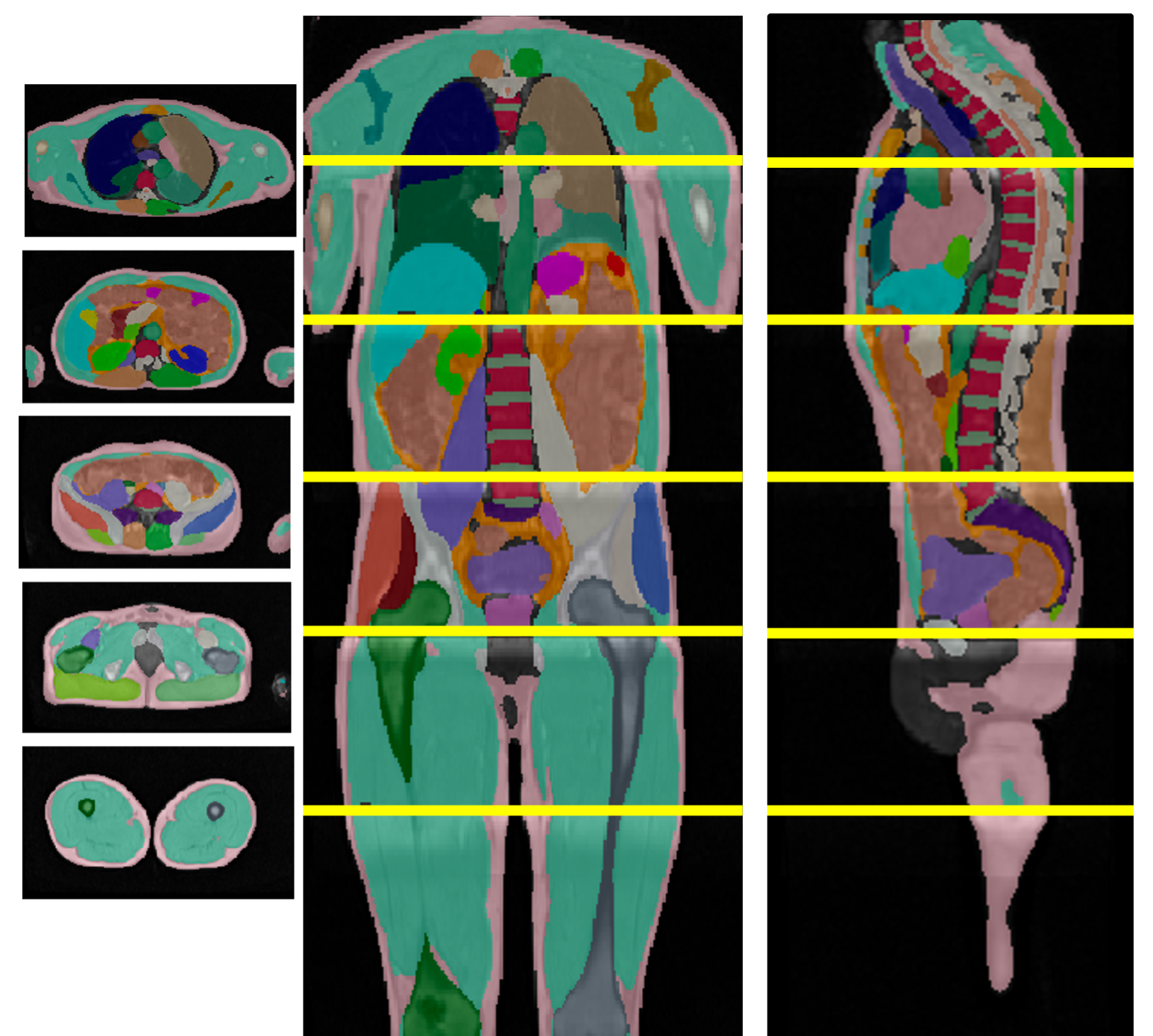} 
\includegraphics[width=0.57\linewidth,trim={1cm 0cm 2.75cm 0cm},clip]{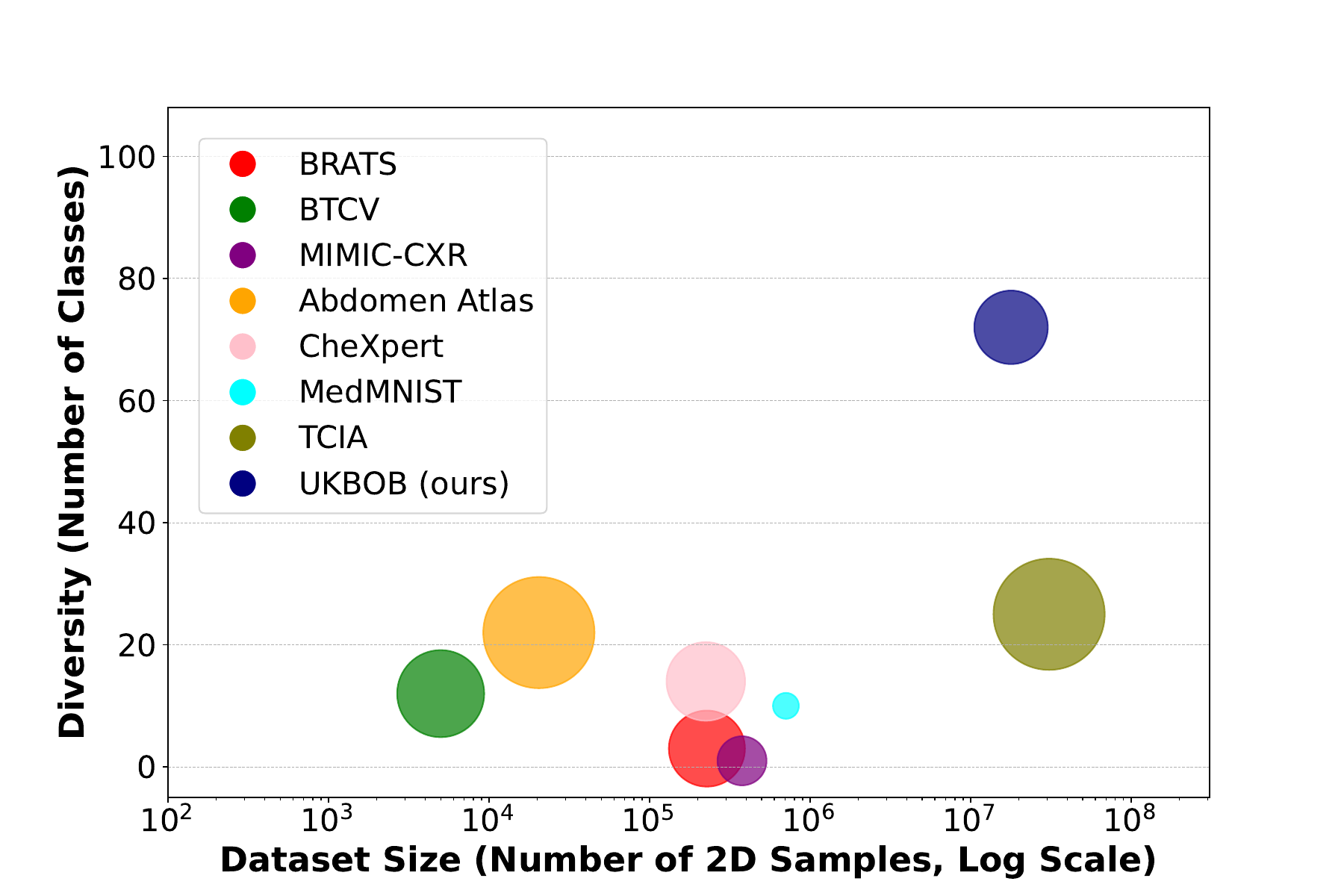}
\captionof{figure}{\textbf{\methodname Size and Diversity.} Our proposed UK Biobank Organs and Bones (\methodname) is the largest labeled medical imaging dataset for segmentation, comprising body organs of 51,761 MRI 3D samples (17.9 M 2D images) and a total of more than 1.37 billion 2D masks of 72 organs. \textit{Left}: we show label examples from \methodname from axial, coronal, and sagittal views. \textit{Right}: We show a plot of the size (number of 2D images) and diversity (number of classes) of our \methodname compared to other medical images datasets. The size of the bubbles indicates 2D image resolution. This new scale in dataset size and diversity should unlock a new wave of applications and methods in the computer vision and medical imaging communities.
\vspace{\baselineskip}}
\label{fig:pullfigure}
}]

\linespread{0.98}
\begin{abstract}
\vspace{-1.25\baselineskip}
In medical imaging, the primary challenge is collecting large-scale labeled data due to privacy concerns, logistics, and high labeling costs. In this work, we present the UK Biobank Organs and Bones (UKBOB), the largest labeled dataset of body organs of 51,761 MRI 3D samples (17.9 M 2D images) and a total of more than 1.37 billion 2D segmentation masks of 72 organs based on the UK Biobank MRI dataset. We utilize automatic labeling, introduce an automated label cleaning pipeline with organ-specific filters, and manually annotate a subset of 300 MRIs with 11 abdominal classes to validate the quality (UKBOB-manual). This approach allows for scaling up the dataset collection while maintaining confidence in the labels. We further confirm the validity of the labels by the zero-shot generalization of trained models on the filtered UKBOB to other small labeled datasets from a similar domain ( \eg abdominal MRI). To further elevate the effect of the noisy labels, we propose a novel Entropy Test-time Adaptation (ETTA) to refine the segmentation output. We use UKBOB to train a foundation model (\textit{Swin-BOB}) for 3D medical image segmentation based on Swin-UNetr, achieving \sota results in several benchmarks in 3D medical imaging, including BRATS brain MRI tumour challenge (+0.4\% improvement), and BTCV abdominal CT scan benchmark (+1.3\% improvement). The pre-trained models and the code are available at ~\url{https://emmanuelleb985.github.io/ukbob}, while filtered labels will be made available with the UK Biobank. 

\end{abstract}

\section{Introduction} \label{sec:introduction}
\vspace{-4pt}
The advent of large-scale labeled datasets such as ImageNet~\cite{IMAGENET} and LAION \cite{LAION} has been a cornerstone in the remarkable progress of computer vision, enabling the development of powerful foundation models~\cite{CLIP,LDM} that excel across various tasks. These models benefit immensely from the abundance of labeled data, which allows them to learn rich and generalizable representations. In stark contrast, the medical imaging domain grapples with a significant scarcity of large-scale labeled datasets due to stringent privacy regulations, complex logistics, and the high costs associated with expert annotations~\cite{ronneberger2015unet,nnUNet}. This limitation hampers the development of robust and generalizable models for critical tasks such as 3D medical image segmentation, which is essential for disease diagnosis, treatment planning, and patient monitoring.

Previous medical image datasets either lack diversity or are too small for generalization \cite{BTCV,Baid2021TheRB,chexpert}.  Recognizing the pressing need for extensive and diverse medical imaging datasets, we introduce the \textbf{UK Biobank Organs and Bones (\methodname)}, the largest labeled segmentation medical imaging dataset to date. Based on the UK Biobank MRI dataset~\cite{Sudlow2015UKBA}, \methodname\ comprises 51,761 3D MRI scans and over \textit{one billion} 2D segmentation masks covering 72 organs. This dataset not only surpasses existing medical imaging datasets in scale but also in anatomical diversity, providing an unprecedented resource for training robust and generalizable models (see \figLabel{\ref{fig:pullfigure}}). Table \ref{tbl:datasets} highlights the differences in scope and other aspects between the different datasets.
To collect the labels of \methodname, we leverage automatic labeling based on the newly released TotalVibe Segmentator \cite{graf2024totalvibesegmentator}. However, the automatic labeling of such a vast dataset introduces challenges related to label noise and quality assurance. To tackle this, we propose a novel mechanism for filtering organ labels based on a statistical Specialized Organ Labels Filter (SOLF). We also collect manual labels from 300 MRIs for 11 abdominal organs acting as validation (\methodname-manual). We further account for noisy labels and dynamically refine the segmentation based on the model's confidence using a novel Entropy Test-Time Adaptation (ETTA). These approaches ensure high-quality labels and enhances the model's robustness. We validate the validity of the labels by demonstrating zero-shot generalization of the trained models on the filtered \methodname\ dataset to other datasets from similar domains, such as the AMOS abdomen MRI dataset~\cite{AMOS} and the BTCV abdomen CT dataset~\cite{BTCV}.

Leveraging the extensive \methodname\ dataset, we train \textit{Swin-BOB}, a foundation model for 3D medical image segmentation based on Swin-UNetr~\cite{Hatamizadeh2022SwinUS}. Our model achieves state-of-the-art performance on several benchmarks in 3D medical imaging, including the BRATS brain tumor MRI challenge~\cite{Baid2021TheRB} and the BTCV abdominal CT scan benchmark~\cite{BTCV}. 
Our contributions can be summarized as follows:

\noindent\textbf{Contributions:} 
\textbf{(i)} We introduce UK Biobank Organs and Bones (\methodname), the largest labeled dataset of organs, consisting of 51,761 MRI 3D samples and a total of 1.37 billion 2D segmentation masks of 72 organs based on the UK Biobank MRI dataset.
\textbf{(ii)} We leverage automatic mechanisms for cleaning and filtering the labels based on body statistics and specialized organ filter, allowing for high-quality scale-up of the labels. The collected labels are validated by a subset of 300 manually annotated labels of 11 abdominal organs.
 \textbf{(iii)} To further elevate the effect of the noisy labels, we propose a novel Entropy Test-time Adaptation (ETTA) to refine the segmentation outputs.
\textbf{(iv)} We train \textit{Swin-BOB}, a foundation model for 3D medical image segmentation based on Swin-UNetr network~\cite{Hatamizadeh2022SwinUS}, achieving state-of-the-art results on standard benchmarks in 3D medical imaging.
\begin{table*}[t]
\centering
\resizebox{0.95\linewidth}{!}{
\tabcolsep=0.12cm
\begin{tabular}{l|cccccc}
\toprule
 & \multicolumn{6}{c}{\textbf{Medical Imaging Segmentation Datasets} } \\
\textbf{Attribute} & \textbf{BRATS} \cite{Baid2021TheRB} & \textbf{BTCV}\cite{BTCV} & \textbf{MIMIC-CXR \cite{Johnson2019MIMICCXRAD}}             & \textbf{Abd.Atlas} \cite{AbdomenAtlas} & \textbf{Total Segmentator} \cite{TotalSegmentator} &  ~~\textbf{\methodname (ours)} \\
\midrule
Number of Classes & 3 & 12   & 1 & 25           & 104 & 72  \\
\rowcolor[HTML]{EFEFEF} 
Number of 3D Samples & 1,470  & 50  & N/A & 20,460 & 1,204 &  \textbf{51,761} \\
Total Number of 2D Images & 227,850 & 5,000 & 377,110 & 673,000 & 400,000  & \textbf{17,902,080}     \\
\rowcolor[HTML]{EFEFEF} 
Number of 2D Label Masks &   581,715      &   425,000   &   N/A    & 16,825,000 & 5,800,000   & \textbf{1,378,913,040} \\
Number of Patients & 1,470 & 50 & 227,835 & N/A   & 1,204  & 50,000 \\
\rowcolor[HTML]{EFEFEF} 
Meta Information & N/A  & N/A  & Text Reports & N/A & N/A & Bone Density + Fat \%   \\
Scope & Brain   & Abdomen  & Chest & Abdomen & Full-Body & Full-Body \\
\rowcolor[HTML]{EFEFEF} 
Modality & MRI & CT & X-rays & CT & CT & MRI \\
Specialty & Tumour   & Organs  & COVID & Organs & Organs/Bones & Organs/Bones \\
\rowcolor[HTML]{EFEFEF} 
2D Image Resolution (axial) & $240 \times 240$ & $314 \times 214$ & $ 2500 \times 3056 $ & $ 280 \times 280$ & $ 512 \times 512$ & $224 \times 174$ \\
\bottomrule
\end{tabular}}
\vspace{2pt}
\caption{\small \textbf{Comparison of Different Medical Imaging Segmentation Datasets}. We compare our proposed \methodname to other well-known medical image segmentation datasets in terms of scope, size, and modality. 
}    \label{tbl:datasets}
\end{table*}

\section{Related Work} \label{sec:related}
\vspace{-4pt}
\mysection{3D Segmentation in Medical Imaging}
Advancements in deep learning have significantly influenced 3D data processing, leading to various approaches such as point-based methods \cite{pointnet, pointnet++}, voxel-based methods \cite{maturanaVoxNet3DConvolutional2015, minkosky}, and view-based methods \cite{mvcnn, mvtn, hamdi2023voint,tracknerf,mai2023egoloc,sparf,vars}.
In medical imaging, the U-Net architecture \cite{ronneberger2015unet} revolutionized image segmentation with its symmetric encoder-decoder structure and skip connections, becoming widely used for tasks such as organ segmentation and tumor detection \cite{Qayyum2017MedicalIA, signals3020018,kirillov2023segment}. For 3D volumetric data like MRI or CT scans, 3D U-Net variants have extended this architecture by replacing 2D operations with 3D counterparts, enhancing performance in volumetric segmentation tasks.
Recent developments in label-free segmentation utilize self-supervised learning and multimodal foundation models \cite{Zhang2022CVPR, Huang2023ICCV, Ha2022CORL, Peng2023CVPR, Kerr2023ICCV, Kobayashi2022NIPS, Ding2023CVPR, Lu2023CVPR, Zeng2023CVPR, Takmaz2023NIPS, Chen2023CVPR,mai2023egoloc} to segment 3D scenes without explicit labels. 
However, all of these models require large-scale medical datasets to generlize well~\cite{MIMICCXR,chexpert,MURA,MedMNISTv2}. While datasets like MIMIC-CXR \cite{MIMICCXR} and CheXpert \cite{chexpert} focus on chest imaging with extensive collections of X-ray images and associated clinical labels, they lack detailed segmentation masks necessary for advanced anatomical analysis. Datasets such as AbdomenAtlas-8K and Abdomen Atlas 1.1~\cite{abdomenatlas8k,abdomenatlas2024} provide valuable multi-organ CT scans with organ-level annotations but are limited to specific regions or modalities. %
The UK Biobank Imaging Study~\cite{Sudlow2015UKBA}, one of the largest clinical trials, has collected extensive MRI data; however, prior works have not fully leveraged its potential for comprehensive organ segmentation. Our proposed \textit{UK Biobank Organs and Bones (\methodname)} dataset leverages this resource, presenting the largest labeled collection of MRI scans with detailed segmentation masks for 72 organs. By introducing a novel filtering mechanism based on normalized body statistics, we ensure high-quality labels while scaling up dataset collection, enabling the training of foundational models for 3D medical image segmentation with significant improvements over existing benchmarks.

\begin{figure}[t]
  \centering
  \setlength{\tabcolsep}{4pt}
  \resizebox{0.99\linewidth}{!}{
  \begin{tabular}{ccccc}
       Spine GT~~ & 3D Spine GT & \methodname labels & ~~Spine label~ & ~~3D Spine~~ \\
    \end{tabular}
    }
           \includegraphics[page=7, trim={0cm 2cm 1cm 4.0cm},clip, width=1\linewidth]{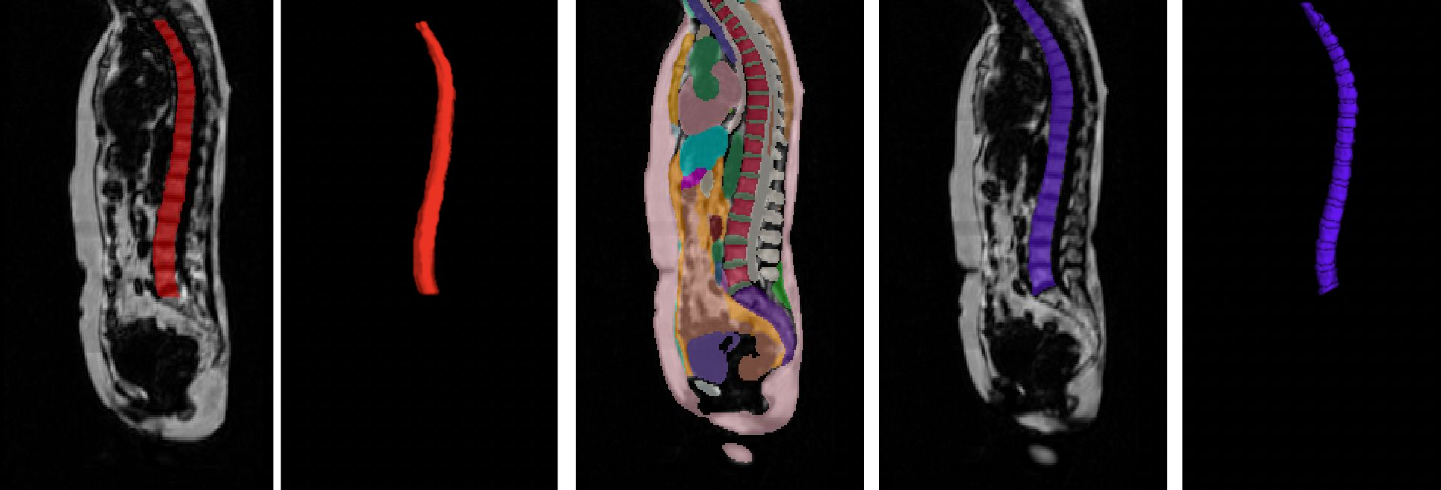}  \\
\caption{\textbf{Accuracy of \methodname Labels.}An example of segmentation labels in \methodname is shown in the sagittal view. The labels include ``\textit{spine}'' (in purple) which we can compare to previously collected hand labels of the spine  \cite{Bourigault23} (in red). We note that the newly collected labels match the manual labels in the spine with a total Dice score of 81.1\% on a set of 250 manually annotated test samples, indicating accurate labels. }
\label{fig:accuracy}
\end{figure}

\mysection{Full-Body MRI Analysis}
Most automatic MRI methods have focused on segmenting individual organs or tumors \cite{Chen2019FullyAM, Doran2017BreastMS, Windsor2020TheLA, Ranjbarzadeh2021BrainTS}, with limited research on whole-body scans. Studies that consider full-body imaging often emphasize the spine \cite{Jamaludin2017SelfsupervisedLF, Jamaludin2018PredictingSI, Windsor2020ACA, Windsor21, bourigault2022scoliosis,xdiffusion}, which is crucial for applications like scoliosis detection. Recently, 
\cite{graf2024totalvibesegmentator,totalsegmentation} released a full torso TotalVibeSegmentator first trained on a subset of NAKO (85 subjects) and UK Biobank (16 subjects) with a nnUNet\cite{nnUNet} network. Their network, while useful, does not provide rich enough information to show improved performance on medical image segmentation tasks. Our work builds upon these efforts by using the TotalVibeSegmentator network to collect labels for the 51,761 samples of UK Biobank %
, filtering MRI labels and verifying their segmentation quality. This enables the training of a general MRI foundation model (Swin-BOB) that can generalize to various tasks and modalities.

\mysection{Domain Adaptation in Medical Imaging}
While U-Net-like networks and their variants perform well in supervised medical image segmentation, significant performance degradation occurs when the test data differs from training data due to variations in protocols, scanners, or modalities \cite{medsam}. Test-time adaptation (TTA) addresses this by fine-tuning model parameters at test time using only test data without ground-truth \cite{KARANI_TTA}. Methods like TENT \cite{TENT} minimize prediction entropy at test time to improve robustness and segmentation performance.  Augmentation-based Test-Time Adaptation is proposed \cite{Zhang2020GeneralizingDL} to improve on the domain gap issue. 
However, these approaches rely on well-calibrated models and may be sensitive to augmentation procedures. Recent work \cite{Dong_2024_CVPR} integrates different predictions using various target domain statistics to enhance performance. Our work tries to address the issue of domain gap when the training domain has noisy labels making the adaptation even more challenging. Our model utilizes the confidence in predictions to adapt based on the entropy map on the test samples. This increases the model's robustness across a wider domain gap.

\begin{figure}[t]
  \centering
  \setlength{\tabcolsep}{4pt}
  \resizebox{0.99\linewidth}{!}{
    \begin{tabular}{cccc}
    \includegraphics[page=1, trim=0.0cm 0.1cm 0.0cm 0cm, clip, width=0.25\linewidth]{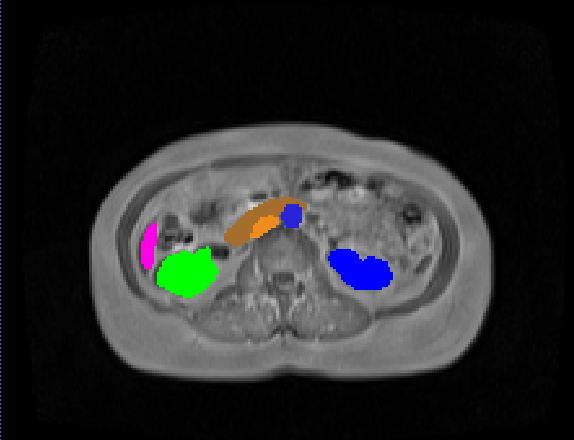} &
    \includegraphics[page=1, trim=0.0cm 0.1cm 0.0cm 0cm, clip, width=0.25\linewidth]{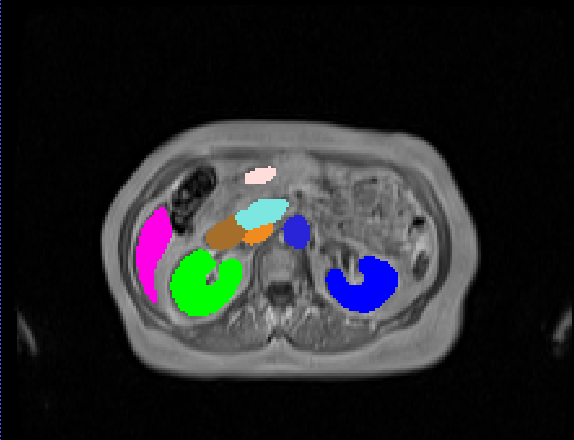} &
    \includegraphics[page=1, trim=0.0cm 0.1cm 0.0cm 0cm, clip, width=0.25\linewidth]{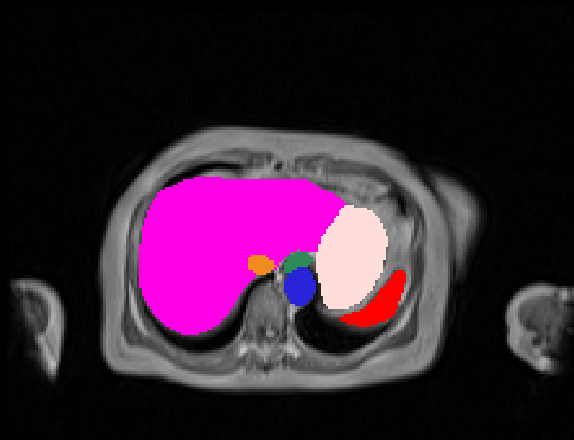} &
    \includegraphics[page=1, trim=0.0cm 0.1cm 0.0cm 0cm, clip, width=0.25\linewidth]{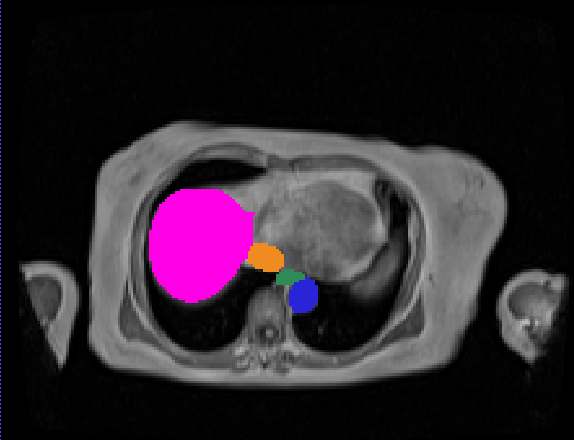} \\
    \multicolumn{4}{c}{%
    \includegraphics[page=1, trim=0.2cm 0.3cm 0.2cm 0cm, clip, width=1\linewidth]{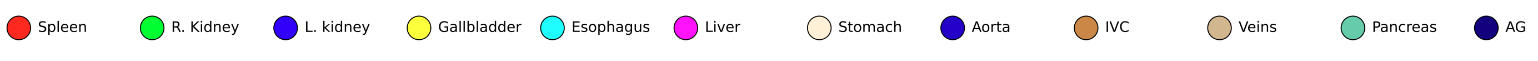}
  }
    \end{tabular}
    }
\caption{\textbf{\methodname-Manual.} We collect manual labels for 300 samples of \methodname for 11 abdominal organs totaling 3,000 images. \methodname-manual acts as manual validation for the large \methodname. Examples of axial slices are shown here.}
\label{fig:ukbob-manual}
\end{figure}
\section{Methodology} \label{sec:methodology}
\vspace{-4pt}
\subsection{\methodname Dataset Labels Collection}
\vspace{-2pt}
UKBiobank is a comprehensive dataset of 51,761 full-body MRIs from more than 50,000 volunteers\cite{Sudlow2015UKBA}, capturing diverse physiological attributes across a broad demographic spectrum. This dataset is unlabeled, which limits the potential applications for medical image understanding.  We construct the \methodname\ dataset by leveraging the UK Biobank MRI Study~\cite{Sudlow2015UKBA}, which consists of 51,761? neck-to-knee 3D MRI scans. Each scan includes four sequences: fat-only, water-only, in-phase, and out-of-phase images. To obtain segmentation labels for $C = 72$ organs, we employ the TotalVibeSegmentator~\cite{graf2024totalvibesegmentator}, an automatic segmentation tool trained on a subset of UK Biobank data. This approach allows us to generate over 1.37 billion 2D segmentation masks. Automatic labeling at this scale is crucial due to the impracticality of manual annotation. While it is not feasible to confirm the quality of 17.9M annotated images manually, we describe next a robust quality control mechanism on the collected labels to insure accurate labels. 

\begin{figure}[t]
    \centering
    \resizebox{\linewidth}{!}{  
    \begin{tabular}{cc|cc}
        \textbf{Low Spher.} & \textbf{High Spher.} & \textbf{Low Ecce.} & \textbf{High Ecce.}\\
        \includegraphics[trim={0.2cm 0cm 0cm 0.2cm},width=0.25\linewidth]{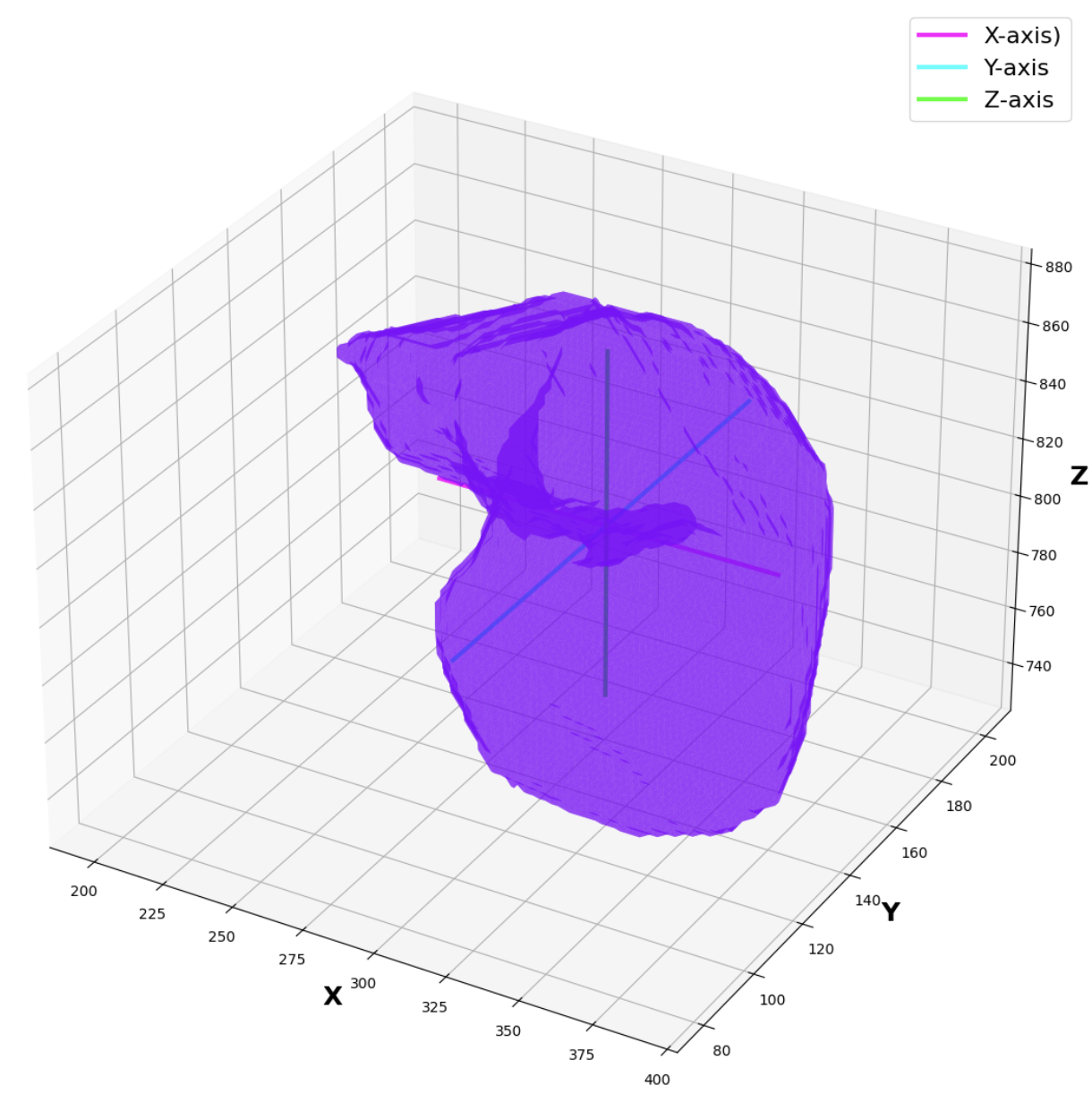} &
        \includegraphics[trim={0.2cm 0cm 0cm 0.2cm},width=0.25\linewidth]{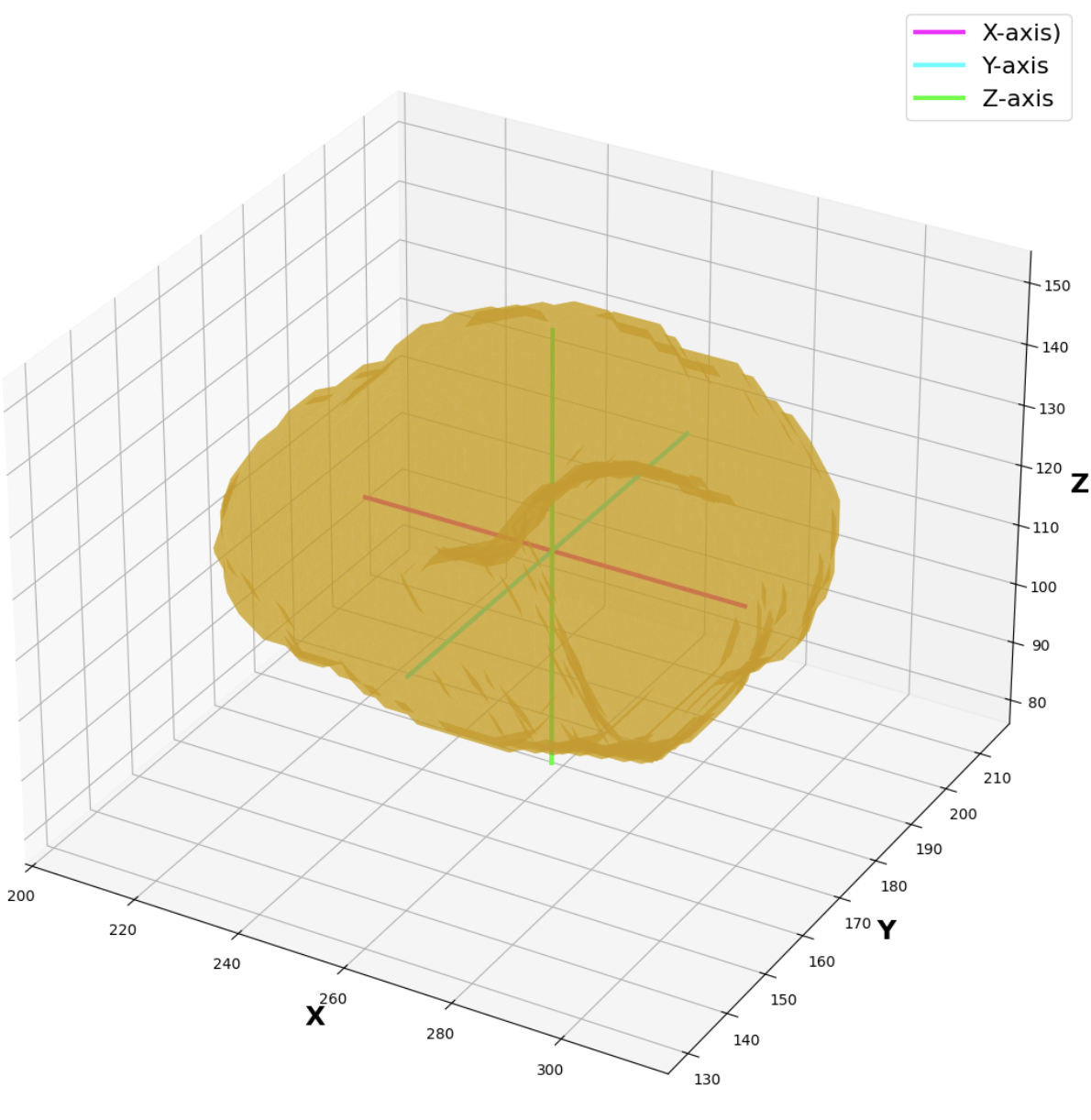} &
        \includegraphics[trim={2.5cm 0cm 0cm 0.2cm},width=0.25\linewidth]{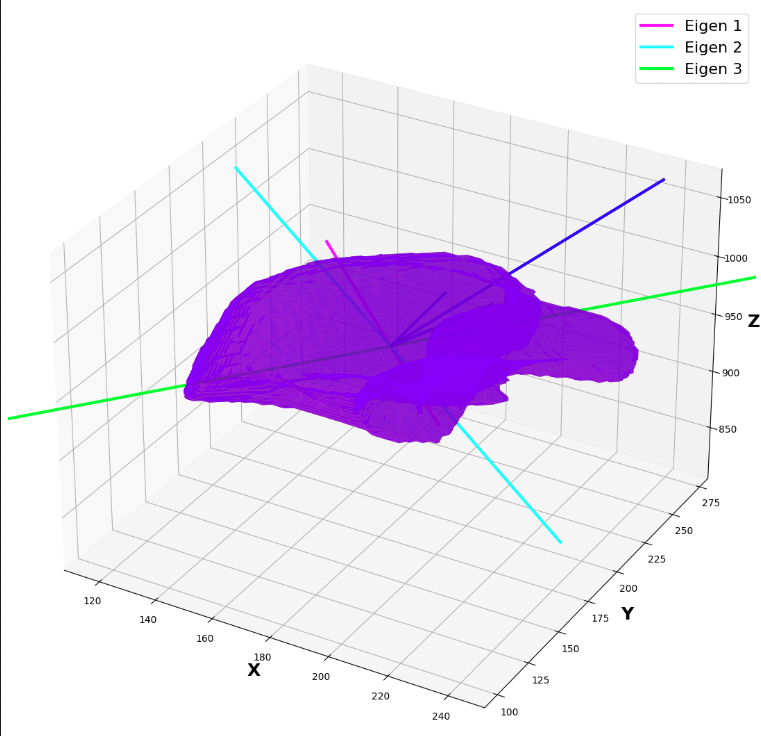}&
        \includegraphics[trim={0.2cm 0cm 0cm 0.2cm},clip,width=0.25\linewidth]{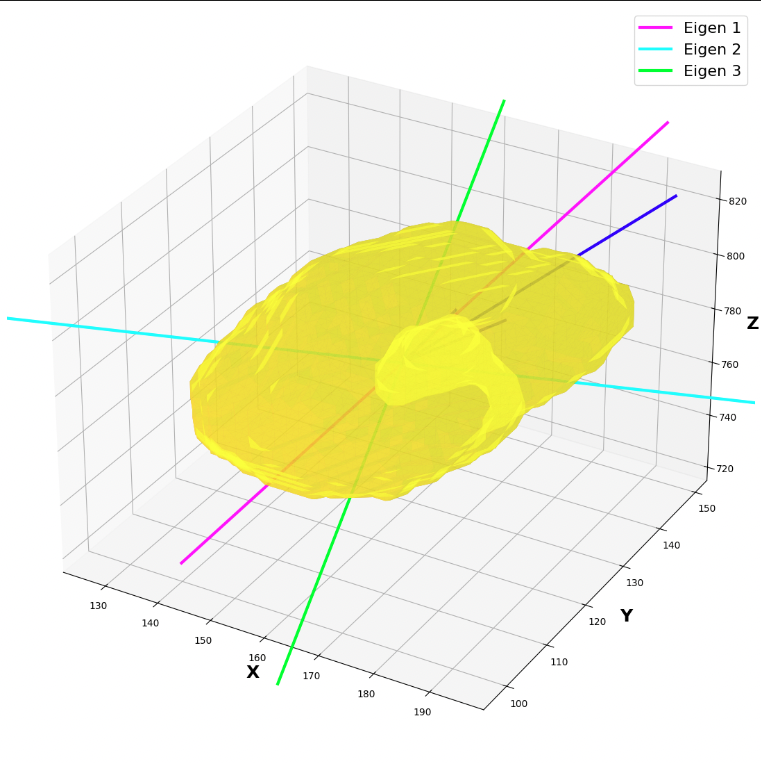}
    \end{tabular}
    }
\caption{\textbf{Specialized Organ Label Filter (SOLF).} SOLF integrates sphericity, eccentricity, and normalized volume to statistically filter out inaccurate organ labels. From left to right, the panels display examples of low sphericity (0.21), high sphericity (0.95), low eccentricity (0.14), and high eccentricity (0.87).}
    \label{fig:Sphericity}
\end{figure}
\subsection{Organ Labels Quality Control}
\label{sec:filtering}
\vspace{-2pt}
\mysection{Manual Labels for Verification}
We design a mechanism to validate these collected labels by humans. To do so we collect manual labels from 3000 2D image from 300 MRI samples for 10 abdominal organs (\methodname-manual). These manual labels (see examples in \figLabel{\ref{fig:ukbob-manual}}) act as a validation for the large \methodname dataset. On these labels, the UKBOB automatic labels obtain an average Dice Score of 0.891 (see Table~\ref{table:manual_labels}). Furthermore, we verify the spine labels of \methodname using previously collected manual labels of 200 3D spine labels \cite{Bourigault23}. 
We show an example in \figLabel{\ref{fig:accuracy}} and we see how the new collected labels match the manual labels in the spine with a total Dice score of 0.811, indicating accurate labels. We discuss in \secLabel{\ref{sec:zeroshot}} another mechanism for verifying the labels by zero-shot generalization of trained models to other similar datasets that has manual labels. 

\mysection{Specialized Organ Label Filter (SOLF)}
While automatic labeling enables creating large datasets, it introduces the possibility of noisy or erroneous labels. To mitigate this, we propose a filtration mechanism that removes outliers from segmentation. A question arises on how to distinguish segmentation failures from common patient abnormalities, \eg enlarged liver. It's important to note that human organs follow typical geometric properties that arise from the body's need to optimize function while minimizing energy expenditure and structural stress. They reflect the underlying biological "blueprint" that has been honed by evolution \cite{shetty2023boss}. inspired by the evolutionary regularity of human organs \cite{shetty2023boss}, we propose the Specialized Organ Label Filter (SOLF), using three features \textit{jointly}: \textit{normalized volume}, \textit{eccentricity}, and \textit{sphericity} (illustrated in \figLabel{\ref{fig:Sphericity}}). 

For each organ class $c$, the normalized volume for some 3D sample is computed as 
$
v_c = \frac{V_c}{V_{\text{body}}},
$ 
where $V_c$ is the voxel count for the organ $c\in \{1, 2, \dots, C\}$ and $V_{\text{body}}$ is the total body voxel count. We define acceptable bounds for each feature by excluding the extreme $\epsilon$ percentiles. For example, the bounds for volume are set as
\begin{equation} \label{eq:filteration}
v_c^{\text{min}} = P_{\epsilon/2}\left(\{v_c\}_{n=1}^{N}\right), ~~
v_c^{\text{max}} = P_{100-\epsilon/2}\left(\{v_c\}_{n=1}^{N}\right)
\end{equation}
, where $P_p(\cdot)$ denotes the $p$-th percentile function. 
Sphericity is defined as $\Phi_c=\pi^{1/3}(6V_c)^{2/3}/A_c$, with $V_c$ computed from voxel counts and $A_c$ as the surface area measured by counting the exposed voxel faces of the organ. Finally, eccentricity is defined as $E_c=\sqrt{1-\lambda_{\min}/\lambda_{\max}}$, where $\lambda_{\min}$ and $\lambda_{\max}$ are the smallest and largest eigenvalues of the covariance matrix of organ $c$ voxel coordinates. 
A sample is flagged as \texttt{inaccurate} if at least two of the three features (normalized volume, eccentricity, and sphericity) fall outside their respective acceptable ranges. Setting $\epsilon$ for SOLF  effectively discards samples with anomalous organ characteristics while retaining valid labels. A single patient with abnormal organs is extremely unlikely to have more than a single independent aspect of deviation \textit{at the same time}, hence indicating inaccurate labels.

We filter collected organ labels using the \textit{patient's full-body statistics}, a novel approach compared to previous methods that rely on flat label statistics (IQR) \cite{cheng2024interactiveseg,kucs2024medsegbench} rather than patient meta-information and organ-specific features .

\begin{figure}[t]
  \centering
  \includegraphics[page=3,trim={0cm 0 0cm 0cm},clip, width=1\linewidth]{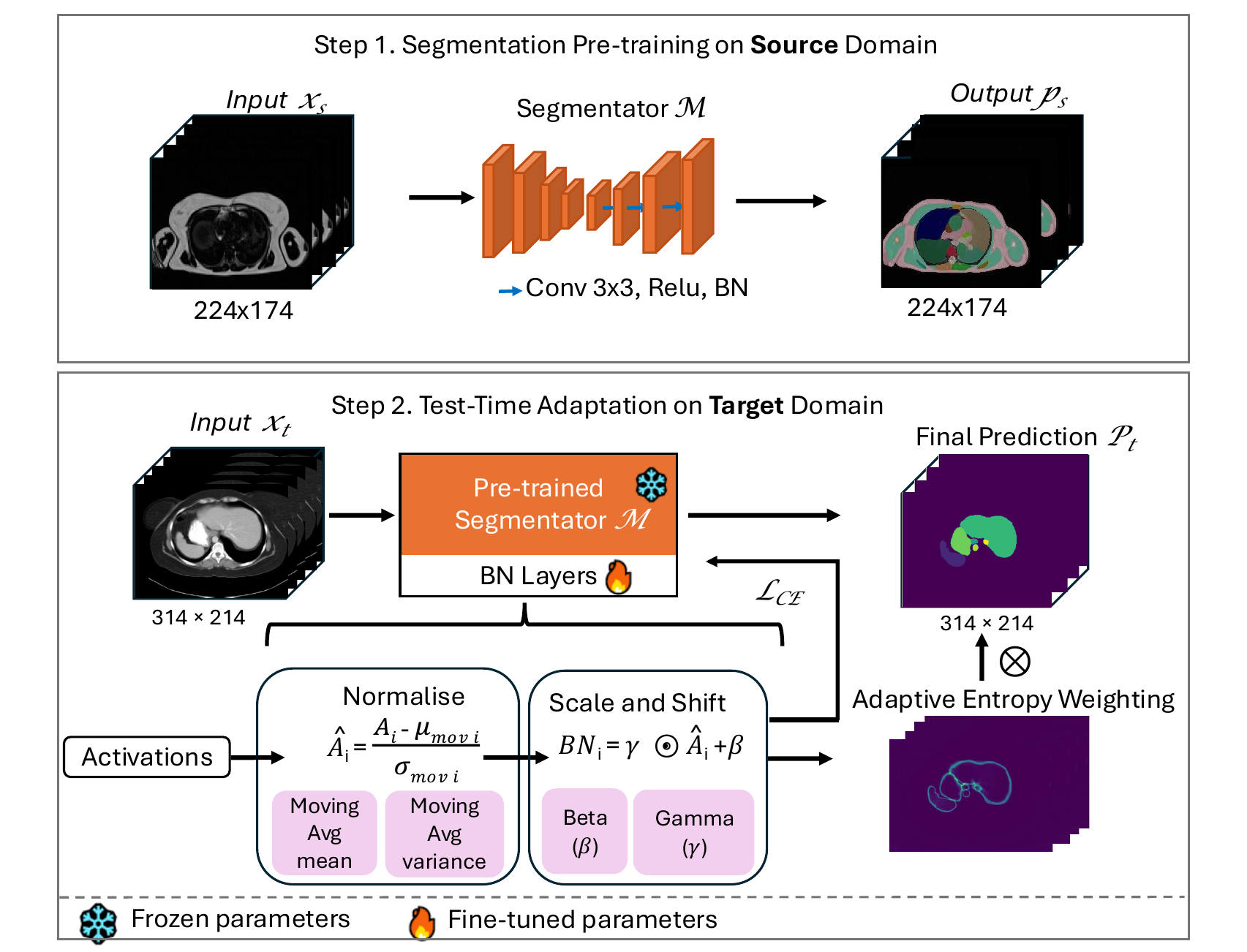}
\caption{\textbf{Entropy Test-Time Adaptation for Image Segmentation.} We use a test-time entropy map to refine the batch norm layer of the network for robust segmentation output. This module is agnostic to the architecture of the deep neural network. Therefore, It can be used with any segmentation network to increase consistency and robustness, especially when trained with noisy labels. 
 }
\label{fig:uncertainty}
\end{figure}
\begin{figure}[t]
    \centering
    \resizebox{0.9\linewidth}{!}{  
    \begin{tabular}{ccc}
        \textbf{Input} & \textbf{Prediction} & \textbf{Entropy Map}\\
        \includegraphics[width=0.25\linewidth]{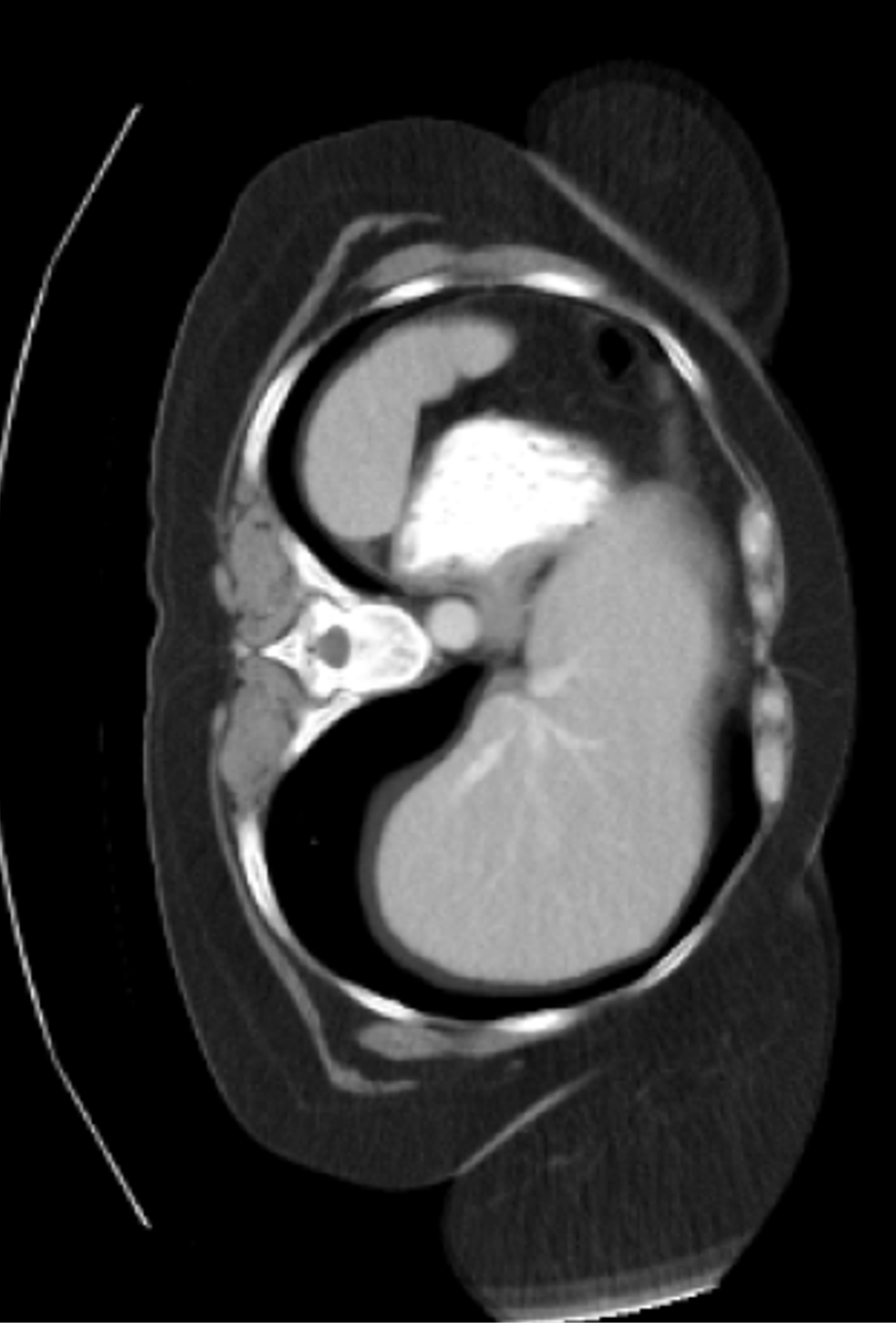} &
        \includegraphics[width=0.25\linewidth]{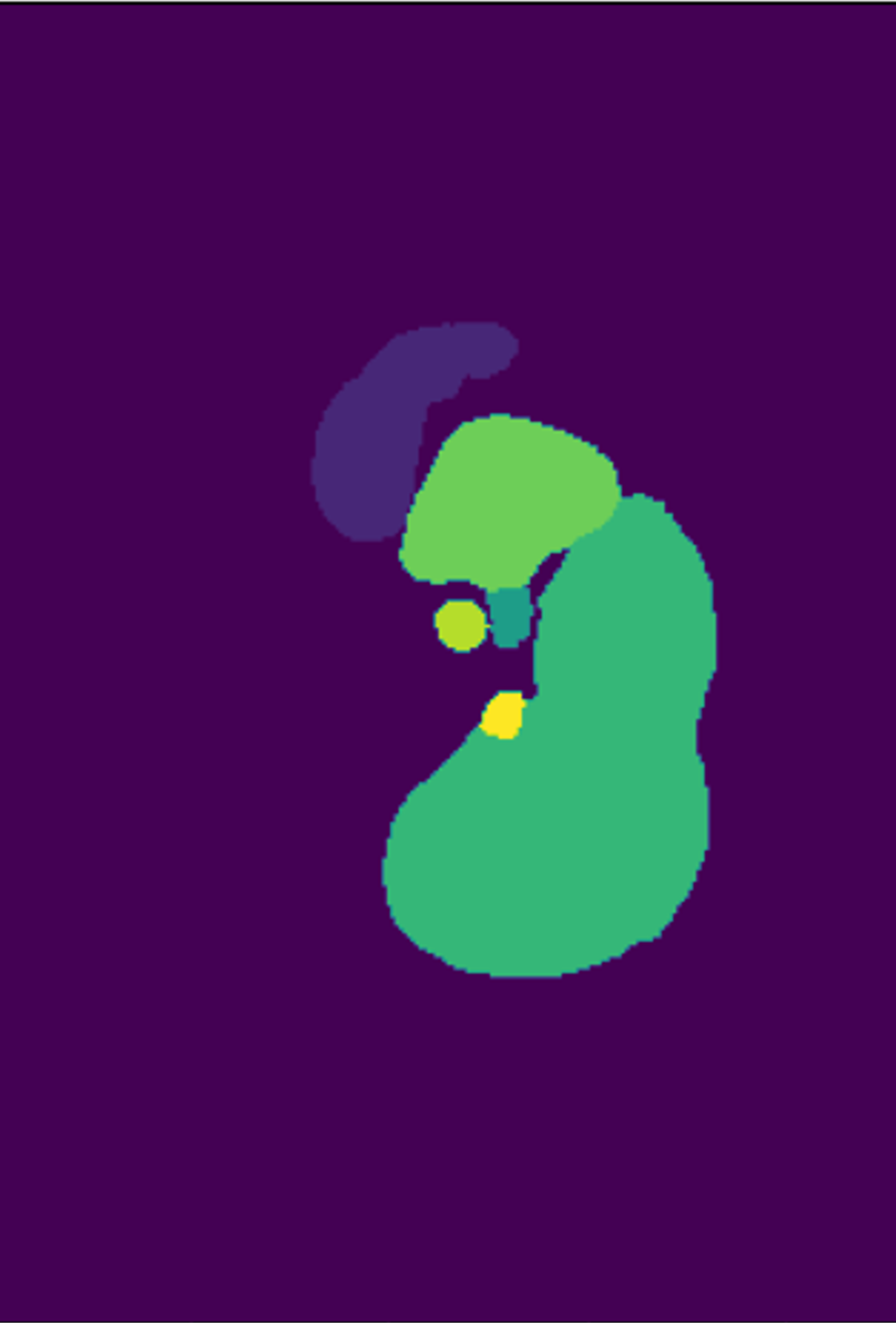} &
        \includegraphics[width=0.25\linewidth]{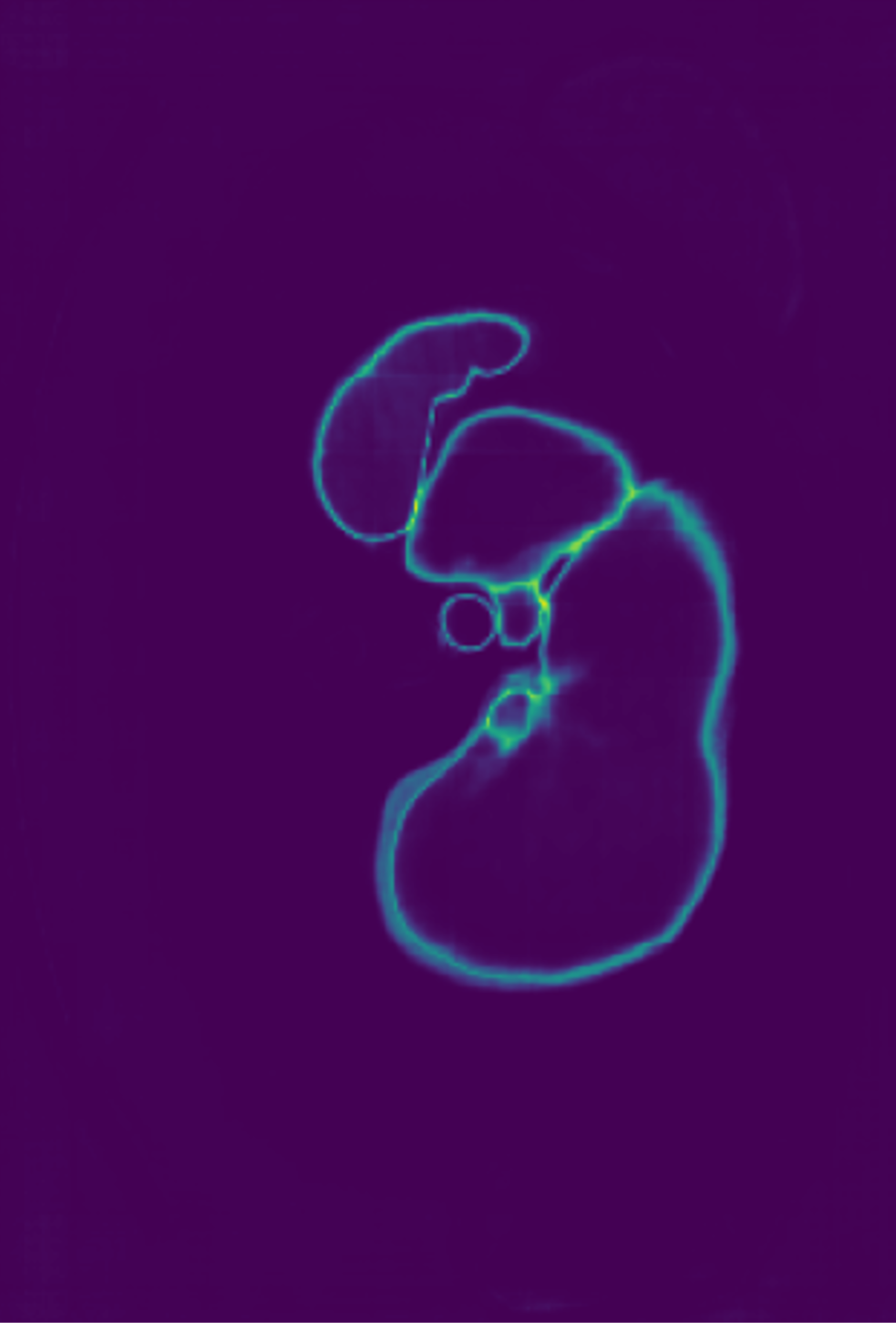}
    \end{tabular}
    }
    \caption{\textbf{Entropy Map Visualization.} We show from left to right an example of input, prediction and entropy map used in the Entropy Test-Time Adaptation on BTCV dataset \cite{BTCV} that can be leveraged to refine the output. Brighter regions indicate higher entropy. %
    }
    \label{fig:btcv_entropy}
\end{figure}

\subsection{Entropy Test-Time Adaptation (ETTA)}
A common practice in medical imaging to addresses the labels distribution-shift is to employ Test-Time Adaptation (TTA) by fine-tuning model parameters at test time using only test data without ground-truth \cite{KARANI_TTA}. To mitigate the impact of any residual label noise at \methodname, we introduce Entropy Test-Time Adaptation (ETTA). ETTA refines the model's predictions on test samples by fine-tuning the batch normalization parameters using the test data itself, guided by minimizing the prediction entropy.
Given a test sample $\mathbf{x}$, we first obtain the network's initial prediction $\mathbf{p} = f_{\theta}(\mathbf{x})$, where $\mathbf{p} \in [0,1]^{N \times C}$ is the softmax probability over $C$ classes at each of the $N$ voxels in the sample. Here, $f_{\theta}$ represents the segmentation network parameterized by $\theta$.
We define the entropy loss $\mathcal{L}_{\text{ent}}$ over the predicted probabilities:
\begin{equation}
\mathcal{L}_{\text{ent}} = - \frac{1}{N} \sum_{i=1}^{N} \sum_{c=1}^{C} p_{i,c} \log p_{i,c}
\end{equation}
where $p_{i,c}$ is the probability of class $c$ at voxel $i$.

We update only the batch normalization parameters $\theta_{\text{BN}}$ while keeping the other network parameters $\theta_{\text{fixed}}$ frozen (see pipeline in Figure~\ref{fig:uncertainty}). The adaptation process involves minimizing the entropy loss with respect to $\theta_{\text{BN}}$:
\begin{equation}
\theta_{\text{BN}}^{*} = \arg\min_{\theta_{\text{BN}}} \mathcal{L}_{\text{ent}}\left(f_{\theta_{\text{fixed}},\, \theta_{\text{BN}}}(\mathbf{x})\right)
\end{equation}
This process adapts the model to the test sample by encouraging confident (low-entropy) predictions, thereby refining the segmentation output.
The adaptation is efficient as it involves updating a small subset of parameters and can be performed online during inference. ETTA leverages the entropy of the network's predictions (\figLabel{\ref{fig:btcv_entropy}}) to guide the adaptation, improving robustness to domain shifts and label noise.

\section{Experiments} \label{sec:experiments}
\vspace{-4pt}

\subsection{Evaluation Datasets and Metrics} \label{sec:datasets}
\vspace{-2pt}

\mysection{Evaluation Datasets}
We evaluate our model on multi-modal publicly available datasets i.e AMOS~\cite{AMOS} of 13 abdominal organs from 100 MRI scans split equally into train and test sets, BTCV (Beyond the Cranial Vault) abdomen CT dataset~\cite{BTCV} of 30 training and 20 testing subjects and 13 labelled organs, and BRATS~\cite{Baid2021TheRB,4b589b6824a64a2a91e8e3b26cc0bf9e,41847efe8ced40078c67adce2164d865}, the largest publicly available dataset for brain tumors 5,880 MRI scans and corresponding annotations.

\mysection{Evaluation Metrics}
We evaluate our model using the Dice Score and the Hausdorff Distance Metric, which are widely used in medical image segmentation \cite{medsam, karimi2019reducing}. The \textit{Dice Score} measures the overlap between predicted and ground truth masks, while the \textit{Hausdorff Distance} assesses the boundary discrepancy, providing a comprehensive evaluation of segmentation performance.

\subsection{Baselines}
\vspace{-2pt}
To evaluate our model, we compared it with a variety of established baselines across the BTCV, BRATS, and UKBOB benchmarks. For the BTCV dataset, baseline models included UNet \cite{ronneberger2015unet}, SegResNet \cite{SegResNet}, TransUNet \cite{transunet}, UNetr \cite{UNETR}, Swin-UNetr\cite{Hatamizadeh2022SwinUS}, nn-UNet \cite{nnUNet}, and AttentionUNet \cite{AttentionUNet}. We also evaluate the base TotalVibe Segmentator (TVS) \cite{graf2024totalvibesegmentator} in zero-shot and finetuning settings. These models represent widely adopted architectures in medical image segmentation, offering a range of network designs from classic CNN-based approaches to transformer-based architectures. For the BTCV dataset, diffusion-based segmentation methods were also considered, including MedSegDiff\cite{MedSegDiff} and  MedSegDiff-V2\cite{MedSegDiffV2}. Together, these baselines provide a comprehensive foundation for evaluating our model’s performance in 3D medical segmentation.
On the BRATS2023 benchmark, additional baselines incorporated V-Net \cite{VNet}, ResUNet++ \cite{ResUNet++}, and nnFormer \cite{nnFormer}, along with UNETR \cite{UNETR} and Swin-UNetr \cite{Hatamizadeh2022SwinUS}. These models were selected to encompass a spectrum of segmentation methods specifically suited to brain tumor segmentation tasks.

\begin{table}[t]
    \centering
    \resizebox{1\linewidth}{!}{%
    \begin{tabular}{r|c|c}
        \hline
        \multicolumn{1}{c}{\textbf{Model}} & \textbf{Dice Score} & \textbf{Hausdorff Distance} \\
        \hline
        ResUNet++\cite{ResUNet++} & 0.876 & 9.431\\
        MedFormer \cite{wang2024medformer} &  0.881 & 8.822 \\
        nnUNet\cite{nnUNet} & 0.915 & 6.442\\
        UNetr\cite{UNETR} &0.902 & 7.968 \\
        Swin-UNetr\cite{Hatamizadeh2022SwinUS} & \textbf{0.918} & \textbf{5.984}\\
        \hline
    \end{tabular}
    }
    \vspace{2pt}
    \caption{\textbf{\methodname 3D Segmentation Benchmark.} We show results on test mean Dice Score (\%) and mean Hausdorff Distance ($n=72$ classes) of our proposed benchmark on \methodname. Note how Swin-UNetr \cite{Hatamizadeh2022SwinUS} achieves the best results, resulting in our Swin-BOB foundation model.}
    \label{tab:segmentation_comparison_ukbb}
\end{table}

\subsection{Implementation Details} 
\label{sec:details}
\vspace{-2pt}

\mysection{Pre-training} 
For pre-training on UKBiobank, we split the dataset into 80-10-10 for training, validation, and testing.
The input is cropped to 96×96×96 voxels from the 3D MRI. In training, for data augmentation, scans are intensity scaled, with random flipping along the 3 axes, random foreground cropping, random rotation 90 degrees with probability 10\%, and random intensity shift with an offset of 0.1. In validation, scans are intensity-scaled.
We use a batch size of 8, AdamW optimizer, $\beta_{1}$ = 0.9, $\beta_{2}$ = 0.999, and cosine learning rate scheduler with a warm 
restart every 200 epochs. We start with the initial learning rate of 
$10^{-4}$ and decay of $10^{-5}$. We train the model on 2 A6000 GPU 
for 3,000 epochs. We use binary Cross-Entropy and Dice Similarity Coefficient (DSC) as our loss function. The best model achieving the highest overall Dice score on the 72 classes is saved at validation. 

\mysection{Fine-tuning} The pre-trained Swin-BOB is used on BTCV, BRATS, and AMOS, keeping the same configuration as in training while reducing the warm-up scheduler to 50 epochs for a total of 500 epochs. 
\begin{table}[t]
    \centering
    \resizebox{0.99\linewidth}{!}{%
    \begin{tabular}{c|cccc}
    \hline
    \textbf{Configuration}  & No Filter & IQR filter & \textbf{SOLF Filter} \\ \hline
    Spine labels \cite{Bourigault23}  &   0.811      &  0.849   &   0.867  \\
    \methodname-manual   &    0.873      & 0.877                    & 0.891                   \\ \hline
    \end{tabular}
    }
    \vspace{2pt}
        \caption{\textbf{Precision of the collected \methodname compared to Manual labels.} We show the Dice Score of the collected \methodname labels on subsets of manual labels on UKBOB for 200 Spines \cite{Bourigault23} and on 300 manual abdominal labels we collect (\methodname-manual). Even without any filtration, \methodname labels are precise, improving in precision with our designed SOLF statistical filter.}
    \label{table:manual_labels}
\end{table}

\section{Results} \label{sec:results}
\vspace{-4pt}
\subsection{Validating \methodname Labels} 
\vspace{-2pt}
\mysection{Manual Verification} In Table \ref{table:manual_labels} we validate the quality of the \methodname labels and the organ quality control against manual spine annotation from \cite{Bourigault23} and our manually annotated 11 abdomen UKBOB organs (\methodname-manual). Even without any filtration, \methodname labels are precise, achieving Dice score of 0.811 and 0.873 on manual spines and \methodname-manual respectively. We Also show that our SOLF filtering approach (when $\epsilon=2$) increases Dice score by 0.056 compared to no filtering on the spine labels and by 0.018 on the labelled abdomen organs. We also show standard inter-quartile range filtering (IQR) \cite{AMOS} for comparison.

\mysection{Zero-Shot Evaluations}  \label{sec:zeroshot}
We also rely on the zero-shot segmentation performance of a model trained solely on those collected labels to be evaluated on a similar domain, namely the AMOS Abdomen MRI dataset \cite{AMOS} and on MRI on BTCV dataset\cite{BTCV} in Table~\ref{table:zero_shot_segmentation_detailed}. AMOS shares 12 class labels with \methodname while BTCV shares 11 class labels. In Table~\ref{table:zero_shot_segmentation_detailed}, we show 10 organs overlap between BTCV and AMOS where we omitted small organ i.e duodenum as it has also been omitted in baseline papers. We combined the left and adrenal gland into one class named AG. We train Swin-UNetr \cite{Hatamizadeh2022SwinUS} from scratch  on different filtration schemes of the \methodname and run the evaluation on the test sets given by BTCV and AMOS (on the shared class labels) and report the mean Dice score and mean Haussdorff distance. We adjust the preprocessing (normalization to [0,1], and resizing) to ensure compatibility with the model’s pretrained dataset. These results highlight the importance of the filtration we followed ensuring better quality labels. 

\begin{table}[t]
    \centering
    \resizebox{1\linewidth}{!}{%
    \begin{tabular}{r|c|c}
        \hline
        \multicolumn{1}{c}{\textbf{Model}} & \textbf{Dice Score} & \textbf{Hausdorff Distance} \\
        \hline
        UNet\cite{ronneberger2015unet}& 0.544 & 39.090 \\
        V-Net\cite{VNet} & 0.842 & 10.891 \\
        ResUNet++\cite{ResUNet++} & 0.784 & 22.249 \\
        AttentionUNet\cite{AttentionUNet} & 0.798 & 20.048 \\
        nnFormer\cite{nnFormer} & 0.812 & 10.070 \\
        UNETR\cite{UNETR} & 0.871 & 9.924 \\
        SegResNet\cite{SegResNet} & 0.890 & 8.650 \\
        Total Vibe Seg. \cite{graf2024totalvibesegmentator} & 0.830 & 8.973 \\
        Swin-UNetr\cite{Hatamizadeh2022SwinUS} & 0.886 & 9.016 \\
        \textbf{Swin-BOB (ours)} & \textbf{0.894} & \textbf{8.650} \\
        \hline
    \end{tabular}
    }
    \vspace{2pt}
    \caption{\textbf{BRATS 3D Segmentation Benchmark.} The proposed Swin-BOB model, pre-trained on UK Biobank organs and fine-tuned on BRATS2023 \cite{Baid2021TheRB} archives \sota results on test mean Dice Score (\%) and mean Hausdorff Distance ($n=3$ classes). Baseline results are reported from the Swin-UNetr paper \cite{Hatamizadeh2022SwinUS}.}
    \label{tab:segmentation_comparison_brats}
\end{table}
\subsection{Swin-BOB: A Foundation Model for 3D Medical Image Segmentation} 
\begin{figure*}[t]
  \centering
   \resizebox{\linewidth}{!}{  
   \begin{tabular}{ccccccccc}
        \textbf{Input} & \textbf{GT} & \textbf{Swin-BOB} & \textbf{TransUNet} & \textbf{UNetr} & \textbf{Swin-UNetr} & \textbf{nn-UNet} & \textbf{Tot.Vibe.Seg.} \\

      \scalebox{-1}[1]{\includegraphics[width=0.111\linewidth]{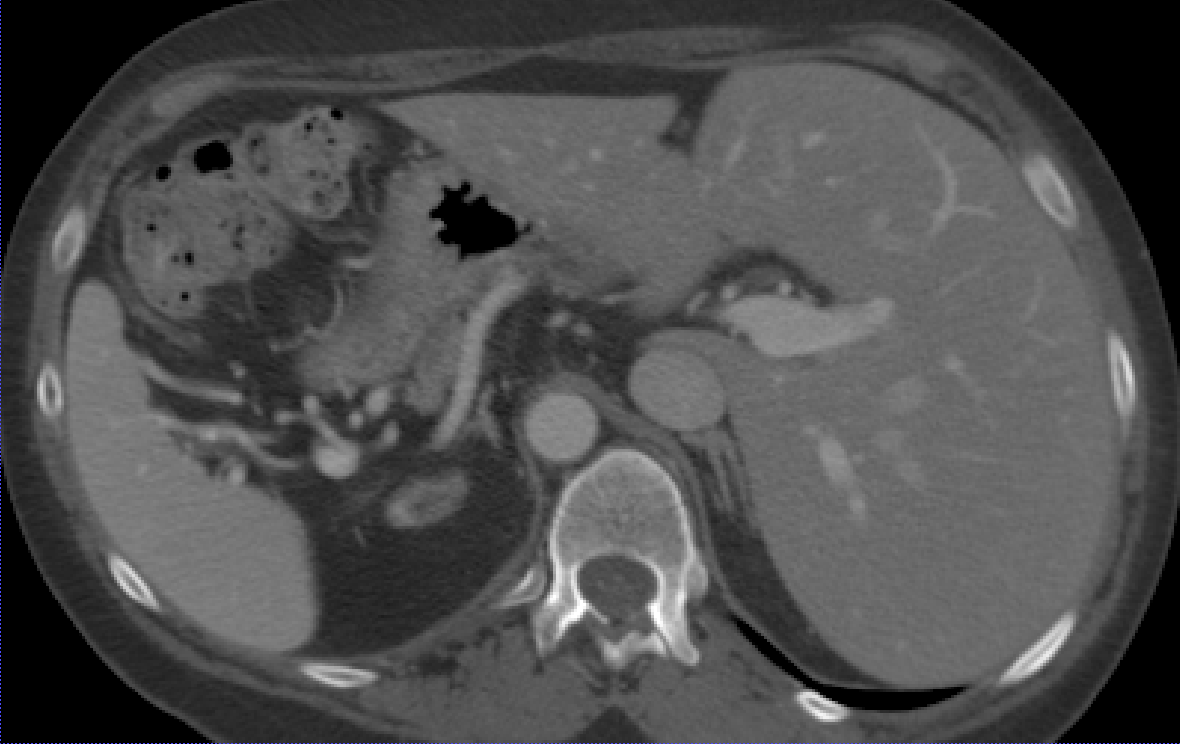}}&
      \scalebox{-1}[1]{\includegraphics[width=0.111\linewidth]{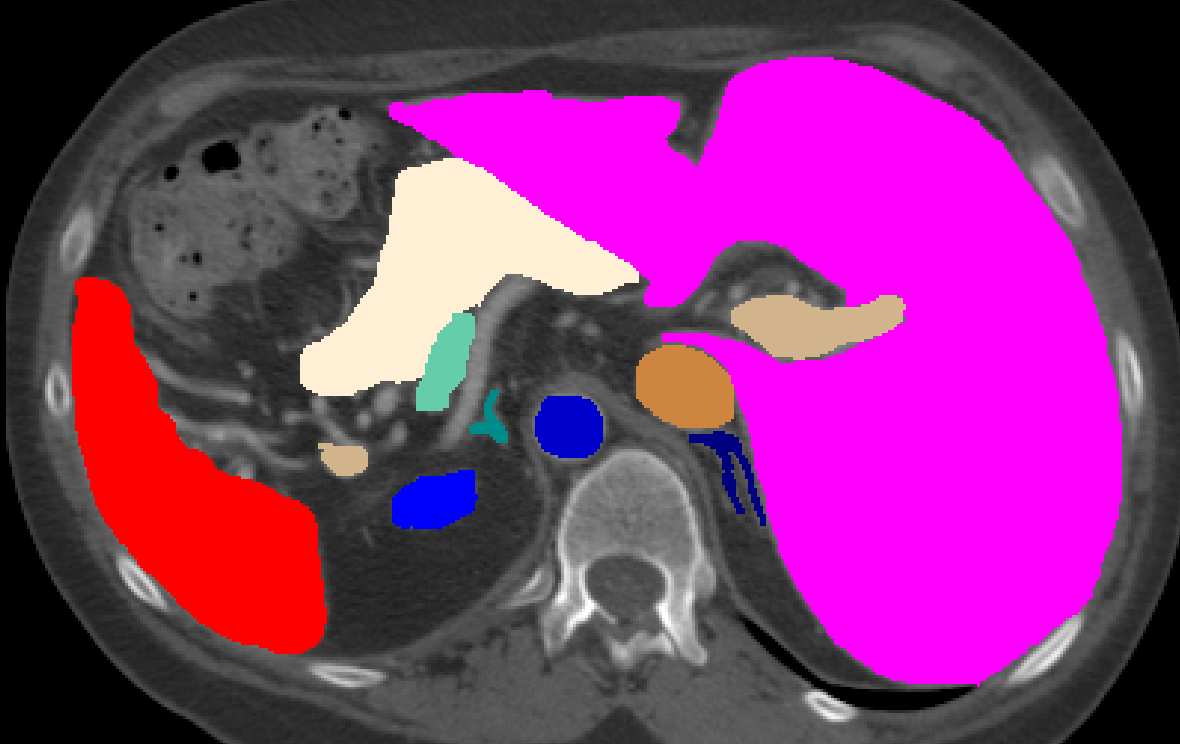}} &\includegraphics[width=0.111\linewidth]{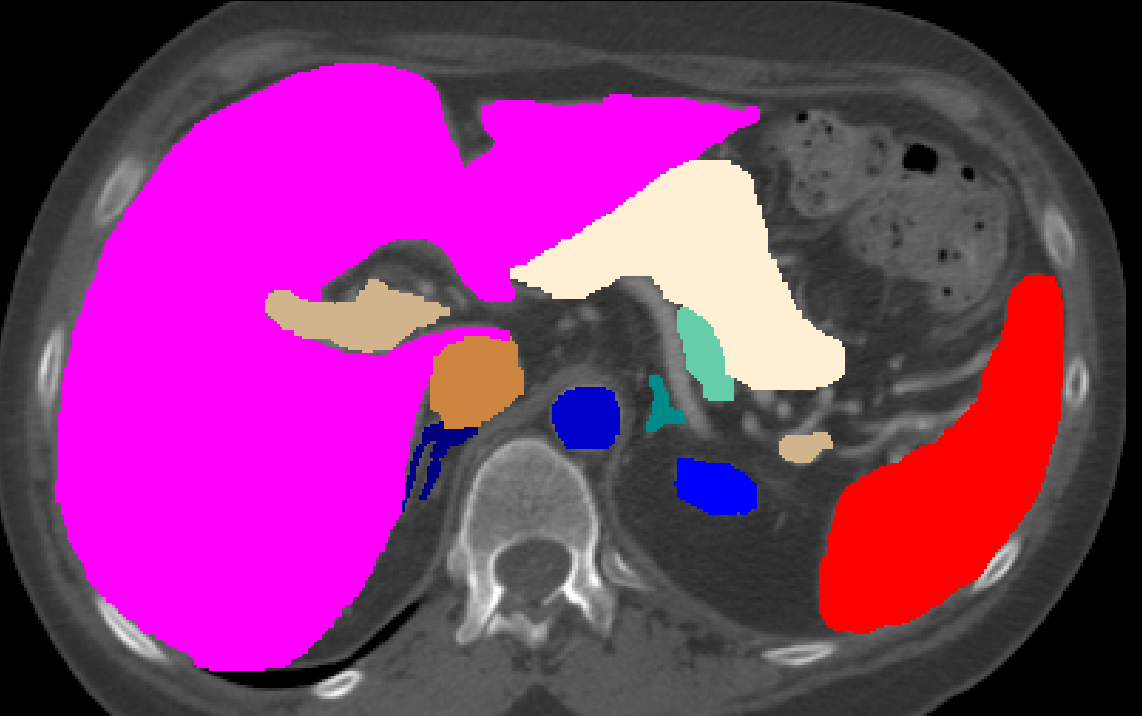}& 
      {\includegraphics[width=0.111\linewidth]{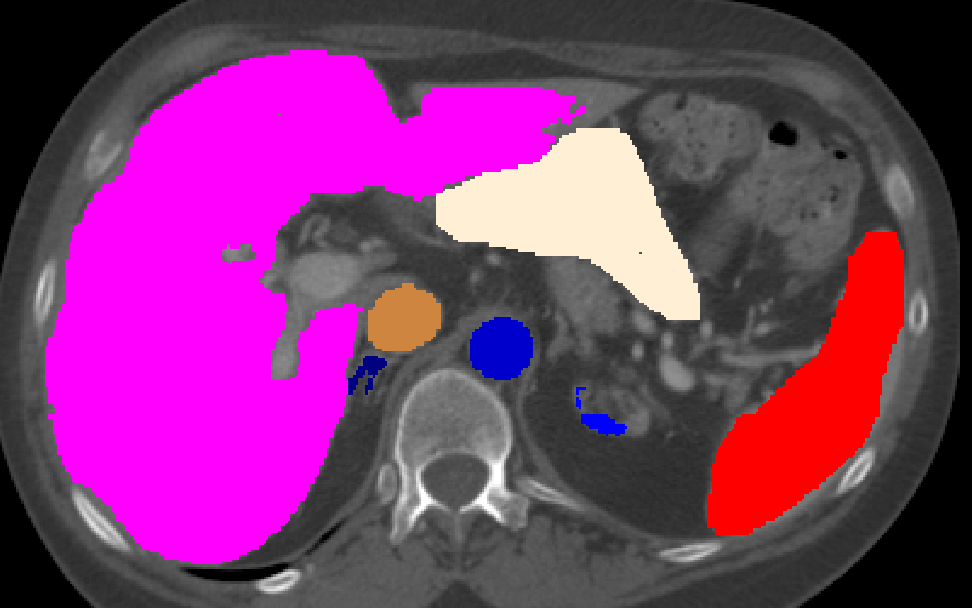}}&
      \includegraphics[width=0.111\linewidth]{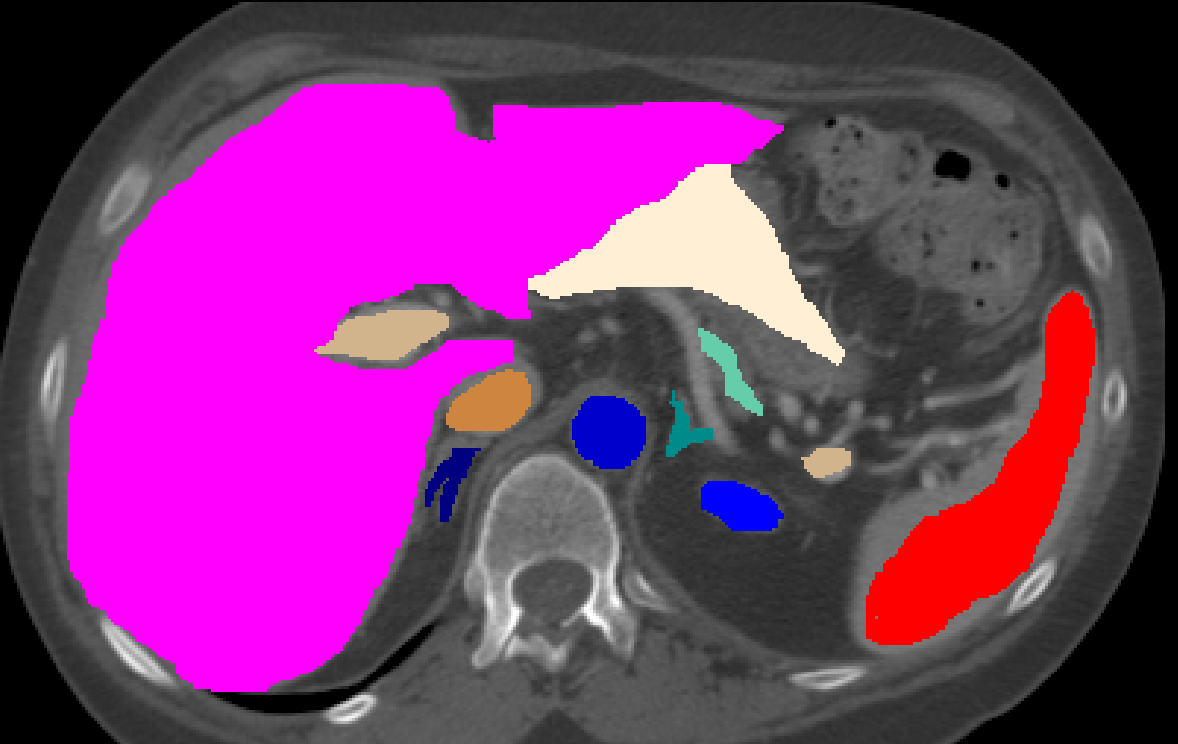}& 
      \includegraphics[width=0.111\linewidth]{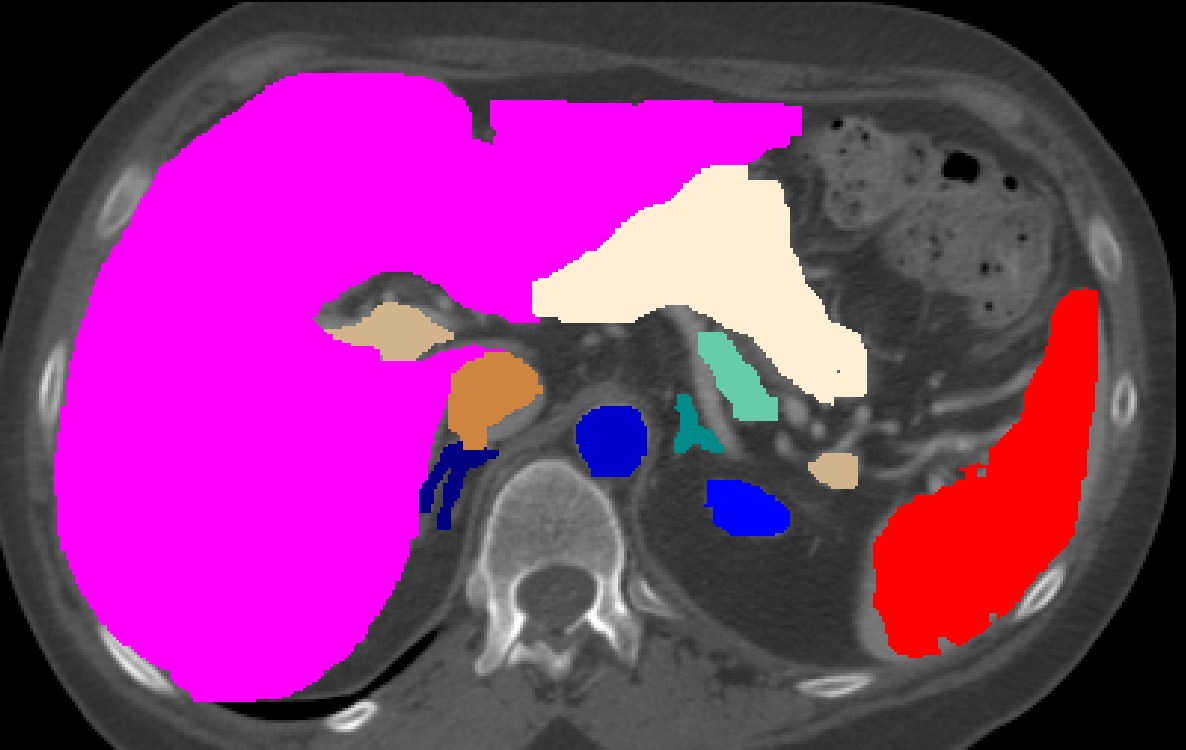} & 
      \includegraphics[width=0.111\linewidth]{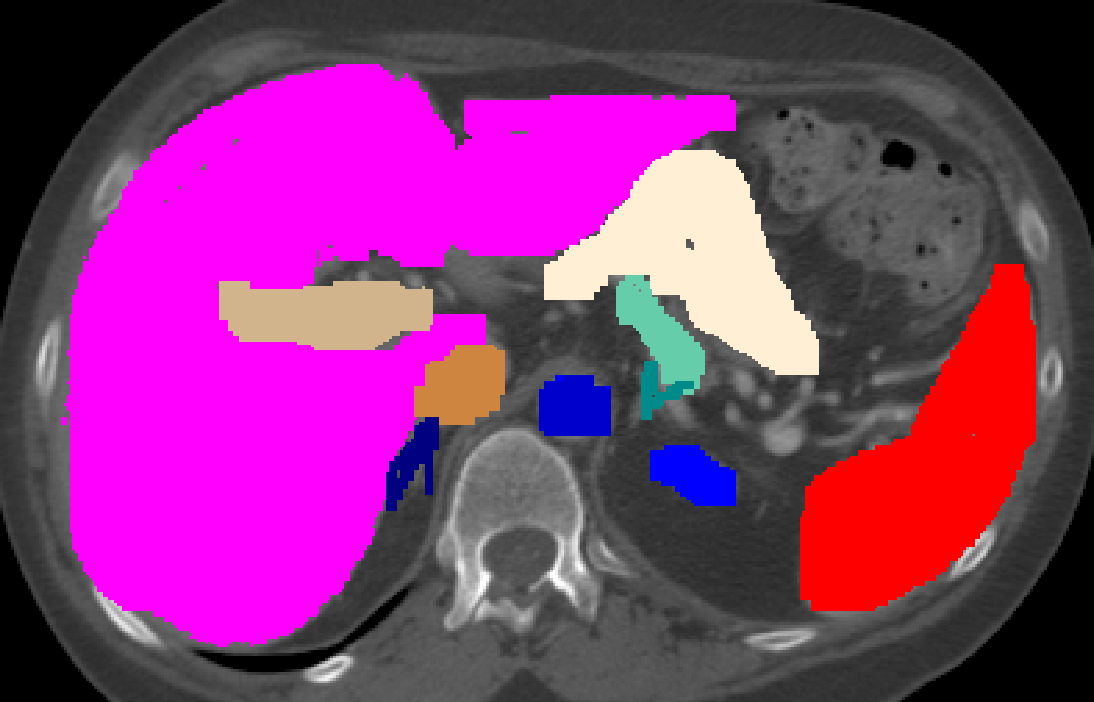} & 
      \includegraphics[width=0.111\linewidth]{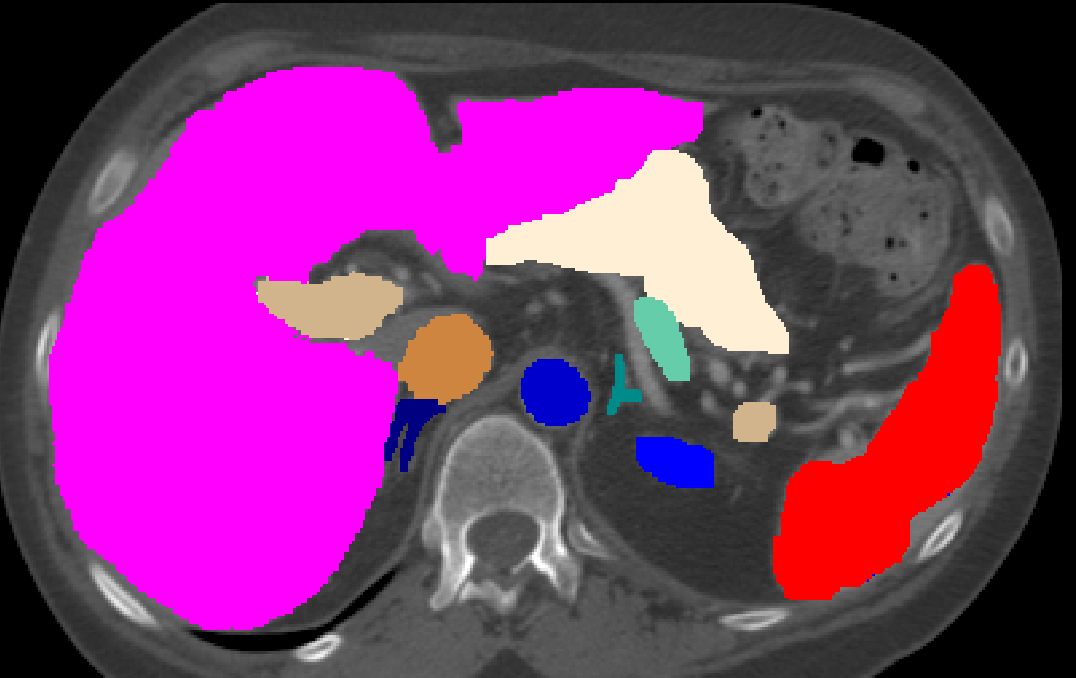}\\[4mm]
      \includegraphics[width=0.111\linewidth]{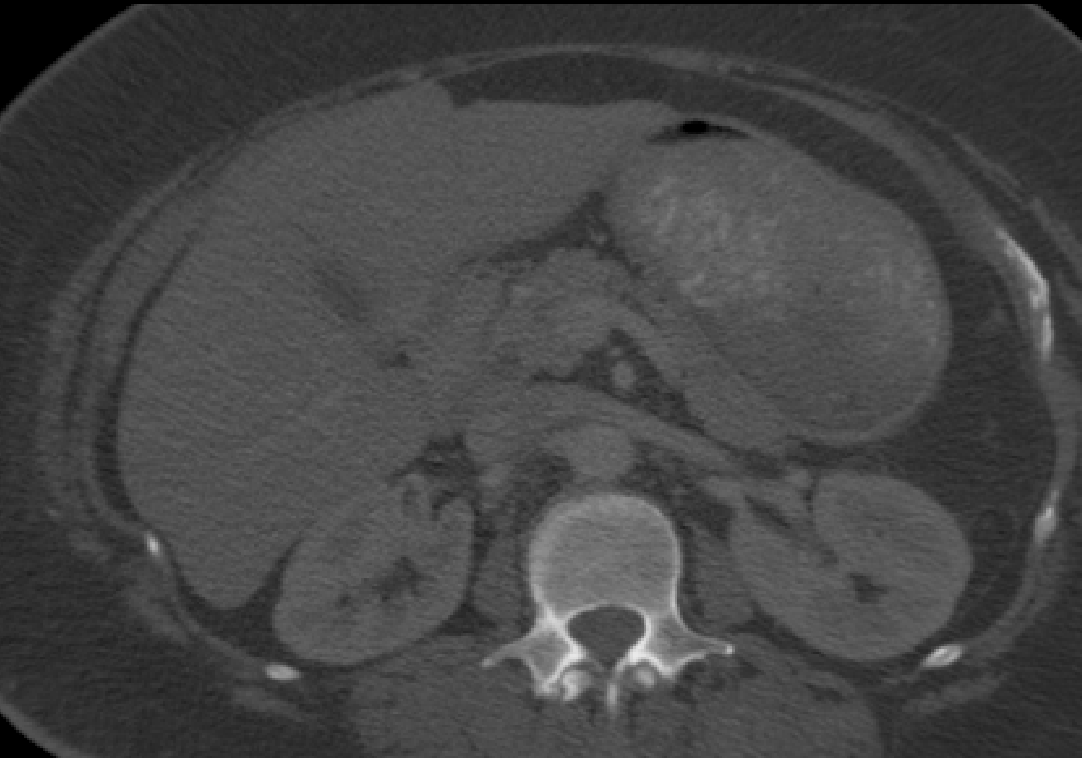} & 
      \includegraphics[width=0.111\linewidth]{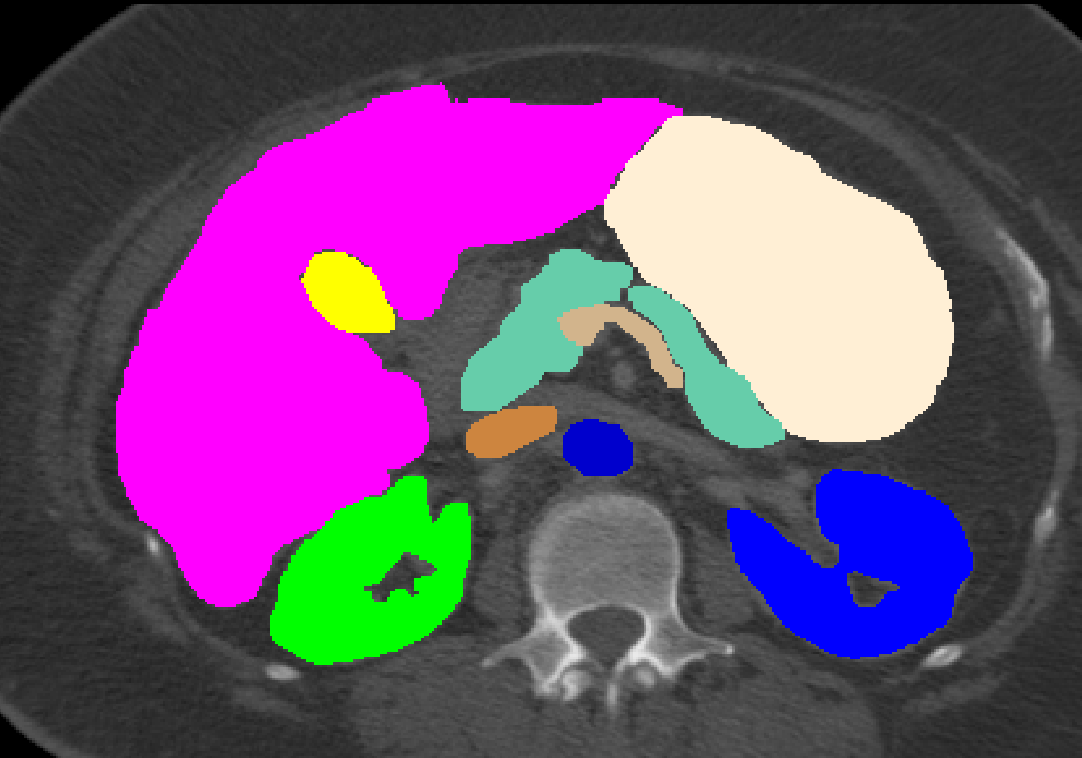} & 
      \includegraphics[width=0.111\linewidth]{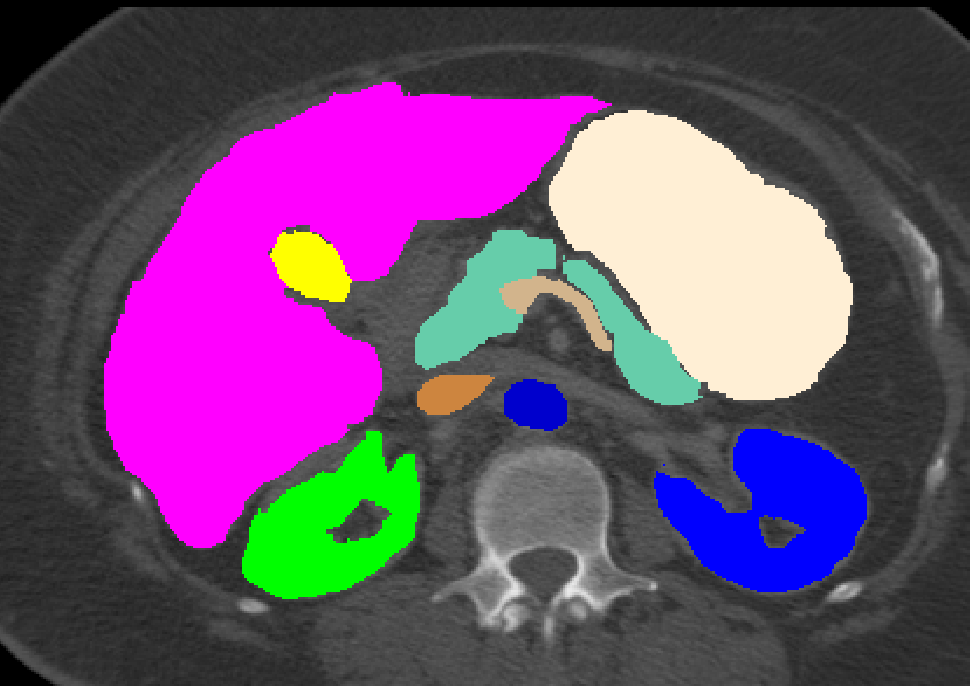} & 
      \includegraphics[width=0.111\linewidth]{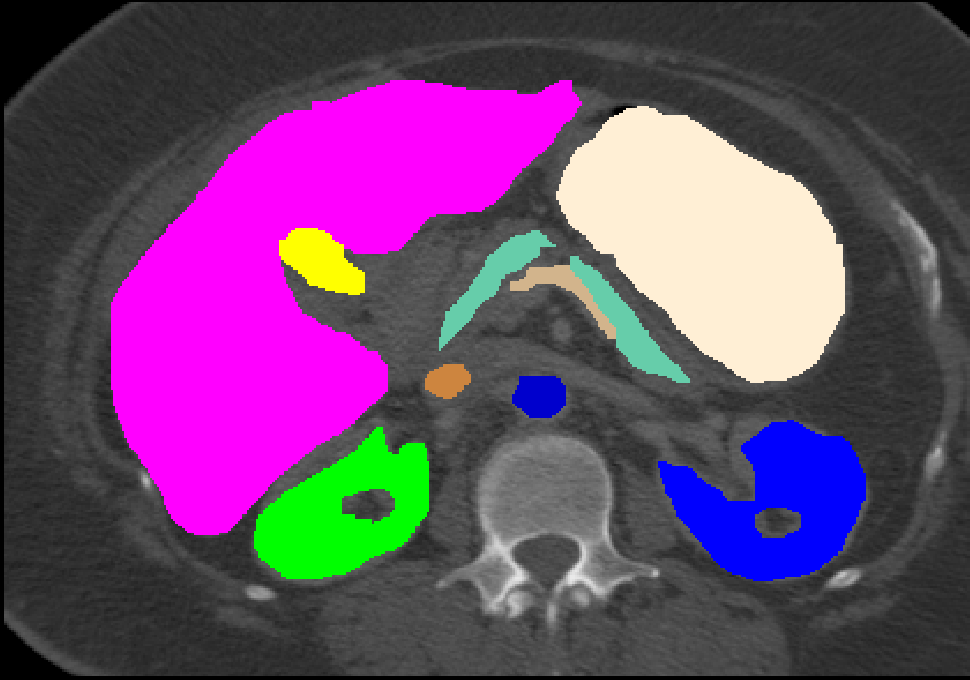} & 
      \includegraphics[width=0.111\linewidth]{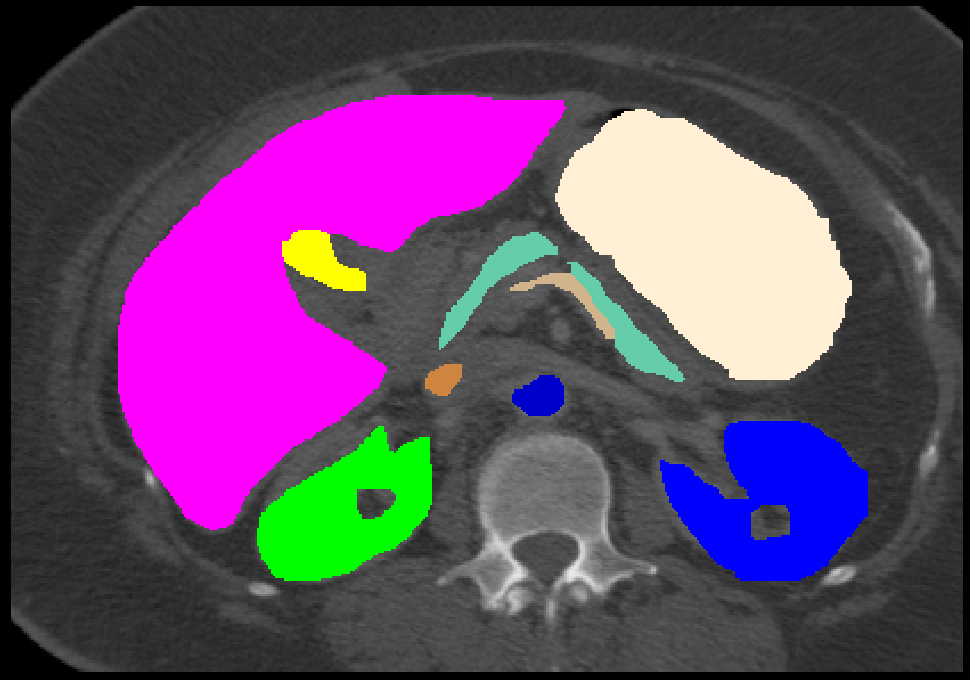} & 
      \includegraphics[width=0.111\linewidth]{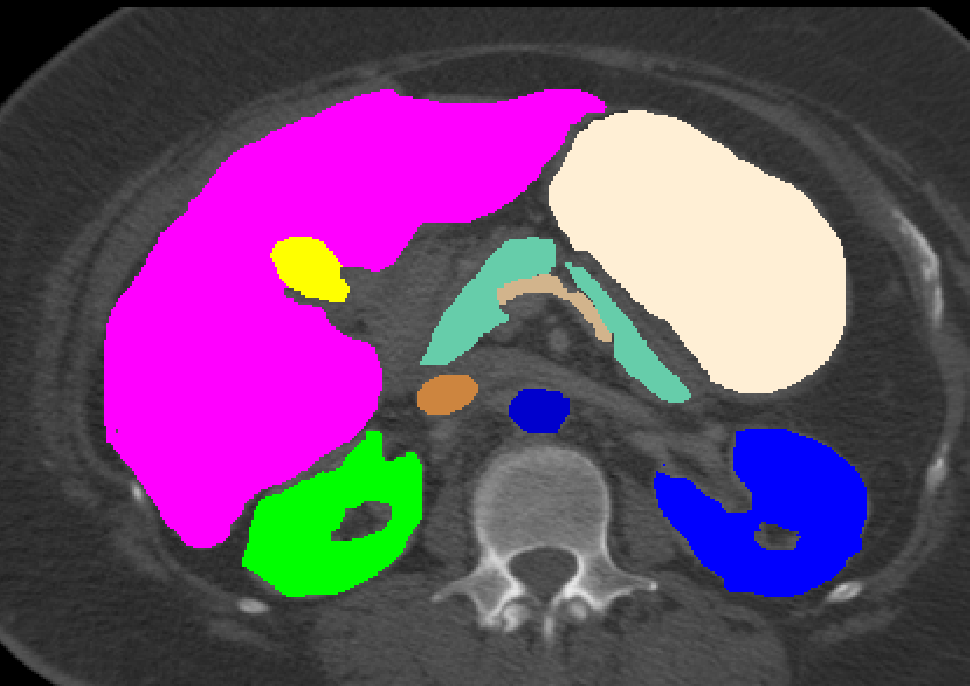} & 
      \includegraphics[width=0.111\linewidth]{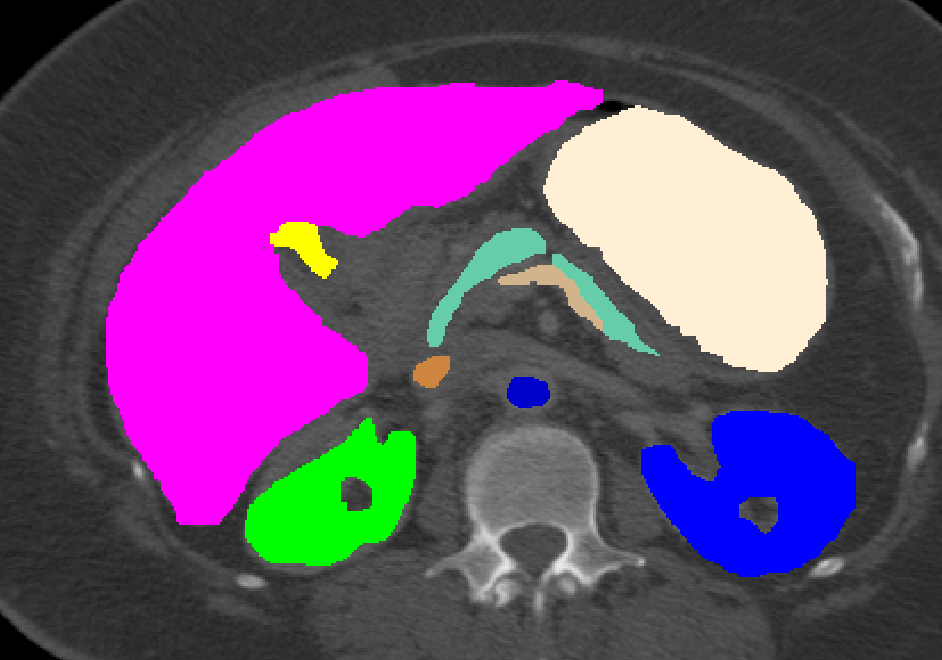} & 
      \includegraphics[width=0.111\linewidth]{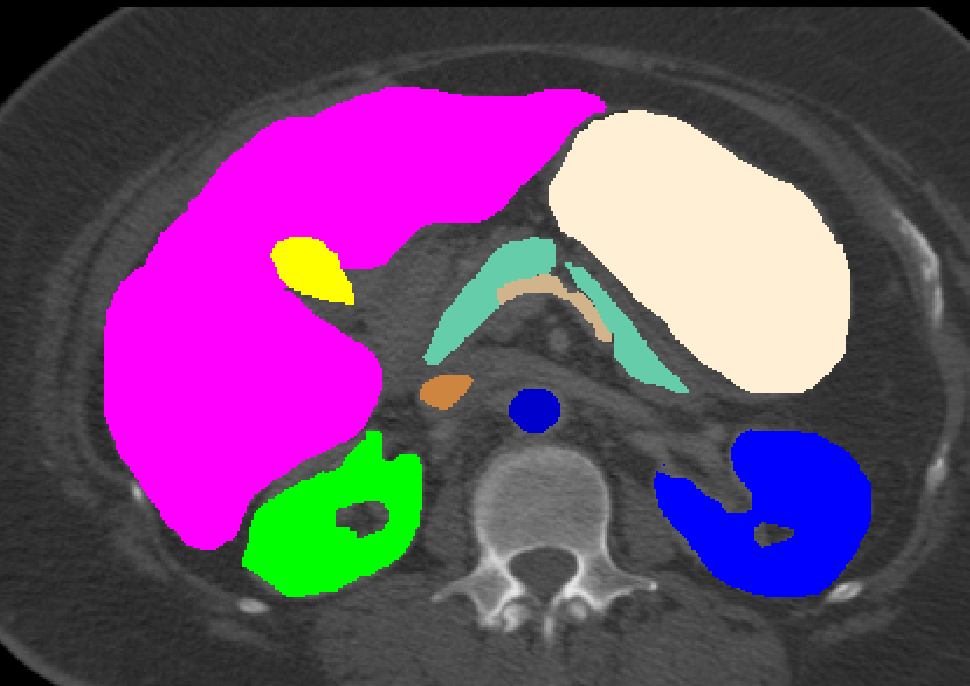} \\
    \multicolumn{8}{c}{
        \includegraphics[page=1, trim=0cm 0cm 0cm 0cm, clip, width=1\linewidth]{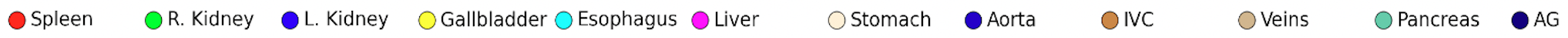}
    }
    \end{tabular}
    
    }
    
\caption{\textbf{Qualitative Results on BTCV.} We show comparisons of 3D segmentation on the abdomen BTCV dataset with 12 organ labels \cite{BTCV}. Note the significant improvement of our Swin-BOB especially on \textit{Stomach} in the first row and \textit{Pancreas} in the second row
}
\label{fig:btcv}
\end{figure*}

\vspace{-2pt}
We train Swin-UNetr \cite{Hatamizadeh2022SwinUS} on our filtered \methodname for a foundation model (Swin-BOB) for 3D segmentation. We evaluate test performance on multiple downstream tasks including BRATS brain MRI benchmark \cite{Baid2021TheRB} in Table \ref{tab:segmentation_comparison_brats} and BTCV Abdomen CT \cite{BTCV} in Table \ref{tab:btcv} (examples in \figLabel{\ref{fig:btcv}}). In both benchmarks, our Swin-BOB achieves \sota with up to 0.02 Dice score improvement and reduction of 2.4 in Mean Hausdorff Distance.  We also establish a \methodname benchmark with reported Dice score and Mean Hausdorff Distance of different networks in Table \ref{tab:segmentation_comparison_ukbb} to aid research in this direction.

\subsection{Entropy Test Time Adaptation Results} 
\vspace{-2pt}
In Table~\ref{table:entropy} we show the benefit of endowing different fine-tuned models with our proposed ETTA and show improvement on 3 different datasets' test performance for 3D segmentation using 3 different networks (including Swin-BOB). In all the 3 networks, we compare the ETTA against augmentation-based test-time adaptation baseline \cite{Zhang2020GeneralizingDL}. This highlights the importance of ETTA in tackling the issue of domain shift in medical imaging especially when the training includes noisy labels, as in the case of the Swin-BOB model.

\begin{table*}[h!]
    \centering
    \resizebox{0.99\linewidth}{!}{%
    \begin{tabular}{cc|cccccccccc|c}
    \hline
    \textbf{Dataset} & \textbf{Config.}                                      
    & \textbf{Spleen} & \textbf{R.Kid} & \textbf{L.Kid} & \textbf{Gall.} & \textbf{Eso.}  & \textbf{Liver} & \textbf{Stom.} & \textbf{IVC}& \textbf{AG} & \textbf{Aorta} & \textbf{Mean}
    \\  \hline
    \multirow{ 4}{*}{\textbf{AMOS}} & TVS & 0.823 & 0.697 & 0.814 & 0.782 & 0.786 & 0.802 & 0.759 & 0.738 & 0.879 & 0.881 & 0.796
    \\ 
         & no filter & 0.908 & 0.931 & 0.942 & 0.657 & 0.658 & 0.958 & 0.822 & 0.874 & 0.529 & 0.906 & 0.818
    \\ 
    & + vol. filter & 0.910 & 0.940 & 0.951 & 0.658 & 0.667 & 0.966 & 0.832 & 0.882 & 0.621 & 0.918 & 0.832 \\
    & \textbf{+ SOLF filter} & 0.919 & 0.943 & 0.962  &0.664  & 0.672 & 0.969 & 0.838 & 0.882 & 0.631 & 0.924 & 0.840
    \\  \hline
        \multirow{ 4}{*}{\textbf{BTCV}} & TVS & 0.848 & 0.721 & 0.801 & 0.785 & 0.797 & 0.795 & 0.737 & 0.715 & 0.861 & 0.862 & 0.792 
    \\ 
     & no filter & 0.883 & 0.884 & 0.932 & 0.795 & 0.790 & 0.946 & 0.885 & 0.871 & 0.784 & 0.799 & 0.856\\
    & + vol. filter & 0.881 & 0.870 & 0.939 & 0.812 & 0.801 & 0.925 & 0.873 & 0.858 & 0.781& 0.852 & 0.875 \\
    & \textbf{+ SOLF filter} & 0.891 & 0.890 & 0.949 &  0.823 & 0.881 & 0.897 & 0.899 & 0.891 & 0.824 & 0.871 & 0.882 \\
    \hline
    \end{tabular}
    }
    \vspace{2pt}
    \caption{\textbf{Detailed Zero-shot 3D Segmentation Performance.} We show Zero-shot Test Dice Score of Swin-BOB on AMOS external MRI data and CT (BTCV) for same organ classes. We show 10 organs that overlap between BTCV and AMOS where we combined left and right adrenal gland into one class named \texttt{AG} while inferior vena cava is briefed as \texttt{IVC}. TotalVibe Segmentator model (TVS) \cite{graf2024totalvibesegmentator} results are shown for reference, while the Swin-BOB model is trained on complete \methodname without filtration, with normalized volume statistical filter, and with full SOLF filter respectively. Note the significant benefit of filtering the collected \methodname labels. 
    }   
    \label{table:zero_shot_segmentation_detailed}
\end{table*}

\begin{table*}[h!]
    \centering
        \resizebox{0.9\linewidth}{!}{%
    \begin{tabular}{l|c|c|c|c|c|c}
        \hline
        & \multicolumn{2}{c|}{\textbf{BTCV}} & \multicolumn{2}{c|}{\textbf{AMOS}} & \multicolumn{2}{c}{\textbf{BRATS}} \\
        \multicolumn{1}{c}{\textbf{Configuration}} & Mean Dice & Mean HD & Mean Dice & Mean HD & Mean Dice & Mean HD \\
        \hline
        nn-UNet \cite{nnUNet}  & 0.804 & 12.141 & 0.795 & 9.623 & 0.812 & 9.787\\
        nn-UNet   + TTA \cite{Zhang2020GeneralizingDL} & 0.811 & 10.901 & 0.830 & 8.465 & 0.832 & 8.327 \\
        nn-UNet   + ETTA (ours) & 0.831 & 8.652 & 0.826 & 7.683 & 0.848 & 7.874 \\
        \hline
        Swin-UNetr \cite{Hatamizadeh2022SwinUS}  & 0.872 & 8.517 & 0.822 & 8.390 & 0.885 & 8.929\\
        Swin-UNetr + TTA \cite{Zhang2020GeneralizingDL} & 0.870 & 8.280 &  0.839 & 7.726 & 0.880 & 8.654\\
        Swin-UNetr + ETTA (ours) & 0.886 & 7.221 & 0.858 & 5.812 & 0.894 & 7.463 \\
        \hline
        Swin-BOB (ours) & 0.883 & 8.261 & 0.847 & 8.105 & 0.882 & 8.624 \\
        Swin-BOB + TTA \cite{Zhang2020GeneralizingDL} & 0.883 & 7.901 & 0.857 & 5.651 & 0.887 & 7.712 \\
        Swin-BOB + ETTA(ours)  & 
        \textbf{0.892} & \textbf{7.381} & \textbf{0.864} & \textbf{7.191} & \textbf{0.894} & \textbf{7.130} \\
        \hline
    \end{tabular}
    }
    \vspace{2pt}
\caption{\textbf{Effect of Entropy Test-Time Adaptation (ETTA).} We demonstrate that the proposed ETTA enhances the performance of fine-tuned models on the BTCV~\cite{BTCV}, BRATS~\cite{Baid2021TheRB}, and AMOS~\cite{AMOS} datasets. The best results are achieved by fine-tuning our baseline Swin-BOB—pre-trained on the \methodname dataset. Our ETTA consistently improves performance across various networks and downstream tasks, outperforming the standard TTA baseline~\cite{Zhang2020GeneralizingDL}.}

    \label{table:entropy}
\end{table*}
\begin{figure}[h]
  \centering
  \includegraphics[width=1\linewidth,trim={1 5 1 5},clip]{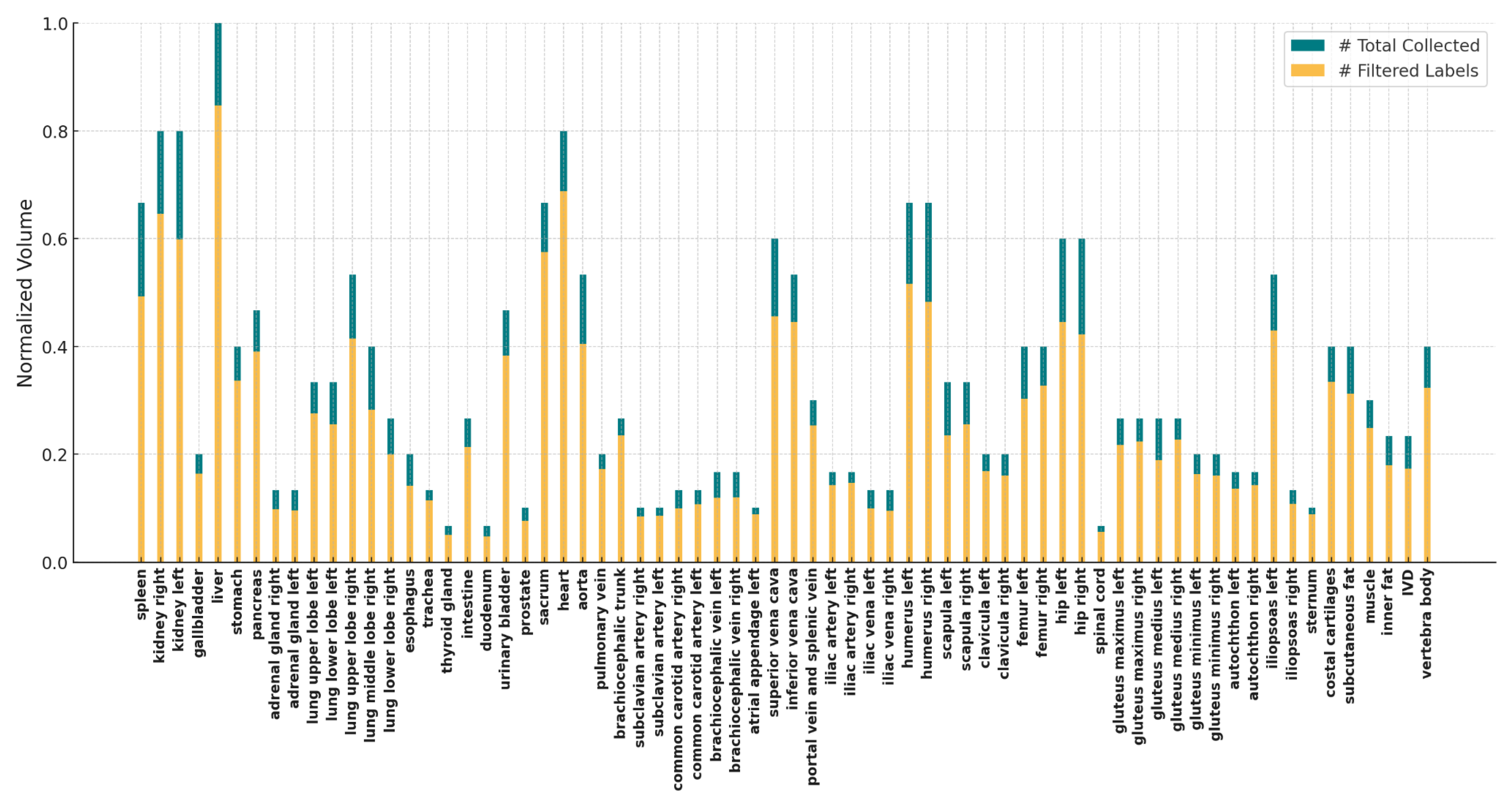}
\caption{\textbf{\methodname Distribution of Labels with our Filtration.} %
We show the distribution of mean normalized volumes of 72 labels before and after SOLF filtration. More examples and classes details are available in \supp.
}
\label{fig:examples1_ukbb}
\end{figure}

\begin{table*}[h]
\centering
\resizebox{\linewidth}{!}{%
\begin{tabular}{r|cccccccccccc|c}
\hline
 \multicolumn{1}{c}{Model}                                                           & Spleen & R.Kid & L.Kid & Gall. & Eso.  & Liver & Stom.  & Aorta & IVC  &Veins & Panc. & AG  & Avg.  \\ \hline
TransUNet \cite{transunet} & 0.952 & 0.927 &  0.929 &0.662 &0.757 &0.969 &0.889 &0.920 &0.833 &0.791 &0.775 &0.637 &0.838 \\
UNetr \cite{UNETR} &0.968 & 0.924 & 0.941 & 0.750 & 0.766 & 0.971 & 0.913 &0.890 & 0.847 & 0.788 & 0.767& 0.741 &0.856 \\
Swin-UNetr \cite{Hatamizadeh2022SwinUS}                                               &0.971 &0.936 &0.943 &0.794 &0.773 &0.975 &0.921 &0.892 &0.853 &0.812 &0.794 &0.765 &0.869\\
nnUNet \cite{nnUNet} & 0.942 &0.894 & 0.910 & 0.704 &0.723 & 0.948 & 0.824 & 0.877 & 0.782 &0.720 & 0.680 & 0.616 & 0.802 \\
Total Vibe Seg. \cite{graf2024totalvibesegmentator} &0.948 & 0.914& 0.917 & 0.736 & 0.741 & 0.954& 0.859 & 0.881 & 0.794 & 0.752 & 0.718 & 0.699 & 0.826 \\
MedSegDiff \cite{MedSegDiff} & 0.973 & 0.930 & 0.955 & 0.812 & 0.815 & 0.973 & 0.924 & 0.907 & 0.868 & 0.825 & 0.788&  0.779 & 0.879 \\
\textbf{Swin-BOB (ours)} & 0.979 & 0.951 & 0.967 & 0.815 & 0.792 & 0.984 & 0.937 & 0.909 & 0.870 & 0.882 & 0.832 & 0.796 & \textbf{0.892}\\
\midrule
MedSegDiff-V2 (ens.)\cite{MedSegDiffV2} & 0.978 &0.941 &0.963 &0.848 &0.818 &0.985 &0.940 &0.928 &0.869 &0.823 &0.831 &0.817 & 0.895\\
\textbf{Swin-BOB (ens.)} &0.981& 0.958& 0.971& 0.817& 0.796& 0.988& 0.942& 0.912& 0.874& 0.886& 0.836& 0.799 & \textbf{0.897}\\
\hline
\end{tabular}}
\vspace{2pt}
\caption{\textbf{3D Segmentation Performance on the BTCV Benchmark.} We evaluate our approach on the 3D segmentation task of the BTCV dataset using the Dice score. For a fair comparison, we also report 10-fold ensembling results (denoted as \textit{ens.}) as presented in MedSegDiff-V2~\cite{MedSegDiffV2}.}

 \label{tab:btcv}
 \end{table*}

\subsection{Analysis and Insights} \label{sec:analysis}
\vspace{-2pt}
\mysection{Filtration Ablation Study}
We study the effect of the filtration threshold $\epsilon$ in the SOLF filter (\eqLabel{\ref{eq:filteration}} of the three features) on the zero-shot generalization of the models trained on the filtered subsets of \methodname. For $\epsilon$ = 0, 1, 2, 3,4, and 5, the performance Dice score on BTCV is 0.792, 0.884, 0.892, 0.766, and 0.745, respectively. We also ablate the features used in the SOLF filter. In Table \ref{table:zero_shot_segmentation_detailed}, when only the normalized volume is used in the SOLF filter(no Sphericity or Eccentricity), the quality of the filtration degrades considerabley, highlighting the importance of each aspect of the SOLF filter to clean the labels.

\begin{figure}[t]
    \centering
    \resizebox{1\linewidth}{!}{
        \begin{tabular}{ccccc}
            \includegraphics[angle=0,trim= 0cm 0cm 0cm 0cm,clip, width=0.5\linewidth]{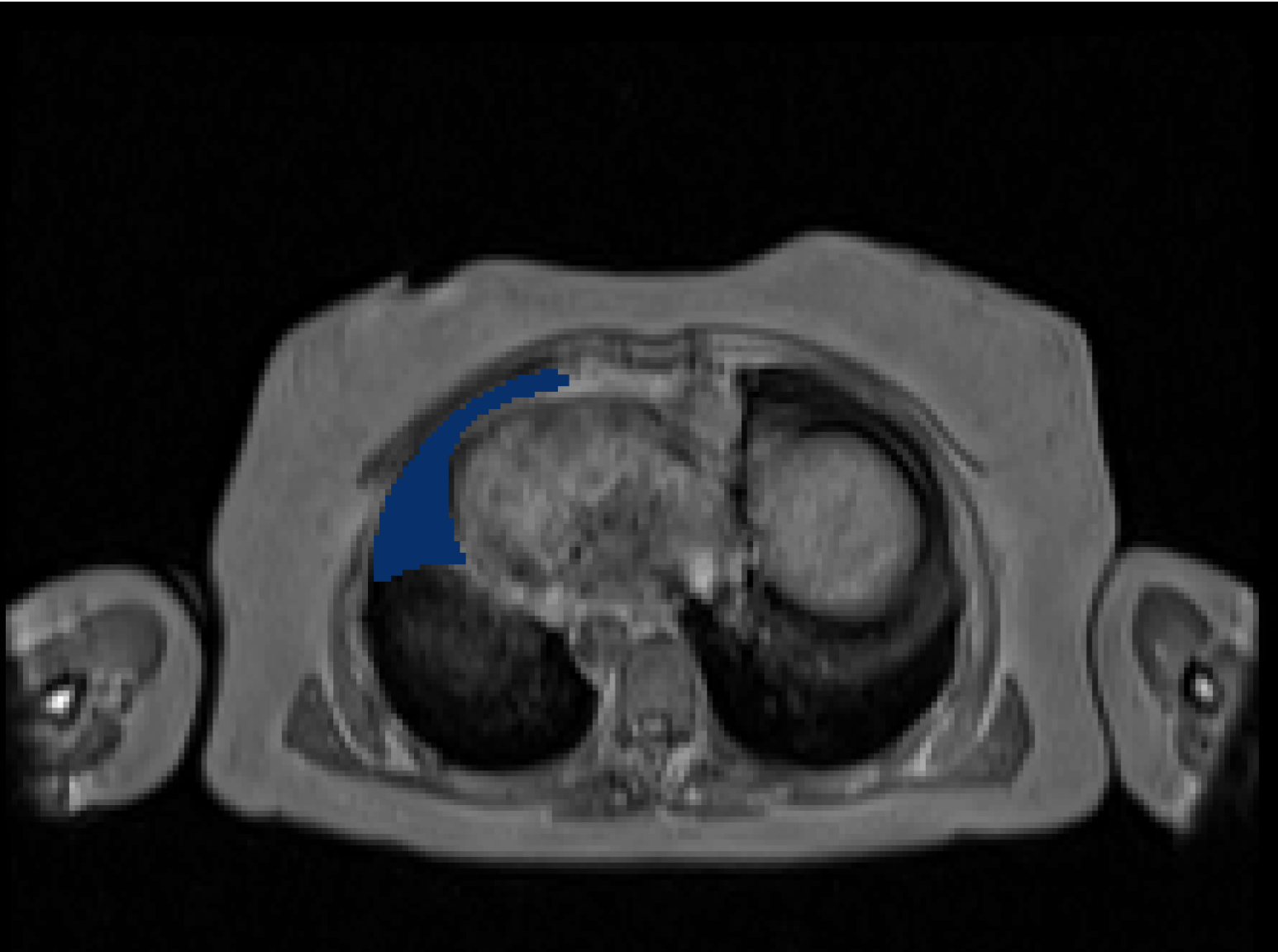}& \includegraphics[angle=0,trim=0cm 0cm 0cm 0cm,clip, width=0.5\linewidth]{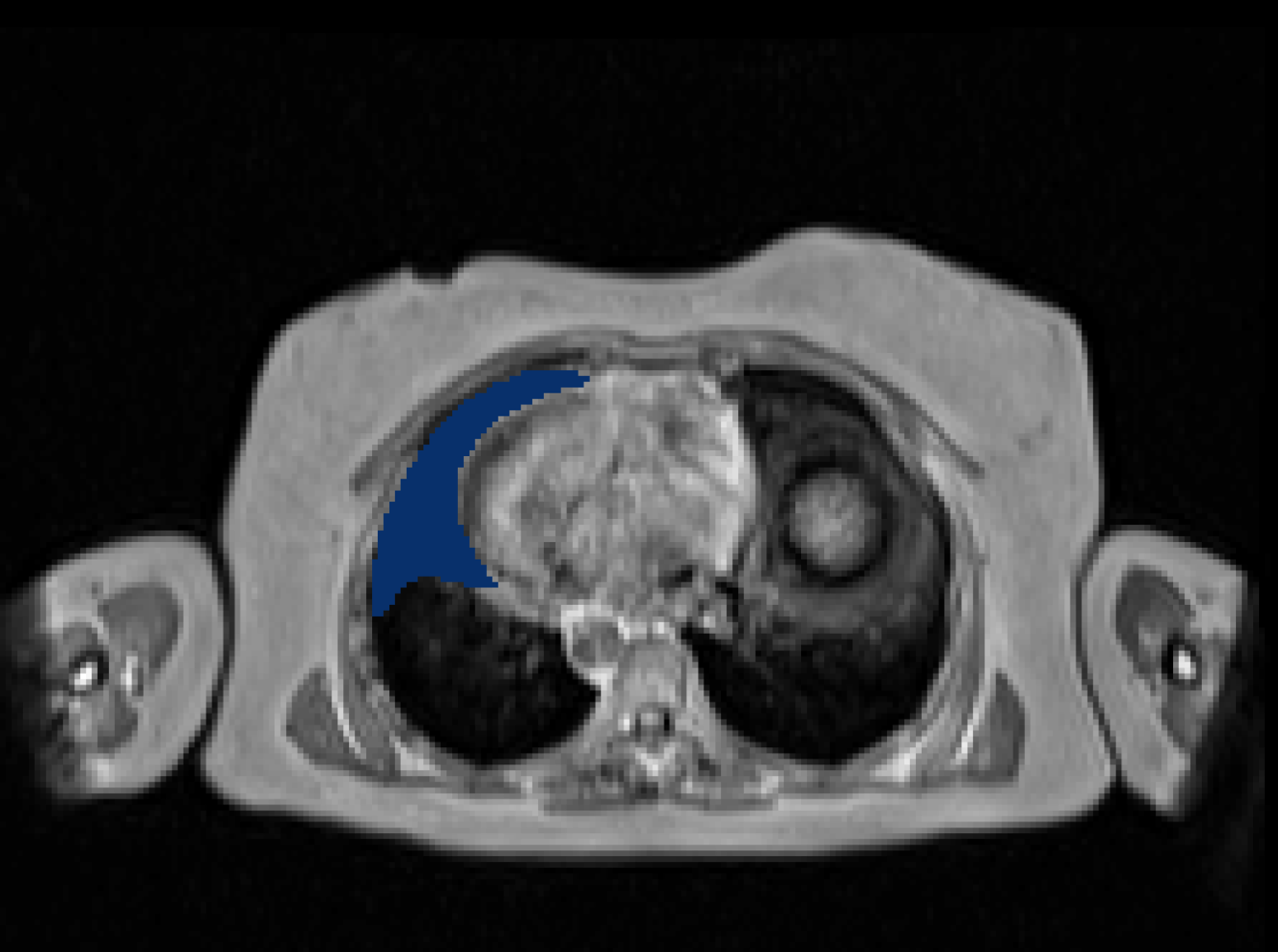}&
            \includegraphics[angle=0,trim= 0cm 0cm 0cm 0cm,clip, width=0.5\linewidth]{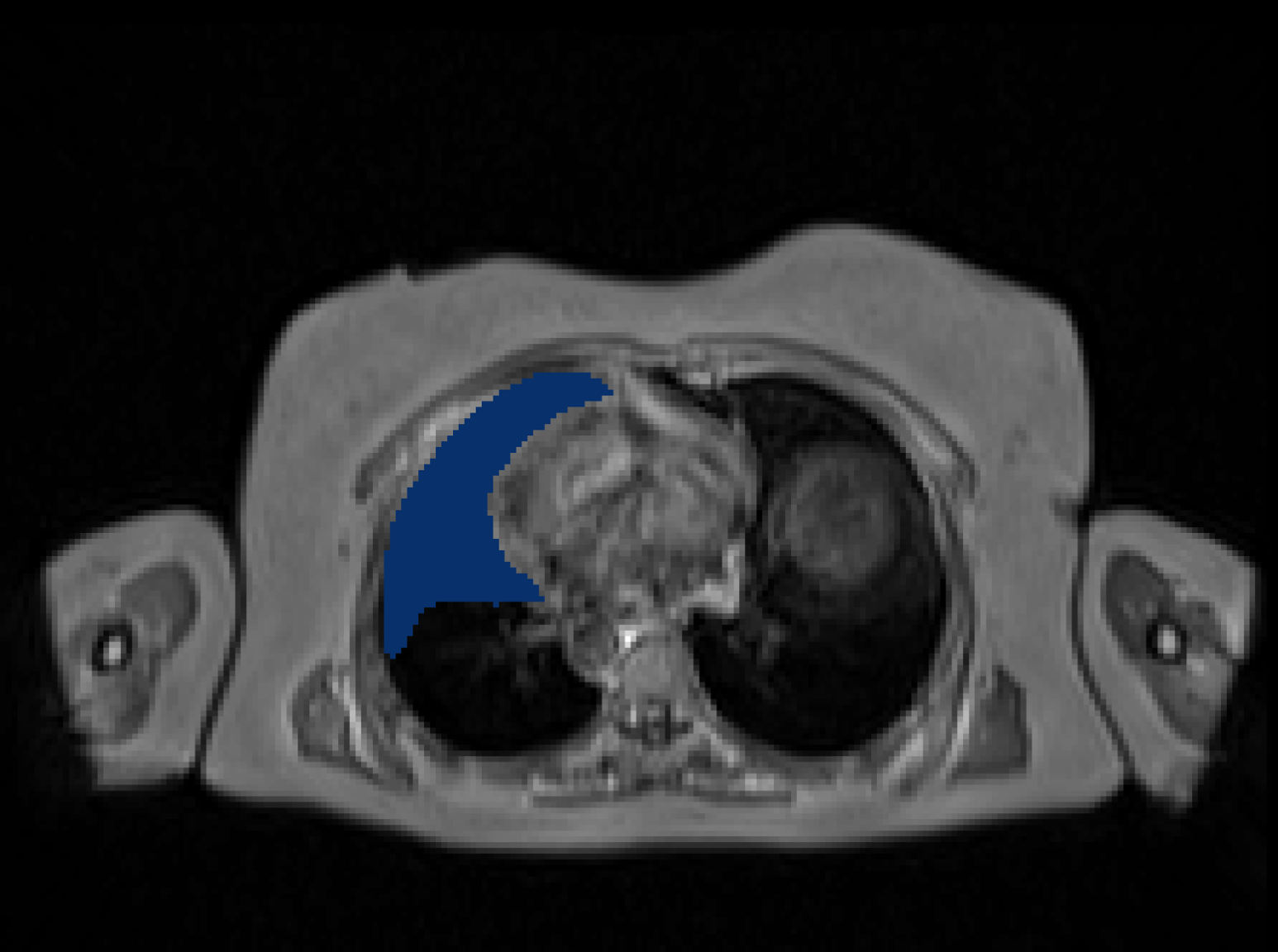}
             & \includegraphics[angle=0,trim=0cm 0cm 0cm 0cm,clip, width=0.5\linewidth]{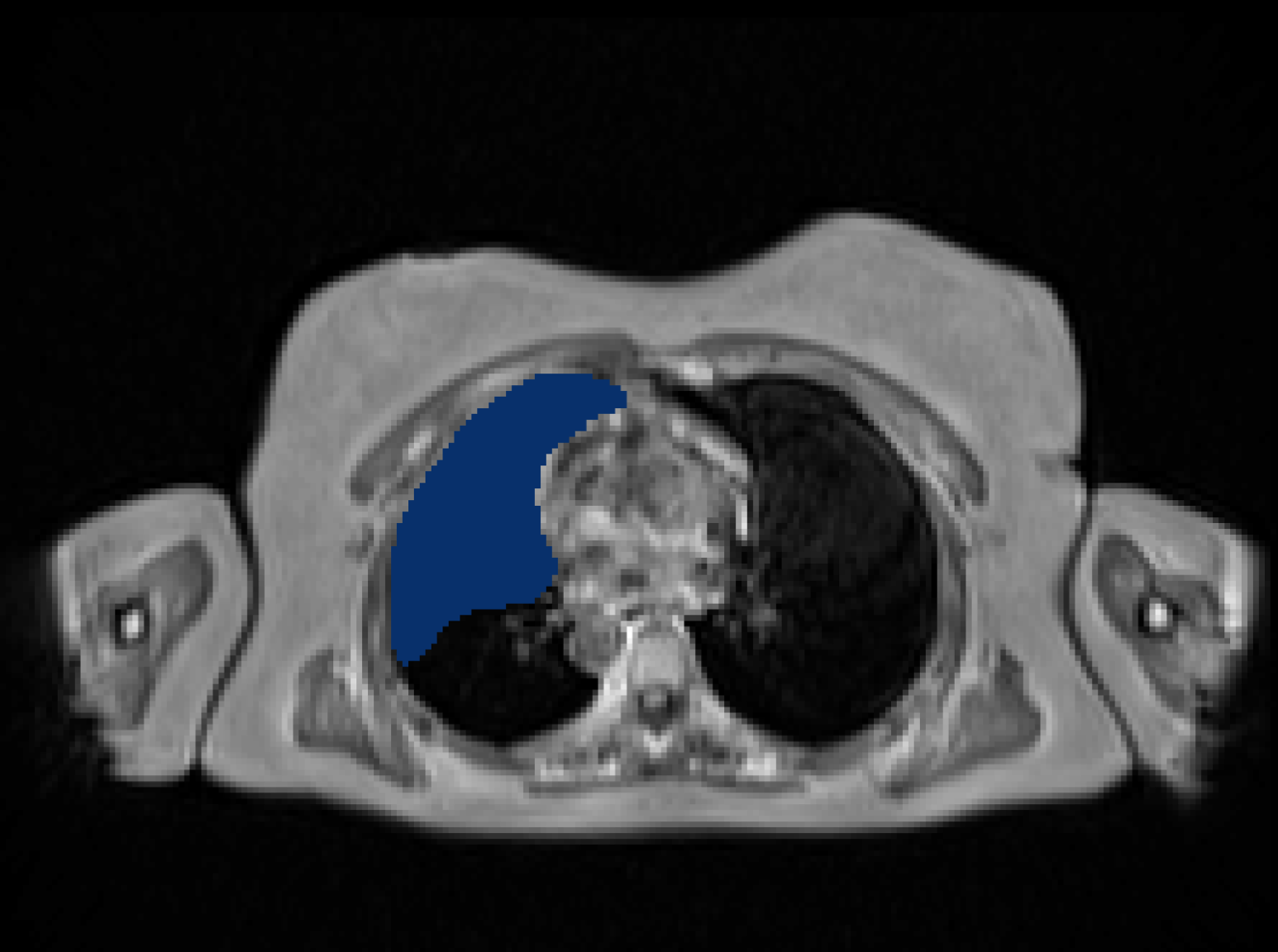}
            & 
           \scalebox{-1}[1]
           {\includegraphics[angle=0,trim=2cm 5cm 2cm 2cm,clip, width=0.5\linewidth]{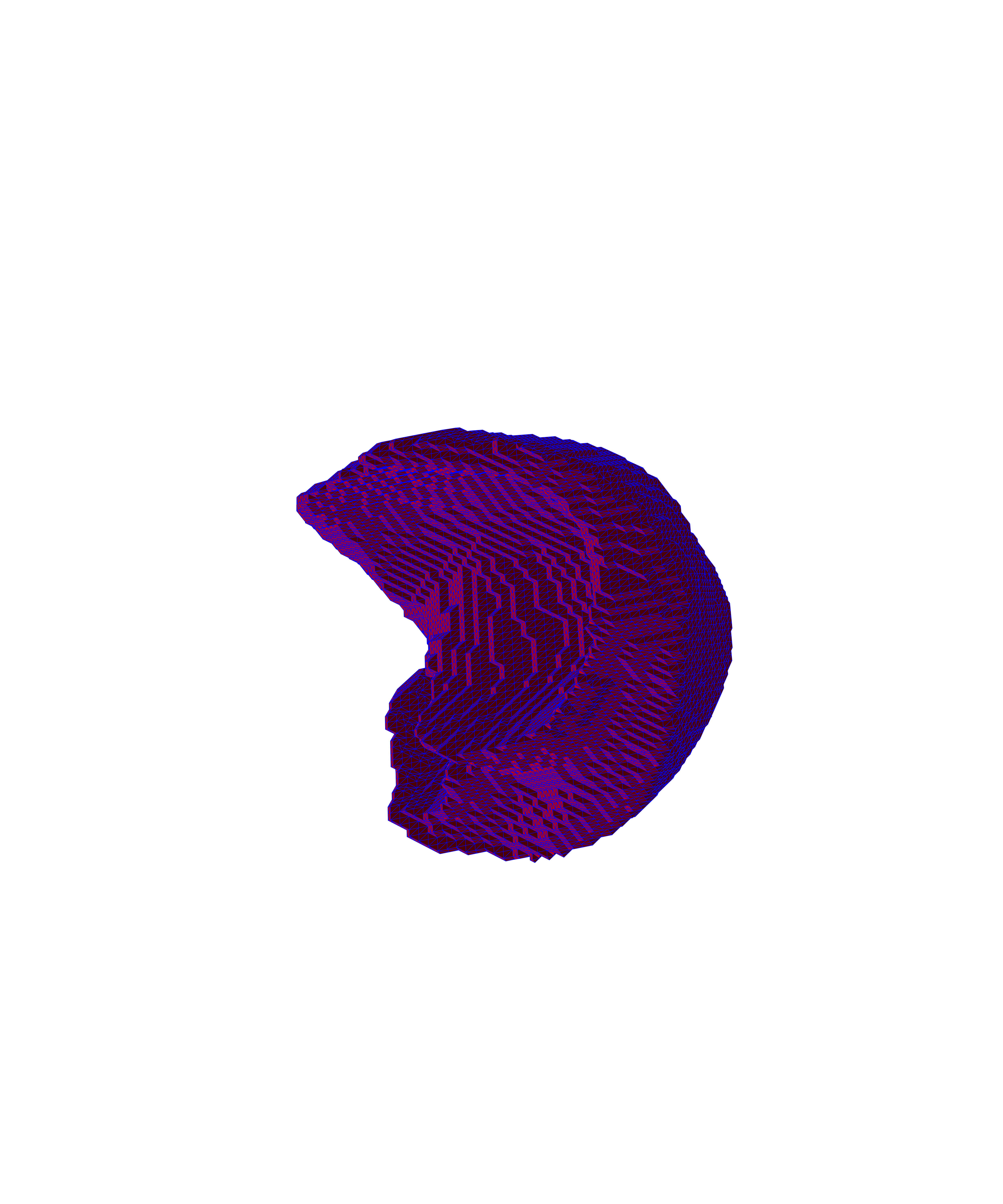}}
            \\
            \includegraphics[angle=0,trim= 0cm 0cm 0cm 0cm,clip, width=0.5\linewidth]{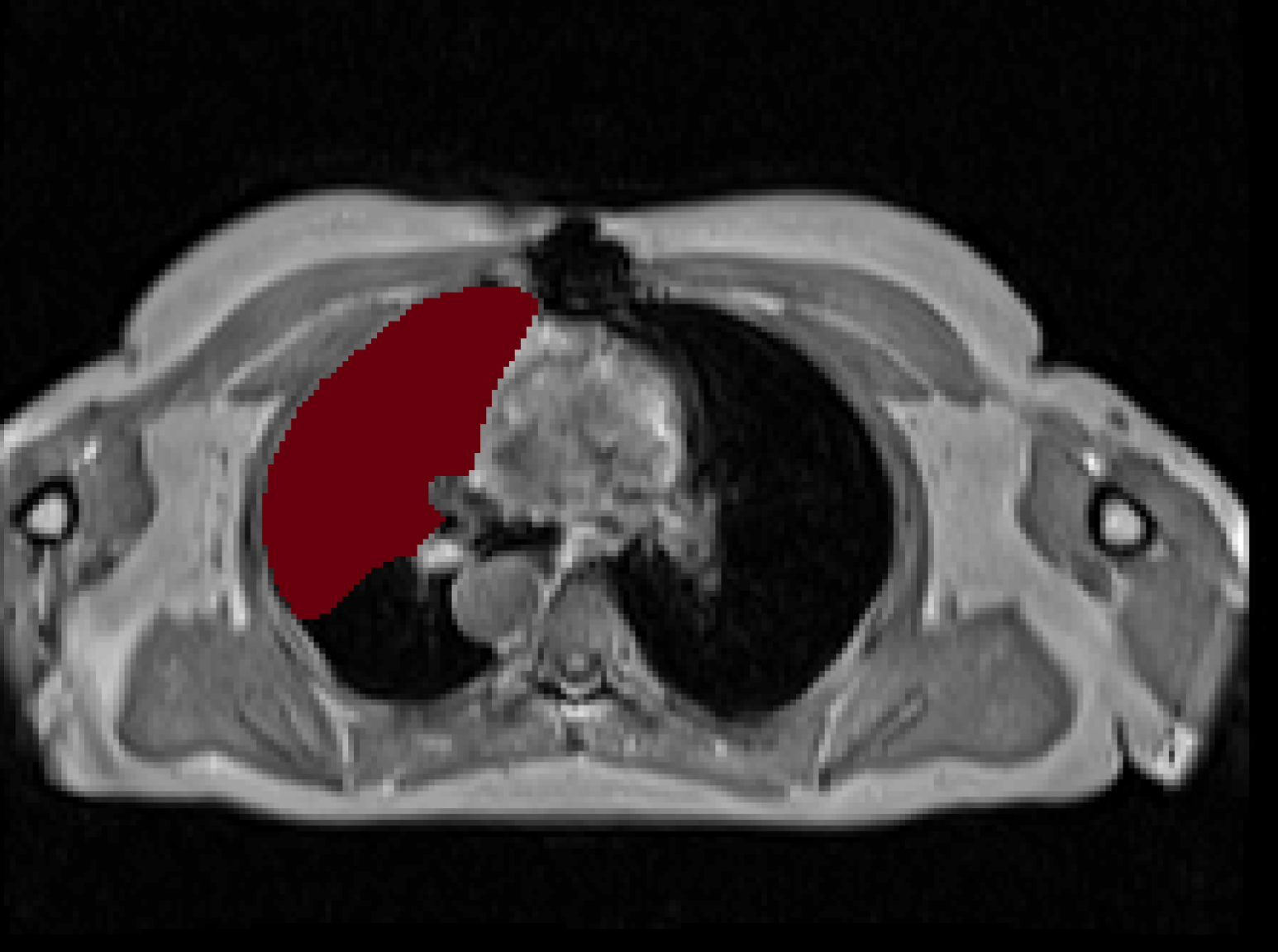}
            & \includegraphics[angle=0,trim=0cm 0cm 0cm 0cm,clip, width=0.5\linewidth]{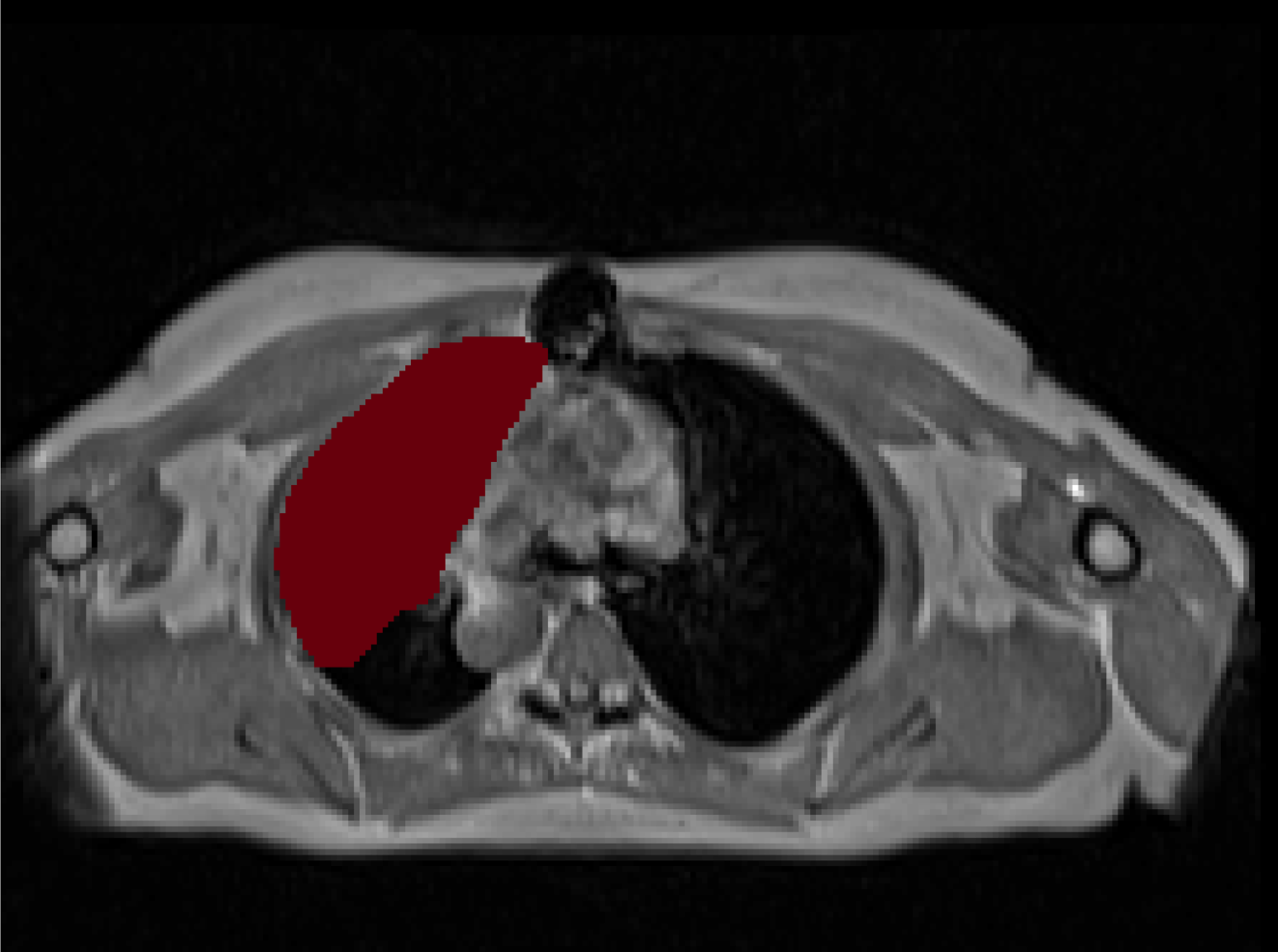}
            &\includegraphics[angle=0,trim=0cm 0cm 0cm 0cm,clip, width=0.5\linewidth]{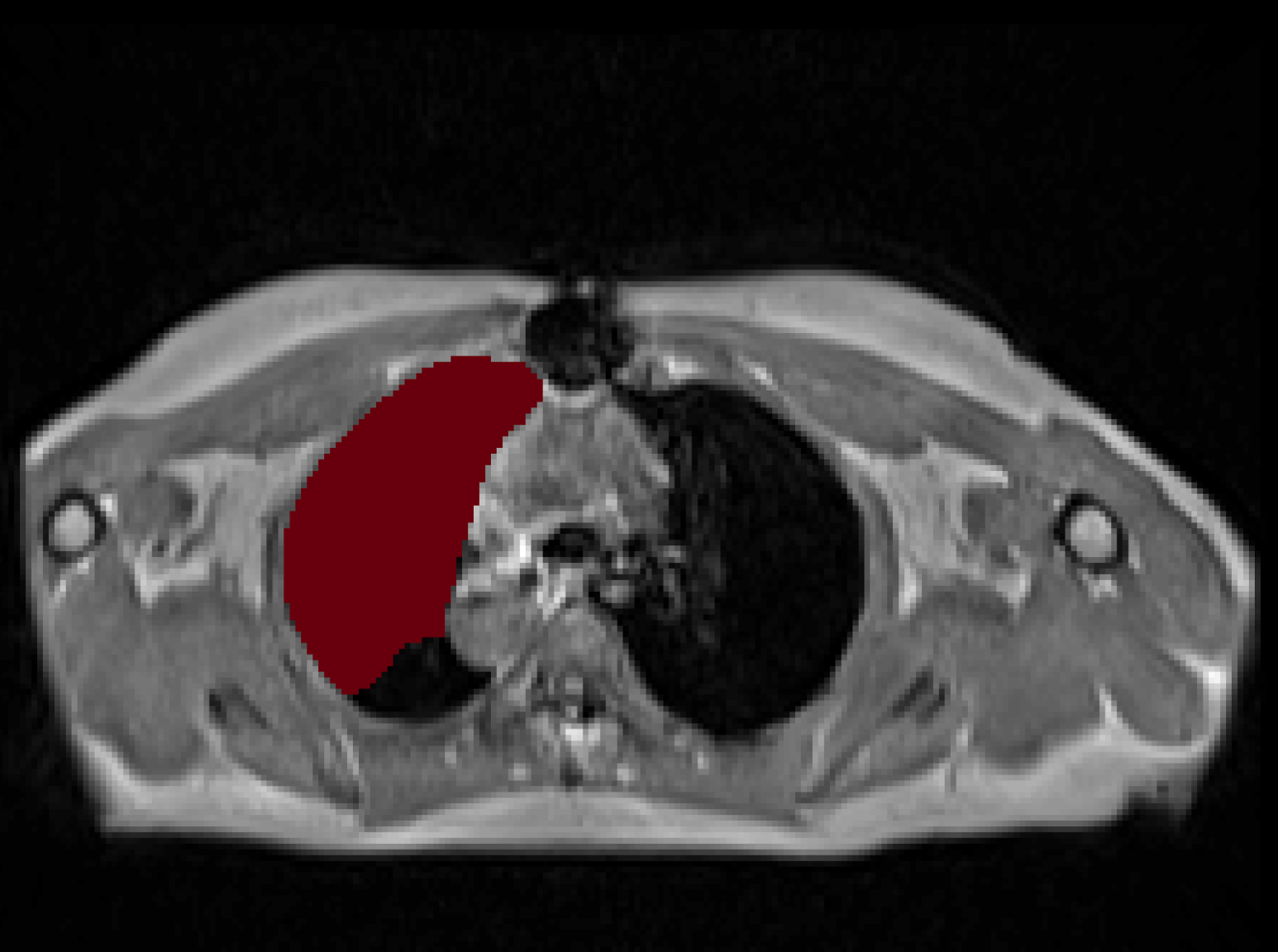}
            &\includegraphics[angle=0,trim=0cm 0cm 0cm 0cm,clip, width=0.5\linewidth]{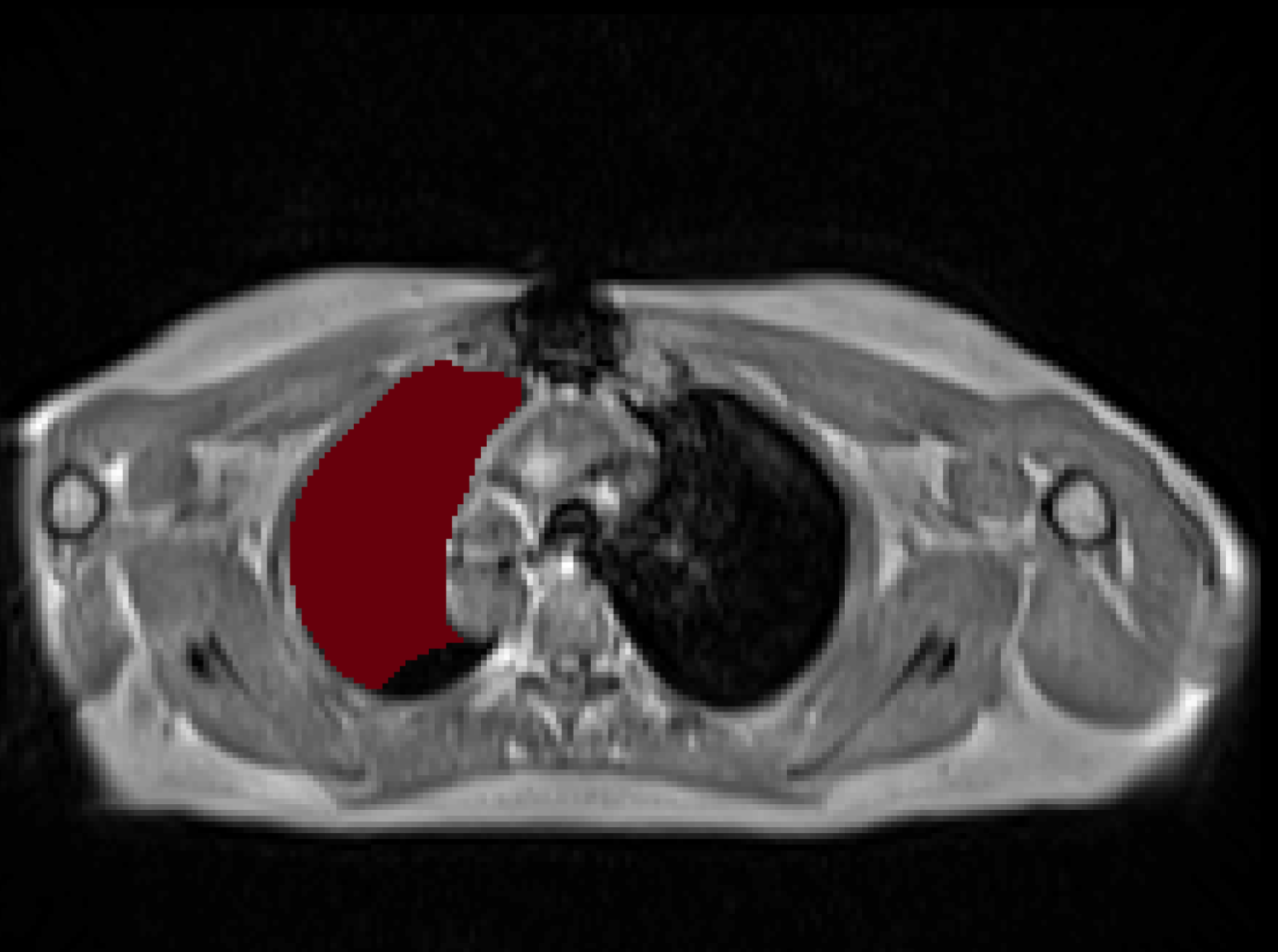}
            &\includegraphics[angle=0,trim=2cm 5cm 2cm 2cm,clip, width=0.7\linewidth]{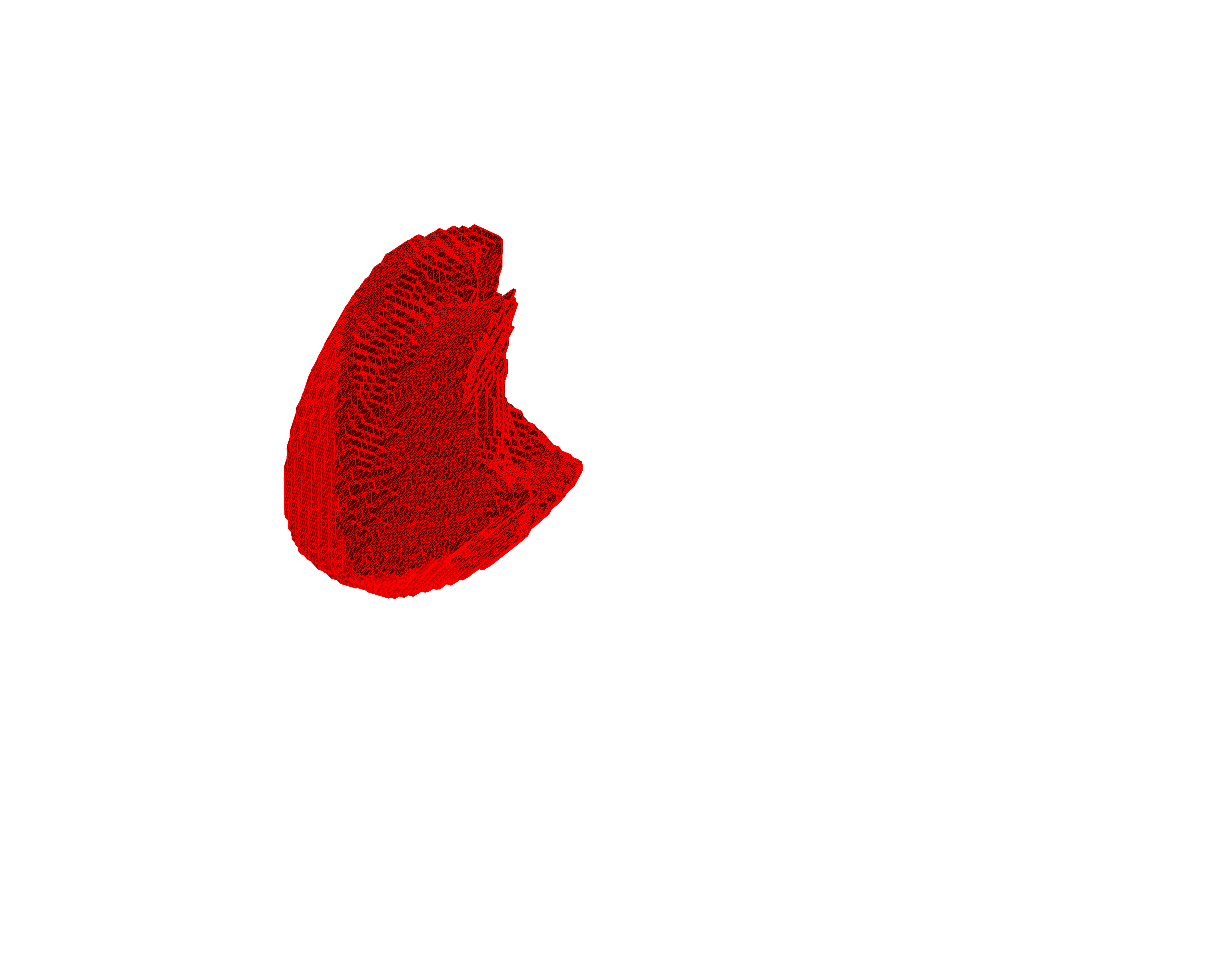}
        \end{tabular} }
    \caption{\textbf{Filtration of Inaccurate Labels.} \textit{top} are manual labels of upper left lung overlaid on scans for several slice indices. \textit{bottom} are filtered out labels (red) of the upper left lung overlaid on scans with corresponding slice indices. The filtered out lung is incomplete and erroneous. %
    }
    \label{fig:filtering}
\end{figure}

\mysection{Filtering Out Patients Abnormalities}
One concern of automatic filtration in \secLabel{\ref{sec:filtering}} is that it might filter out some natural abnormalities or pathologies in the patients, mistaken as wrong labels. We visualize some of these filtered-out labels in Figure \ref{fig:filtering} and show that indeed lack quality labels rather than the patients have obvious abnormalities. 
The combination of normalized volume, sphericity and eccentricity makes the filtration mostly about the quality of the labels rather than filtering out patients with abnormality. \figLabel{\ref{fig:examples1_ukbb}} shows the distribution of organs normalized volumes in \methodname before and after filtration.

\begin{figure}[t]
  \centering
  \includegraphics[trim={0cm 0 0cm 0cm},clip, width=0.99\linewidth]{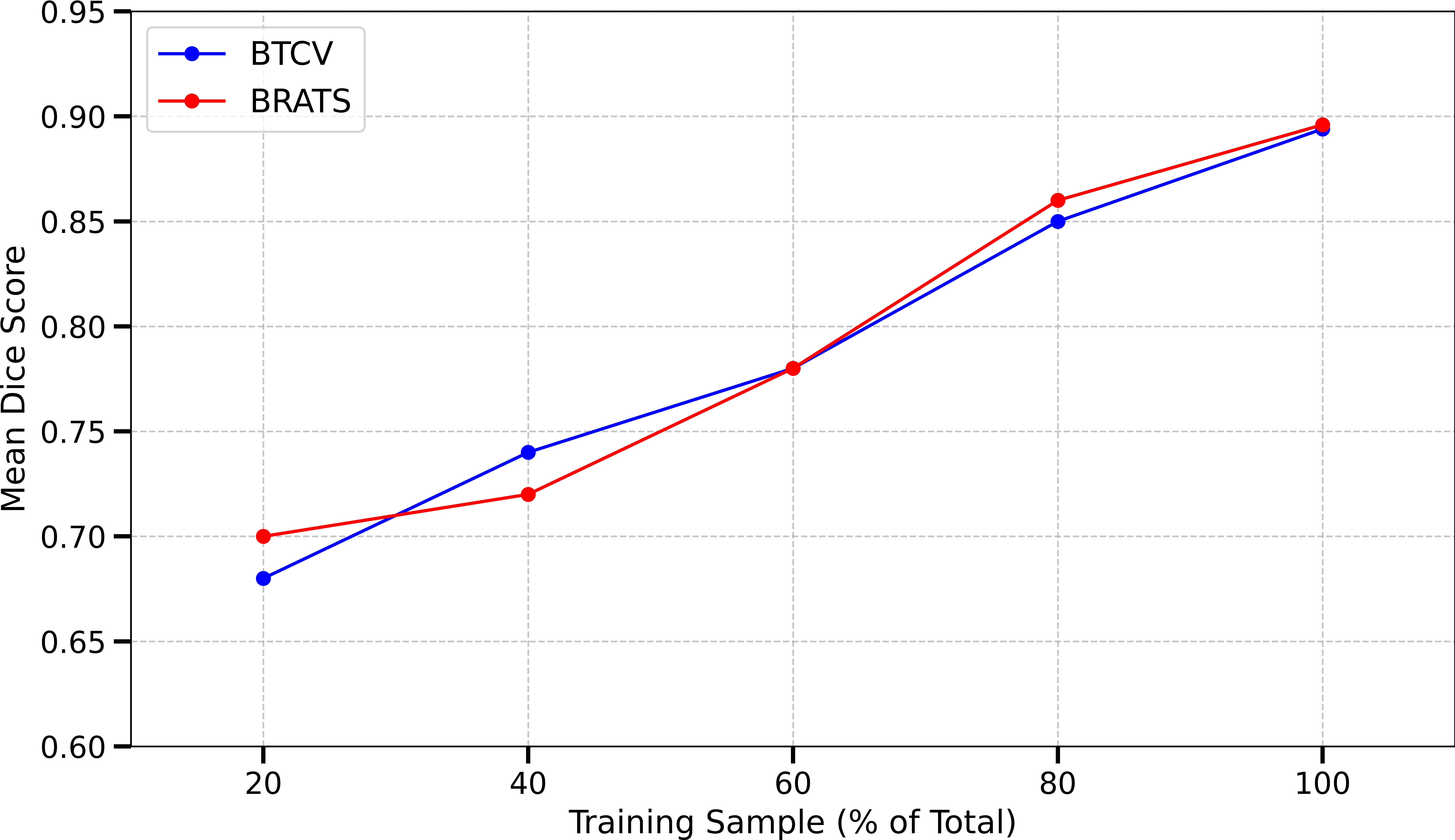}
\caption{\textbf{Effect of \methodname Pre-Training Dataset Size on Downstream Segmentation Performance.} We observe a consistent increase in test Dice Score for both BRATS \cite{Baid2021TheRB} and BTCV \cite{BTCV} when doubling the size of pre-training Swin-BOB on \methodname, acting as a foundation model. 
 }
\label{fig:training_sizde}
\end{figure}

\begin{figure}[t]
  \centering
  \includegraphics[trim={0cm 0 0cm 0cm},clip, width=0.8\linewidth]{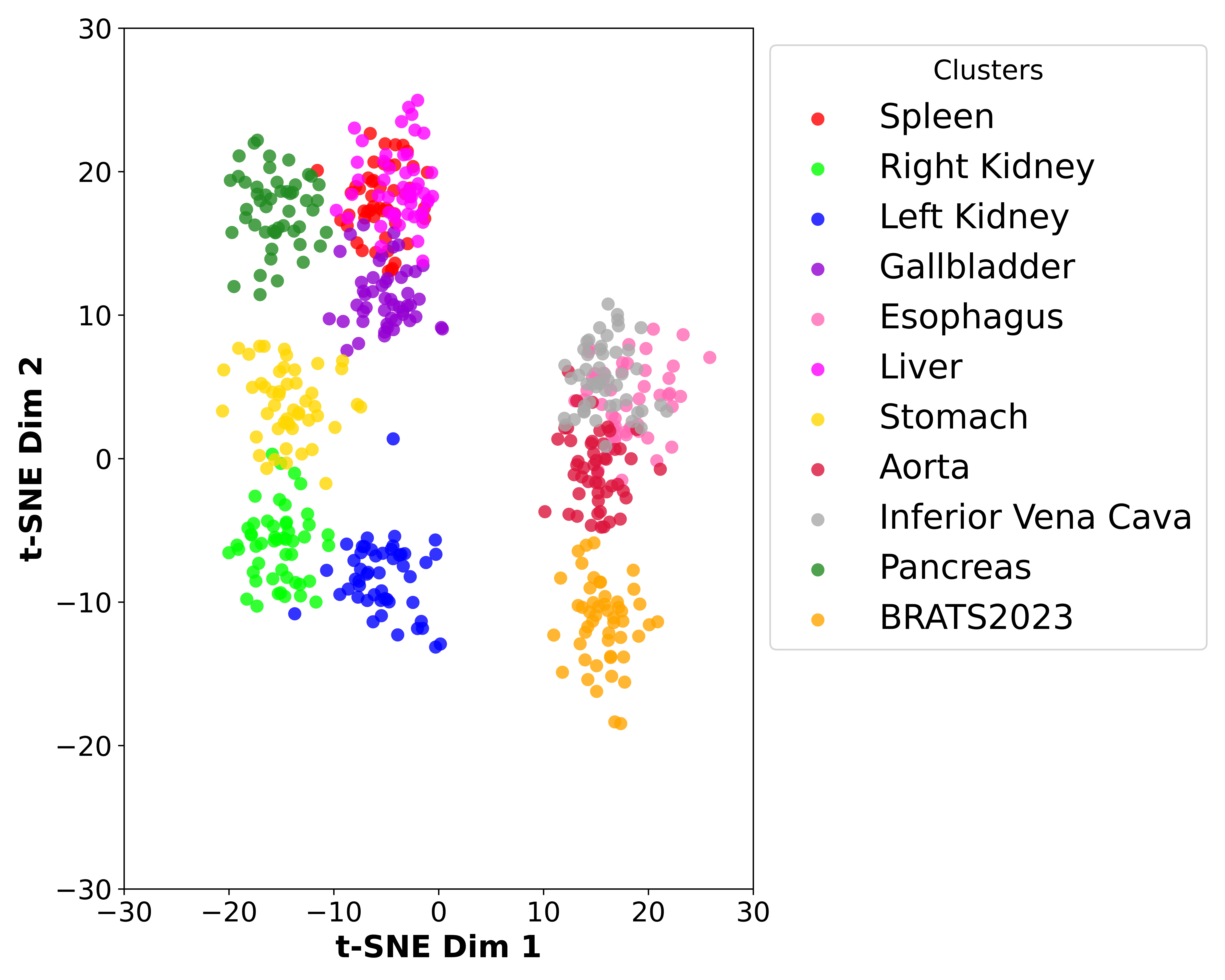}
\caption{\textbf{Distribution of Feature Embeddings on BTCV organs and BRATS23}. Each category is represented with a unique color. We reduce features embeddings to 2D for each class using t-sne \cite{tsne}. The low dispersion of the clusters between each other indicates that the features of different classes probably share similar patterns and this explains the beneficial effect of large pre-training. }

\label{fig:tsne}
\end{figure}

\mysection{Scaling-Law of 3D Medical Segmentation}
We study the impact of the scale of training data of \methodname on the downstream 3D segmentation performance, to justify the large scale \methodname. We independently trained the Swin-UNetr network on subsets of \methodname (\ie 10\%, 20\%, 40\%, 60\%, and 80\% and show in \figLabel{\ref{fig:training_sizde}} the Average test Dice Score for both BTCV \cite{BTCV} and BRATS \cite{Baid2021TheRB}. We also show the t-sne visualization of the features in \figLabel{\ref{fig:tsne}} illustrating the quality of the features. 

\section{Conclusions and Future Works} \label{sec:conclusion}
\vspace{-4pt}
In this work, we introduced the \methodname\ dataset, the largest labeled medical imaging dataset to date, comprising $51,761$ 3D MRI scans and over 1.37 billion 2D segmentation masks covering $72$ organs. 
Our models trained on \methodname\ demonstrate strong zero-shot generalization to other medical imaging datasets and achieve state-of-the-art performance on several benchmarks in 3D medical image segmentation.

\mysection{Limitations and Future Works} 
While \methodname\ significantly expands the availability of large-scale labeled data for medical imaging, it is limited to neck-to-knee MRI scans and may not encompass the full diversity of imaging modalities and anatomical regions. Despite our filtration process, the automatic labeling may still introduce residual label noise that could impact model training. Future work includes extending the dataset to cover additional imaging modalities such as CT scans and incorporating more anatomical regions. Additionally, exploring advanced adaptation techniques and integrating clinical metadata could enhance model robustness and applicability across diverse clinical settings.

\mysection{Acknowledgments}
 This work was supported by the Centre for Doctoral Training in Sustainable Approaches to Biomedical Science: Responsible and Reproducible Research (SABS: R3), University of Oxford (EP/S024093/1), and by the EPSRC Programme Grant Visual AI (EP/T025872/1). We are also grateful for the support from the Novartis-BDI Collaboration for AI in Medicine. Part of the support is also coming from KAUST Ibn Rushd Postdoc Fellowship program.

{\small
\bibliographystyle{ieee_fullname}
\bibliography{egbib}
}
\clearpage
\appendix
\renewcommand{\thesection}{\Alph{section}}
\renewcommand{\thetable}{\Roman{table}}
\renewcommand{\thefigure}{\Roman{figure}}

\setcounter{section}{0}
\setcounter{table}{0}
\setcounter{figure}{0}

\section{Detailed Setup} \label{secsup:setup}
\subsection{Datasets} \label{supsec:datasets}

We conducted our experiments on four primary datasets:
    \mysection{UK Biobank} A more comprehensive dataset of 51,761 full-body MRIs from more than 50,000 volunteers\cite{Sudlow2015UKBA}, capturing diverse physiological attributes across a broad demographic spectrum. UK Biobank MRIs are resampled to be isotropic and cropped to a consistent resolution (501 $\times$ 160 $\times$ 224). 
    
    \mysection{BRATS} The largest public dataset of brain tumours consisting of 5,880 MRI scans from 1,470 brain diffuse glioma patients, and corresponding annotations of tumours\cite{Baid2021TheRB,4b589b6824a64a2a91e8e3b26cc0bf9e,41847efe8ced40078c67adce2164d865}. All scans were skull-stripped and resampled to 1 mm isotropic resolution. All images have resolution 240 $\times$ 240 $\times$ 155. Tumours are annotated by expert clinicians for three classes: Whole Tumour (WT), Tumour Core (TC), and Enhanced Tumour Core (ET).

    \mysection{BTCV}
    BTCV (Beyond the Cranial Vault)
    abdomen dataset\cite{BTCV}. This dataset involves 30 training and
    20 testing subjects and 13 labelled organs: spleen, right kidney, left kidney, gallbladder, esophagus, liver, stomach, aorta, inferior vena cava, portal vein and splenic vein, pancreas, right adrenal gland and left adrenal gland. We combine the left and right adrenal gland into one. Scans are resampled to consistent resolution (224 $\times$ 224 $\times$85) and intensity scaled in the range [-175,250] Hounsfield Units (HU).

    \mysection{AMOS}
    AMOS Abdomen MRI \cite{AMOS} from the MICCAI AMOS Challenge, which consists of segmentation of abdominal organs from 100 MRI scans split equally into train and test sets. The organs include the liver, spleen, pancreas, kidneys, stomach, gallbladder, esophagus, aorta, inferior vena cava, adrenal glands, and duodenum.
    Scans are resampled to consistent resolution (256 $\times$ 256 $\times$ 125) and scans normalised for intensity channel wise in the range [0,1].
    
\subsection{Evaluation Metrics}
\label{subsec:Evaluation Metrics}
\begin{itemize}

\item \textbf{Dice Score}
The Dice Score, or Dice Coefficient, is a statistical measure used to assess the similarity between two samples. It is widely utilized in medical image analysis due to its sensitivity to variations in object size. The Dice Score is calculated by doubling the area of overlap between the predicted and ground truth segmentations and dividing by the total area of both. The formula is:
\[
\mathrm{Dice} = \frac{2 \times \mathrm{Area}(S_{\text{pred}} \cap S_{\text{gt}})}{\mathrm{Area}(S_{\text{pred}}) + \mathrm{Area}(S_{\text{gt}})}
\]
This metric ranges from 0 to 1, with a value of 1 indicating perfect agreement between the prediction and the ground truth. The Dice Score is particularly robust against variations in the size of the segmented objects, making it extremely useful in medical applications where such variability is common.

Both IoU and Dice Score offer comprehensive insights into model accuracy, with the Dice Score being especially effective in scenarios involving significant variations in object size.

\item \textbf{Hausdorff Distance}
The Hausdorff Distance is a metric used to measure the extent of discrepancy between two sets of points, often applied to evaluate the accuracy of object boundaries in image segmentation tasks. It is particularly useful for quantifying the worst-case scenario of the distance between the predicted segmentation boundary and the ground truth boundary.

The Hausdorff Distance calculates the greatest distance from a point in one set to the closest point in the other set. In image segmentation, this involves finding the largest distance from any point on the predicted boundary to the nearest point on the ground truth boundary, and vice versa. The mathematical definition is:
\[
\mathrm{HD} = \max\left\{ \sup_{p \in P} \inf_{q \in Q} d(p, q), \sup_{q \in Q} \inf_{p \in P} d(p, q) \right\}
\]
where \(P\) and \(Q\) are the sets of boundary points of the predicted segmentation and the ground truth segmentation, respectively, and \(d(p, q)\) represents the Euclidean distance between points \(p\) and \(q\).
   
    \end{itemize}

\subsection{Segmentation Details}

We perform a series of experiments to determine the best segmentation model on UKBOB using state-of-the-art multi-resolution CNN (UNet\cite{ronneberger2015unet}, SegResNet\cite{SegResNet}, nn-UNet\cite{nnUNet}) and transformer-based networks  (TransUNet\cite{transunet}, UNetr\cite{UNETR}, Swin-UNetr\cite{Hatamizadeh2022SwinUS}). We report segmentation performance in Table~\ref{tabsup:ukbb} where Swin-UNetr outperforms baselines by a margin, followed by nn-UNet.
We show visual examples of the 72 class labels in UKBOB in Figure~\ref{supp:UKBOB_Cor} and Figure~\ref{supp:UKBOB_Sag}.

\begin{figure}[t]
  \centering
  \includegraphics[trim={0cm 0 0cm 0cm},clip, width=0.9\linewidth]{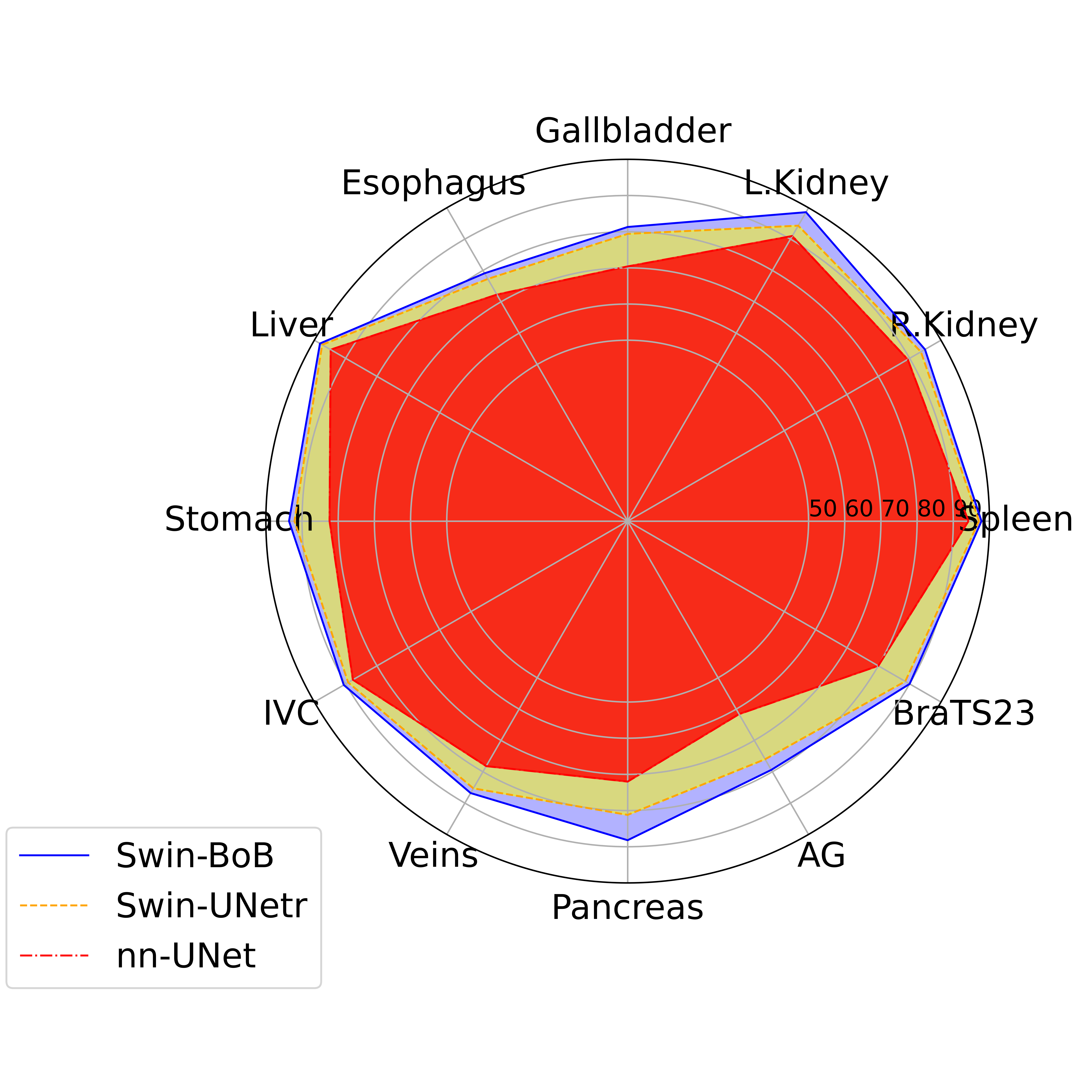}
\caption{\textbf{Per-Class Performance Comparison with Specialized Segmentation models.} We compare the Dice Score performance of our Swin-BOB model and baselines Swin-UNetr\cite{Hatamizadeh2022SwinUS} and nn-UNet\cite{nnUNet} on abdominal organ segmentation (BTCV) and brain tumour segmentation (BRATS).%
}
\label{fig:radar_plot_comparison}
\end{figure}

We also show detailed baseline comparison for BTCV and AMOS in Table~\ref{tab_supp:segmentation_comparison_btcv} and Table~\ref{tab_supp:mean_dice_score_amos} respectively. We provide radar plot in Figure~\ref{fig:radar_plot_comparison} that summarizes the performance of our segmentation model Swin-BOB compared to baseline segmentation models on different classes from BTCV and BRATS23 class average.

\begin{table*}[h!]
    \centering
    \resizebox{0.6\linewidth}{!}{%
    \begin{tabular}{c|c|c}
        \hline
        \textbf{Model} & \textbf{Mean Dice Score} & \textbf{Mean Hausdorff Distance} \\
        \hline
        UNet\cite{ronneberger2015unet} & 0.782 & 8.374 \\
        SegResNet\cite{SegResNet}& 0.794 & 7.912\\
        TransUNet\cite{transunet}& 0.838 & 6.258 \\
        UNetr\cite{UNETR} & 0.856 & 4.317 \\
        Swin-Unetr\cite{Hatamizadeh2022SwinUS} & 0.869 & 3.801 \\
        nn-UNet\cite{nnUNet}& 0.802 & 6.782\\
        AttentionUNet\cite{AttentionUNet} & 0.816 & 5.848\\
        \hline
    \end{tabular}
    }
        \vspace{2pt}
    \caption{\textbf{Comparison of segmentation model performance on BTCV (n = 12 classes).}}
    \label{tab_supp:segmentation_comparison_btcv}
\end{table*}

\begin{figure}[t]
  \centering
  \includegraphics[page=1,trim={0cm 0 0cm 0cm},clip, width=0.8\linewidth]{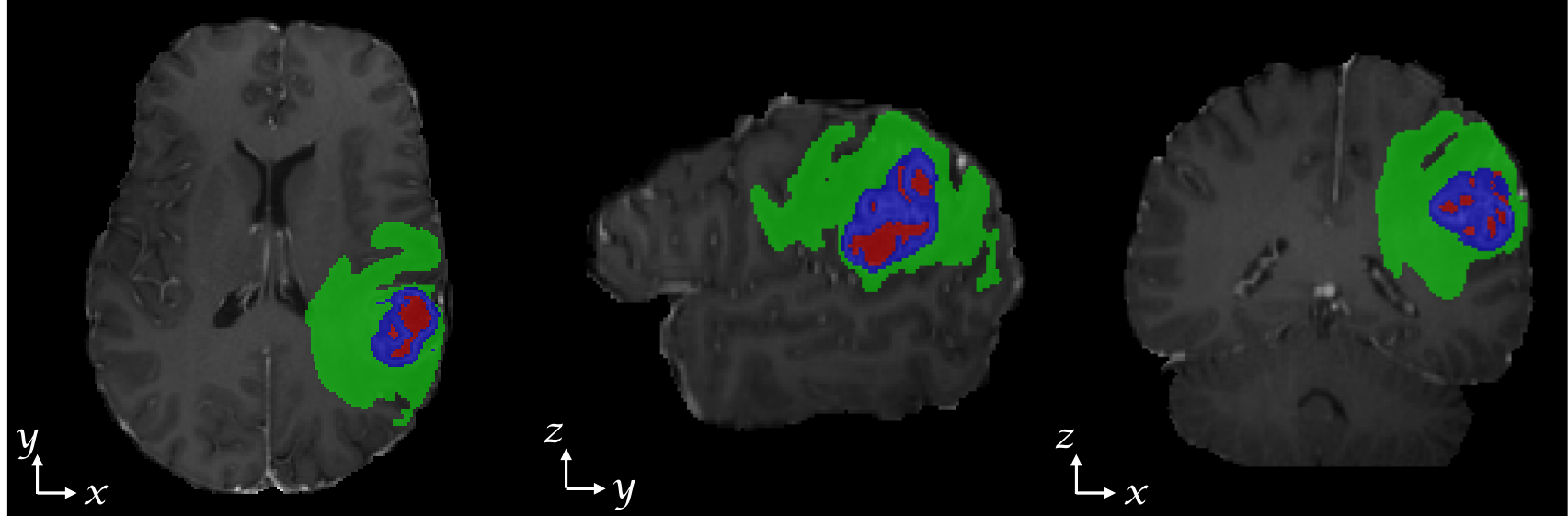}
  \includegraphics[page=3,trim={0cm 0 0cm 0cm},clip, width=0.8\linewidth]{images/BRATS/brats_filtering.pdf}
\caption{\textbf{Qualitative Performance on BRATS.} We show the ground-truth \textit{top} and output \textit{bottom} of our pre-trained Swin-BOB model for 3D segmentation on the brain tumour BRATS dataset with 3 tumour class labels \cite{Baid2021TheRB}. .
 }
\label{fig:brats}
\end{figure}

\begin{figure*}
    \centering
    \includegraphics[page=1, trim= 0.0cm 0.1cm 0.0cm 0cm,clip, width=0.19\linewidth]{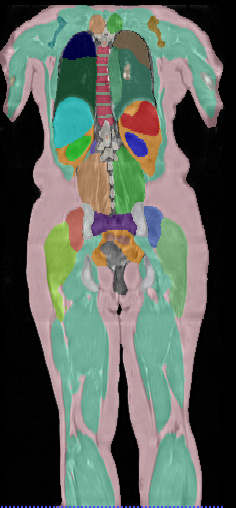} 
    \includegraphics[page=1, trim= 0.0cm 0.1cm 0.0cm 0cm,clip, width=0.19\linewidth]{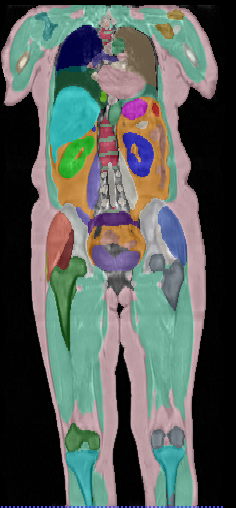}
    \includegraphics[page=1, trim= 0.0cm 0.1cm 0.0cm 0cm,clip, width=0.19\linewidth]{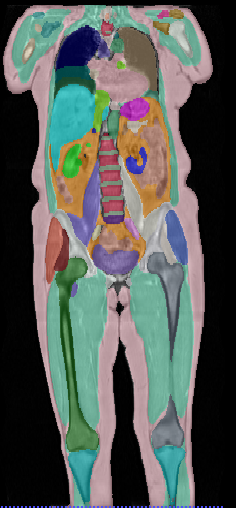}
    \includegraphics[page=1, trim= 0.0cm 0.1cm 0.0cm 0cm,clip, width=0.19\linewidth]{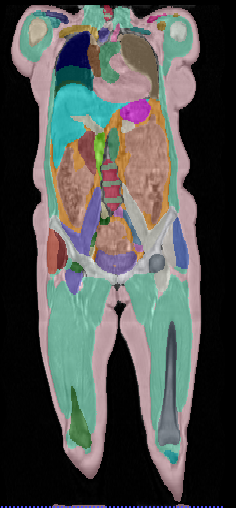}
    \includegraphics[trim= 0.0cm 0cm 0.0cm 0cm,clip, width=0.19\linewidth]{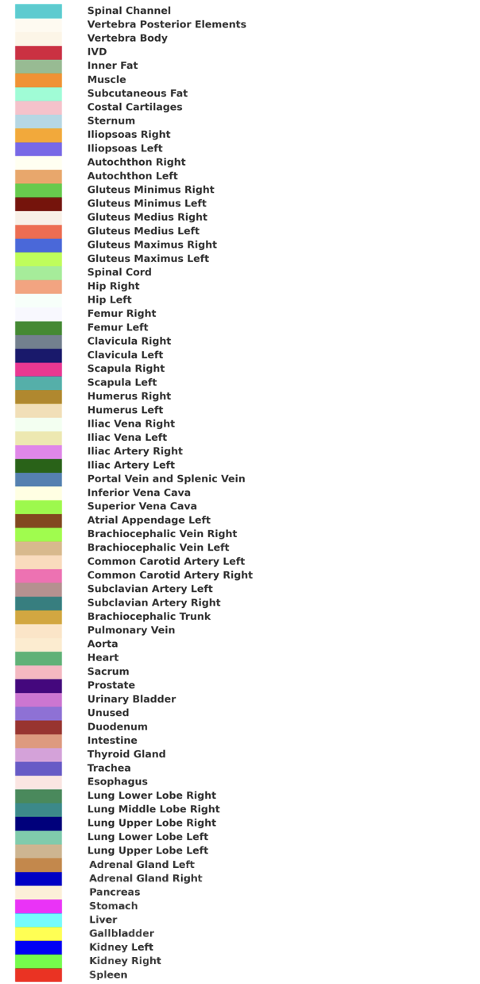}
    \caption{\textbf{Visualisation of UKBOB Segmentation Coronal Plane.} We show an example of 3D MRI from UKBOB for on coronal plane. %
    }
    
    \label{supp:UKBOB_Cor}
\end{figure*}

\begin{figure*}
    \centering
    \includegraphics[page=1, trim= 0.1cm 0cm 0.0cm 0cm,clip, width=0.19\linewidth]{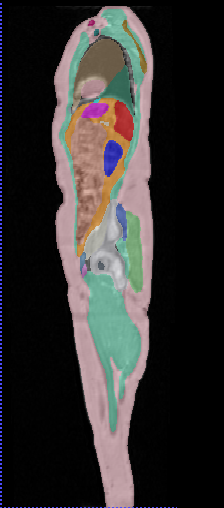}    \includegraphics[page=1, trim= 0.1cm 0cm 0.0cm 0cm,clip, width=0.19\linewidth]{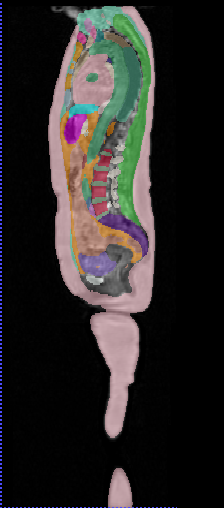}
    \includegraphics[page=1, trim= 0.1cm 0cm 0.0cm 0cm,clip, width=0.19\linewidth]{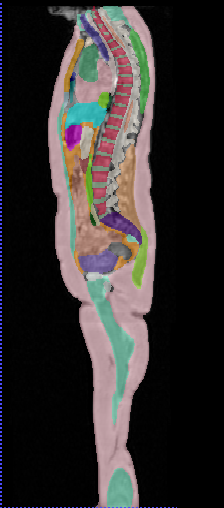}
    \includegraphics[page=1, trim= 0.1cm 0cm 0.0cm 0cm,clip, width=0.19\linewidth]{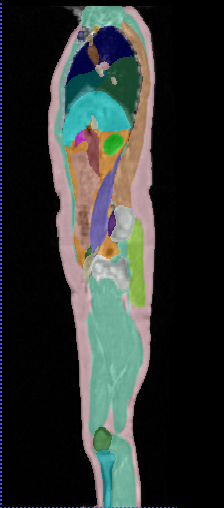}
    \includegraphics[trim= 0.0cm 0cm 0.0cm 0cm,clip, width=0.19\linewidth]{images/supp/ukbb_label_palette.png}
    \caption{\textbf{Visualisation of UKBOB Segmentation Sagittal Plane.} We show an example of 3D MRI from UKBOB for on sagittal plane.
    }
    \label{supp:UKBOB_Sag}
\end{figure*}

We show visual comparison on BRATS (Figure~\ref{fig:brats}) of our model segmentation relative to ground-truth.

\begin{table}[h]
\centering
\begin{tabular}{@{}lc@{}}
\toprule
\textbf{Model} & \textbf{Mean Dice Score} \\
\midrule
TransBTS\cite{transbts} & 0.792 \\
UNETR\cite{UNETR} & 0.762 \\
nnFormer\cite{nnFormer} & 0.790 \\
SwinUNETR\cite{Hatamizadeh2022SwinUS} & 0.880 \\
3D UX-Net\cite{3duxnet} & 0.900 \\
\bottomrule
\end{tabular}
        \vspace{2pt}
\caption{\textbf{Comparison of Segmentation Models for AMOS Segmentation (n = 14 classes).}}
\label{tab_supp:mean_dice_score_amos}
\end{table}

\begin{table}[h!]
\centering
\begin{tabular}{lcc}
\toprule
\textbf{Model} & \textbf{Mean Dice Score} & \textbf{Mean Hausdorff Distance} \\
\midrule
$\epsilon = 3$ & 0.891 & 7.126\\
$\epsilon = 2$ & 0.884 & 7.528\\
$\epsilon = 1$ & 0.792 & 8.247\\
$\epsilon = 4$ & 0.766 & 8.594\\
$\epsilon = 5$ & 0.745 & 8.972\\
\bottomrule
\end{tabular}
        \vspace{2pt}
\caption{\textbf{Effect of Filtration Threshold on Segmentation Performance on manual annotated set of abdomen organs (300) from UK Biobank.} The 11 abdomen organs and bones that have been manually annotated represent the overlap organs with BTCV\cite{BTCV} and UK Biobank\cite{graf2024totalvibesegmentator}.
}
\label{tab_supp:filtration_thresholds}
\end{table}

\begin{table}[h!]
\centering
\begin{tabular}{@{}lcc@{}}
\toprule
\textbf{Dataset} & \textbf{Mean Dice Score} & \textbf{Hausdorff Distance} \\ \midrule
AMOS & 0.831 & 7.647 \\
BTCV & 0.837 & 5.138 \\ \bottomrule
\end{tabular}
\caption{\textbf{Zero-shot performance on external datasets.}}
\label{tab:zeroshot}
\end{table}

\begin{table*}[h!]
    \centering
    \resizebox{0.9\linewidth}{!}{%
    \begin{tabular}{c|cccccccccc|c}
    \hline
    \textbf{Configuration}                                                           
    & \textbf{Spleen} & \textbf{R.Kid} & \textbf{L.Kid} & \textbf{Gall.} & \textbf{Eso.}  & \textbf{Liver} & \textbf{Stom.} & \textbf{IVC}& \textbf{AG} & \textbf{Aorta} 
    \\  \hline
    \textbf{AMOS} & 0.9084 & 0.9311 & 0.9421 & 0.6516 & 0.6582 & 0.9581 & 0.8216 & 0.8740 &0.5292 & 0.9062 
    \\ 
    \textbf{AMOS + filtering} & 0.9102 & 0.9397 & 0.9508 & 0.6582 & 0.6673 & 0.9662 & 0.8315 & 0.8824 & 0.6209 & 0.9183 
    \\  \hline
    \textbf{BTCV} & 0.883 & 0.884 & 0.932 & 0.795 & 0.790 & 0.946 & 0.885 &  0.871 & 0.784 & 0.799\\
    \textbf{BTCV + filtering} & 0.889 & 0.889 & 0.941 & 0.813 & 0.825 & 0.949 & 0.893 &  0.883 & 0.799& 0.869 \\
    \hline
    \end{tabular}
    }        \vspace{2pt}
        \caption{\textbf{Zero-shot 3D Segmentation Performance of Swin-BOB on AMOS external MRI data and CT (BTCV) for same organ classes}.}
    \label{tablesup:zero_shot_segmentation}
\end{table*}

\section{Dataset Filtration Details}

\subsection{Threshold Selection}

Full ablation experiments for threshold selection is available in Table~\ref{tab_supp:filtration_thresholds}. Results on impact of filtration on BTCV and AMOS are reported in Table~\ref{tablesup:zero_shot_segmentation}.
We therefore ensure high-quality labels by removing outliers adequately.

\subsection{Zero-Shot Generalization}
Our zero-shot evaluation on the AMOS and BTCV datasets highlights the robustness of filtered labels. Metrics are detailed in Table~\ref{tab:zeroshot}.

\subsection{Residual Label Noise}
While filtration reduces label noise, some false positives persist. To further improve the quality of the segmentation, we could incorporate human-in-the-loop approaches that turned efficient as shown in \cite{graf2024totalvibesegmentator,Bourigault23}.

\subsection{Filtering Out Patients Abnormalities}
One concern of automatic filtration is that it might filter out some natural abnormalities or pathologies in the patients, mistaken as wrong labels. We visualize some of these filtered-out labels in Figure~9 (main paper) and show that indeed lack quality labels rather than the patients have obvious abnormalities. To quantify this behavior, we measure the 50-sample average LPIPS distance (the lower the more similar) between any two 3D mid-abdominal slices from full UKBOB (0.315), between filtered/filtered-out samples (0.329), between filtered/filtered samples (0.303), and between filtered-out/filtered-out samples (0.339). This shows that all the distances are almost identical, indicating mostly homogeneous organs in the dataset partitioning and hence the filtration is mostly about the quality of the labels rather than filtering out patients with abnormality.

\section{Entropy Test-Time Adaptation (ETTA)}
\subsection{Algorithm Details}
In this section, we detail the algorithmic process for our test-time adaptation (ETTA). It works by  refining predictions minimizing entropy:
\[
L_{\text{ent}} = -\frac{1}{N} \sum_{i=1}^N \sum_{c=1}^C p_{i,c} \log p_{i,c}.
\]
During test time, only batch normalization parameters are fine-tuned while keeping other parameters fixed.
The method is simple, and efficient computationally since it does not require retraining the full model. We show detailed step by step procedure in Algorithm~\ref{supp:alg_bn}.

\begin{algorithm}[h!]
\caption{Algorithm for Test-Time Adaptation Using Batch Normalization}
\label{alg:test_time_bn}
\begin{algorithmic}[1]
\Require Pre-trained segmentation model $M$, test dataset $D_{test}$, loss function $\mathcal{L}$ (optional), optimizer $O$ (optional), epochs $N_{epochs}$ (optional)
\Ensure Adapted model $M'$

\State \textbf{Function} \textsc{FreezeExceptBN}($M$):
\State \hspace{1em} \textbf{For each parameter} $p$ \textbf{in} $M$:
\State \hspace{2em} \textbf{If} $p$ does not belong to a BatchNorm layer:
\State \hspace{3em} $p.requires\_grad \gets \textbf{False}$
\State \hspace{1em} \textbf{End For}
\State \textbf{End Function}

\State \textbf{Function} \textsc{UpdateBNStatistics}($M, D_{test}$):
\State \hspace{1em} Set $M$ to training mode: $M.train()$
\State \hspace{1em} \textsc{FreezeExceptBN}($M$)
\State \hspace{1em} \textbf{For each batch} $x \in D_{test}$:
\State \hspace{2em} Compute predictions: $M(x)$
\State \hspace{1em} \textbf{End For}
\State \textbf{End Function}

\State \textbf{Function} \textsc{FineTuneBN}($M, D_{test}, \mathcal{L}, O, N_{epochs}$):
\State \hspace{1em} Set $M$ to training mode: $M.train()$
\State \hspace{1em} \textsc{FreezeExceptBN}($M$)
\State \hspace{1em} \textbf{For} $epoch \gets 1$ \textbf{to} $N_{epochs}$:
\State \hspace{2em} \textbf{For each batch} $(x, y_{true}) \in D_{test}$:
\State \hspace{3em} Compute predictions: $y \gets M(x)$
\State \hspace{3em} Compute loss: $\ell \gets \mathcal{L}(y, y_{true})$
\State \hspace{3em} Zero gradients: $O.zero\_grad()$
\State \hspace{3em} Backpropagate: $\ell.backward()$
\State \hspace{3em} Update parameters: $O.step()$
\State \hspace{2em} \textbf{End For}
\State \hspace{1em} \textbf{End For}
\State \textbf{End Function}

\State \textbf{Define Function} \textsc{Infer}($M, D_{test}$):
\State \hspace{1em} Set $M$ to evaluation mode: $M.eval()$
\State \hspace{1em} \textbf{For each batch} $x \in D_{test}$:
\State \hspace{2em} Compute predictions: $y \gets M(x)$
\State \hspace{1em} \textbf{End For}
\State \textbf{End Function}

\end{algorithmic}
\label{supp:alg_bn}
\end{algorithm}

\begin{figure*}[t]
  \centering
  \includegraphics[width=1\linewidth,trim={1 4 1 4},clip]{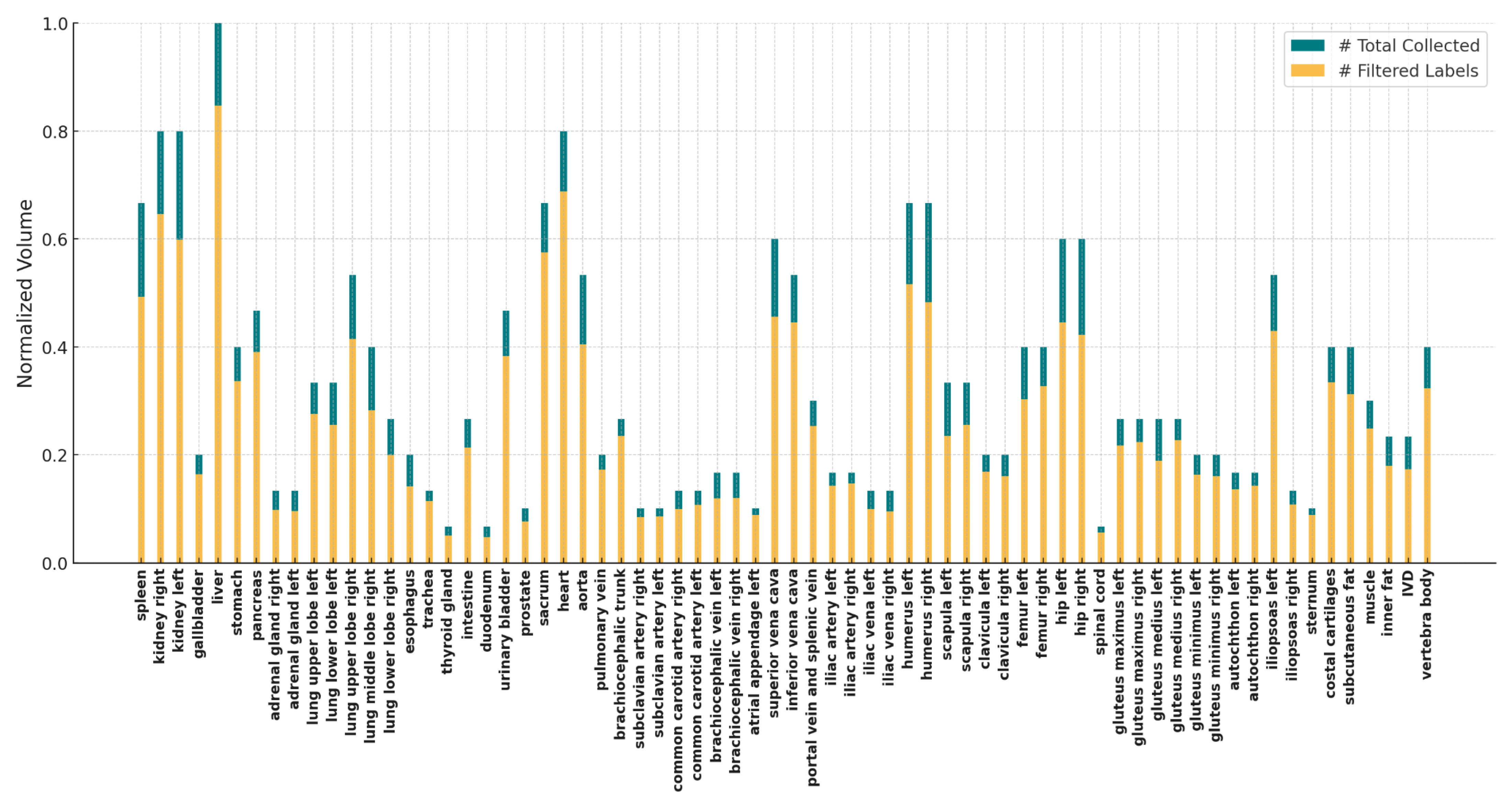}
\caption{\textbf{\methodname Distribution of Labels with our Filtration.} %
We show the distribution mean normalised volumes of 72 labels before and after filtration.}
\label{figsup:examples1_ukbb}
\end{figure*}

\begin{table*}[h]
\centering
\resizebox{0.5\linewidth}{!}{%
\begin{tabular}{c|ccccc}
\hline
\\

Model                                                           & ResUNet & UNetr & nnUNet  & Swin-UNetr & MedFormer \\ \hline
spleen
    & 0.91 & 0.92 & 0.94 & 0.94 & 0.93\\ 
kidney right
    & 0.87 & 0.89 & 0.91 & 0.92 & 0.90\\
kidney left
    & 0.88 & 0.90 & 0.92 & 0.93 & 0.91\\ 
gallbladder
    & 0.82 & 0.84 & 0.85 & 0.85 & 0.84\\
liver
    & 0.94 & 0.96 & 0.97 & 0.96 & 0.96\\
stomach
     & 0.88 & 0.89 & 0.90 & 0.91 & 0.89\\
pancreas
    & 0.85 & 0.87 & 0.89 & 0.90 & 0.88\\
adrenal gland right
    & 0.81 & 0.83 & 0.84 & 0.86 & 0.84\\
adrenal gland left
    & 0.81 & 0.83 & 0.84 & 0.86 & 0.83\\
lung upper lobe left
     & 0.93 & 0.94 & 0.96 & 0.96 & 0.95\\
lung lower lobe left
     & 0.94 & 0.95 & 0.96 & 0.96 & 0.94\\
lung upper lobe right
    & 0.94 & 0.95 & 0.96 & 0.96 & 0.94\\
lung middle lobe right
    & 0.93 & 0.94 & 0.96 & 0.96 & 0.95\\
lung lower lobe right
    & 0.93 & 0.95 & 0.96 & 0.96 & 0,95\\
esophagus
     & 0.86 & 0.88 & 0.9 & 0.91 & 0.89\\
trachea 
    & 0.87 & 0.89 & 0.92 & 0.92 & 0.91\\
thyroid gland
    & 0.74 & 0.75 & 0.76 & 0.77 & 0.75\\
intestine
    & 0.87 & 0.91 & 0.93 & 0.92 & 0.91\\
duodenum
    & 0.81 & 0.84 & 0.86 & 0.87 & 0.85\\
urinary bladder
    & 0.89 & 0.93 & 0.95 & 0.96 & 0.94\\
prostate
    & 0.91 & 0.92 & 0.94 & 0.94 & 0.94 \\
sacrum
    & 0.91 & 0.92 & 0.96 & 0.95 & 0.04\\
heart
    & 0.92 & 0.96 & 0.97 & 0.97 & 0.96\\
aorta
    & 0.91 & 0.93 & 0.95 & 0.94 & 0.93\\
pulmonary vein
    & 0.87 & 0.89 & 0.91 & 0.92 & 0.91\\
brachiocephalic trunk
    & 0.83 & 0.86 & 0.88 & 0.89 & 0.88\\
subclavian artery right
    & 0.81 & 0.85 & 0.86 & 0.88 & 0.86\\
subclavian artery left
    & 0.81 & 0.85 & 0.86 &0.88  & 0.86\\
common carotid artery right
    & 0.81 & 0.83 & 0.84 & 0.86 & 0.86\\
common carotid artery left  
    & 0.81 & 0.83 & 0.85 & 0.87 & 0.85\\
brachiocephalic vein left
    & 0.82 & 0.85 & 0.88 & 0.89 & 0.85\\
brachiocephalic vein right
    & 0.83 & 0.85 & 0.87 & 0.88 & 0.85\\
atrial appendage left   
    & 0.79 & 0.82 & 0.84 & 0.84 & 0.83\\
superior vena cava
    & 0.89 & 0.91 & 0.93 & 0.93 & 0.91\\
inferior vena cava
    & 0.89 & 0.90 & 0.92 & 0.92 & 0.90\\
portal vein and splenic vein
    & 0.76 & 0.79 & 0.82 & 0.82 & 0.80\\
iliac artery left
    & 0.83 & 0.85 & 0.87 & 0.87 & 0.85\\
iliac artery right
    & 0.82 & 0.84 & 0.86 & 0.86 & 0.84\\
iliac vena left
    & 0.85 & 0.88 & 0.91 & 0.91 & 0.89\\
iliac vena right
    & 0.85 & 0.88 & 0.90 & 0.90 & 0.88\\
humerus left
    & 0.90 & 0.93 & 0.94 & 0.93 & 0.93\\
humerus right
    & 0.90 & 0.93 & 0.94 & 0.94 & 0.93\\
scapula left
    & 0.86 & 0.89 & 0.91 & 0.91 & 0.89\\
scapula right
    & 0.88 & 0.89 & 0.91 & 0.91 & 0.89\\
clavicula left
    & 0.86 & 0.88 & 0.90 & 0.90 & 0.88\\
clavicula right
    & 0.86 & 0.88 & 0.90 & 0.90  & 0.88\\
femur left
    & 0.91 & 0.94 & 0.97 & 0.96 & 0.95\\
femur right
    & 0.90 & 0.93 & 0.96 & 0.95 & 0.93\\
hip left
    & 0.92 & 0.95 & 0.97 & 0.98 & 0.96\\
hip right
    & 0.91 & 0.94 & 0.96 & 0.97 & 0.95\\
spinal cord
    & 0.85 & 0.87 & 0.88 & 0.90 & 0.88\\
gluteus maximus left
    & 0.93 & 0.95 & 0.98 & 0.98 & 0.95\\
gluteus maximus right
    & 0.93 & 0.95  & 0.98 & 0.98 & 0.95\\
gluteus medius left
    & 0.94 & 0.97 & 0.98 & 0.98  & 0.97\\
gluteus medius right
    & 0.94 & 0.97 & 0.97 & 0.97  & 0.97\\
gluteus minimus left
    & 0.94 & 0.97 & 0.94 & 0.95 & 0.97\\
gluteus minimus right
    & 0.94 & 0.97 & 0.94 & 0.95 & 0.97\\
autochthon left
    & 0.94 & 0.96 & 0.97 & 0.97 & 0.96\\
autochthon right
    & 0.94 & 0.96 & 0.97 & 0.97 & 0.96\\
iliopsoas left
    & 0.92 & 0.94 & 0.96 & 0.96 & 0.95\\
iliopsoas right
    & 0.91 & 0.93 & 0.96 & 0.96 & 0.95\\
sternum
    & 0.86 & 0.88 & 0.92 & 0.92 & 0.89\\
costal cartilages
    & 0.85 & 0.87 & 0.90 & 0.91 & 0.88\\
subcutaneous fat
    & 0.89 & 0.92 & 0.95 & 0.96 & 0.93\\
muscle
    & 0.91 & 0.93 & 0.96 & 0.97 & 0.94\\
inner fat
    & 0.86 & 0.88 & 0.90 & 0.91 & 0.89\\
IVD
    & 0.86 & 0.88 & 0.90 & 0.91 & 0.88\\
vertebra body
    & 0.89 & 0.92 & 0.94 & 0.94 & 0.93\\
vertebra posterior elements
    & 0.82 & 0.84 & 0.86 & 0.88 & 0.85\\
spinal channel   & 0.87 & 0.89 & 0.91 & 0.91 & 0.89\\
bone other  & 0.82 & 0.84 & 0.86 & 0.87 & 0.84 \\
\hline
\end{tabular}}
        \vspace{2pt}
\caption{\textbf{3D Segmentation Performance on UK Biobank dataset}. We compare our \methodname on 3D medical segmentation task on the UK Biobank test set (n=10,353) with 5-fold cross validation compared to other methods using the average Dice score and average Haussdorff Distance (HD) per class as metric. Standard deviations are shown next to the mean Dice Score and HD values.}
\label{tabsup:ukbb}
\end{table*}

\section{Dataset Access and Code for Reproducibility}
The dataset and pre-trained Swin-BOB models will be made available publicly via UK Biobank. Documentation for reproducing experiments is provided in supplementary materials. Code is available at \url{https://anonymous.4open.science/r/UK_BOB-ACF1/}

\end{document}